\pdfoutput=1
\documentclass{article}
    \PassOptionsToPackage{numbers, sort&compress}{natbib}

 \usepackage[eandd,final]{neurips_2026}

\usepackage[utf8]{inputenc} 
\usepackage[T1]{fontenc}    
\usepackage{hyperref}       
\usepackage{url}            
\usepackage{titletoc}
\usepackage{etoc} 
\usepackage{booktabs}       
\usepackage{amsfonts}       
\usepackage{nicefrac}       
\usepackage{microtype}      
\usepackage[dvipsnames]{xcolor}         
\usepackage{tcolorbox}
\usepackage{amsmath}
\usepackage{soul} 
\usepackage{multirow}
\usepackage{booktabs}
\usepackage{adjustbox}
\usepackage{hhline}
\usepackage{tabularx} 
\usepackage{subcaption}
\usepackage{enumitem}
\usepackage{wrapfig}
\usepackage{needspace}
\usepackage{algorithm}
\usepackage{algpseudocode}
\usepackage{colortbl}
\usepackage{tcolorbox}
\tcbuselibrary{breakable}
\usepackage{xparse}
\usepackage{stfloats}
\usepackage{float}
\usepackage{highlight}
\usepackage{adjustbox}
\usepackage{graphicx}
\usepackage{makecell}
\usepackage{longtable}
\usepackage{upgreek}
\usepackage{tikz}
\usepackage{listings}
\definecolor{lstkw}{HTML}{0033B3}    
\definecolor{lststr}{HTML}{067D17}   
\definecolor{lstcom}{HTML}{8C8C8C}   
\definecolor{lsthead}{HTML}{B8860B}  
\lstdefinelanguage{mdlite}{
  morecomment=[l]{\#},
  morestring=[b]`,
  morekeywords={Workspace,Rules},
  sensitive=true,
}
\lstdefinelanguage{jsonlite}{
  morestring=[b]",
  morekeywords={true,false,null},
  literate={:}{{{\color{lstkw}:}}}1
           {\{}{{{\color{lstkw}\{}}}1
           {\}}{{{\color{lstkw}\}}}}1
           {[}{{{\color{lstkw}[}}}1
           {]}{{{\color{lstkw}]}}}1,
  sensitive=true,
}
\usetikzlibrary{positioning,fit,backgrounds,arrows.meta,calc}
\usepackage[dvipsnames]{xcolor}
\definecolor{ForestGreen}{RGB}{34,139,34}

\usepackage{authblk}
\usepackage{authblk}

\makeatletter
\renewcommand\AB@affilsepx{, \protect\Affilfont}
\makeatother

\usepackage{amsmath,amsfonts,bm}

\def\eqref#1{equation~\ref{#1}}

\def\1{\bm{1}}

\DeclareMathAlphabet{\mathsfit}{\encodingdefault}{\sfdefault}{m}{sl}
\SetMathAlphabet{\mathsfit}{bold}{\encodingdefault}{\sfdefault}{bx}{n}

\usepackage[nameinlink,noabbrev]{cleveref}

\crefname{appendix}{appendix}{appendices}
\Crefname{appendix}{Appendix}{Appendices}

\DeclareRobustCommand{\ourmethod}{%
  \texorpdfstring{\raisebox{0.25ex}{\scalebox{1.1}{$\upchi$}}-Bench}{chi-Bench}%
}

\DeclareRobustCommand{\ourworld}{%
  \texorpdfstring{\raisebox{0.25ex}{\scalebox{1.1}{$\upchi$}}-World}{chi-World}%
}

\newcommand{\symbolimg}[2][0.3cm]{%
  \ensuremath{\vcenter{\hbox{\includegraphics[height=#1]{#2}}}}%
}

\newcolumntype{/}{!{\color{white}\vline width 7pt}}

\newcommand{\sharded}{\symbolimg[0.35cm]{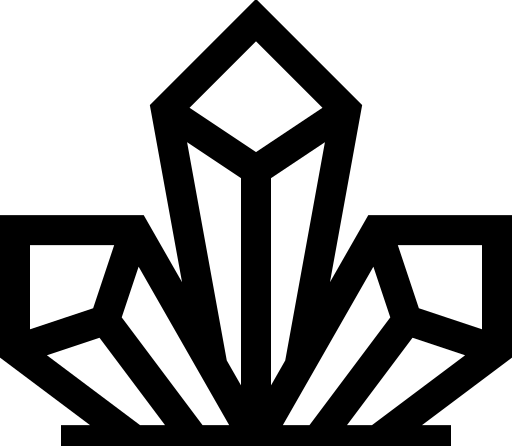}}

\newcommand{\huggingface}{\symbolimg[0.3cm]{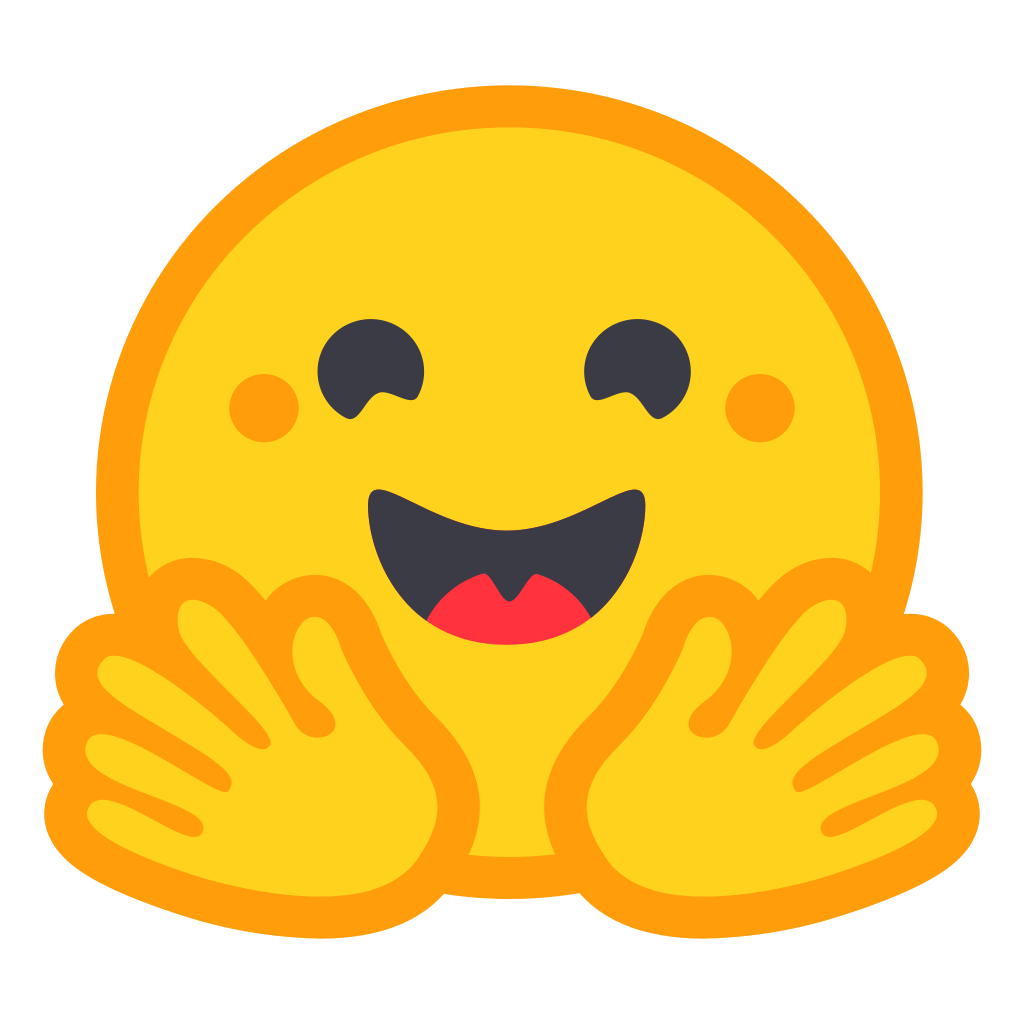}}
\newcommand{\github}{\symbolimg[0.3cm]{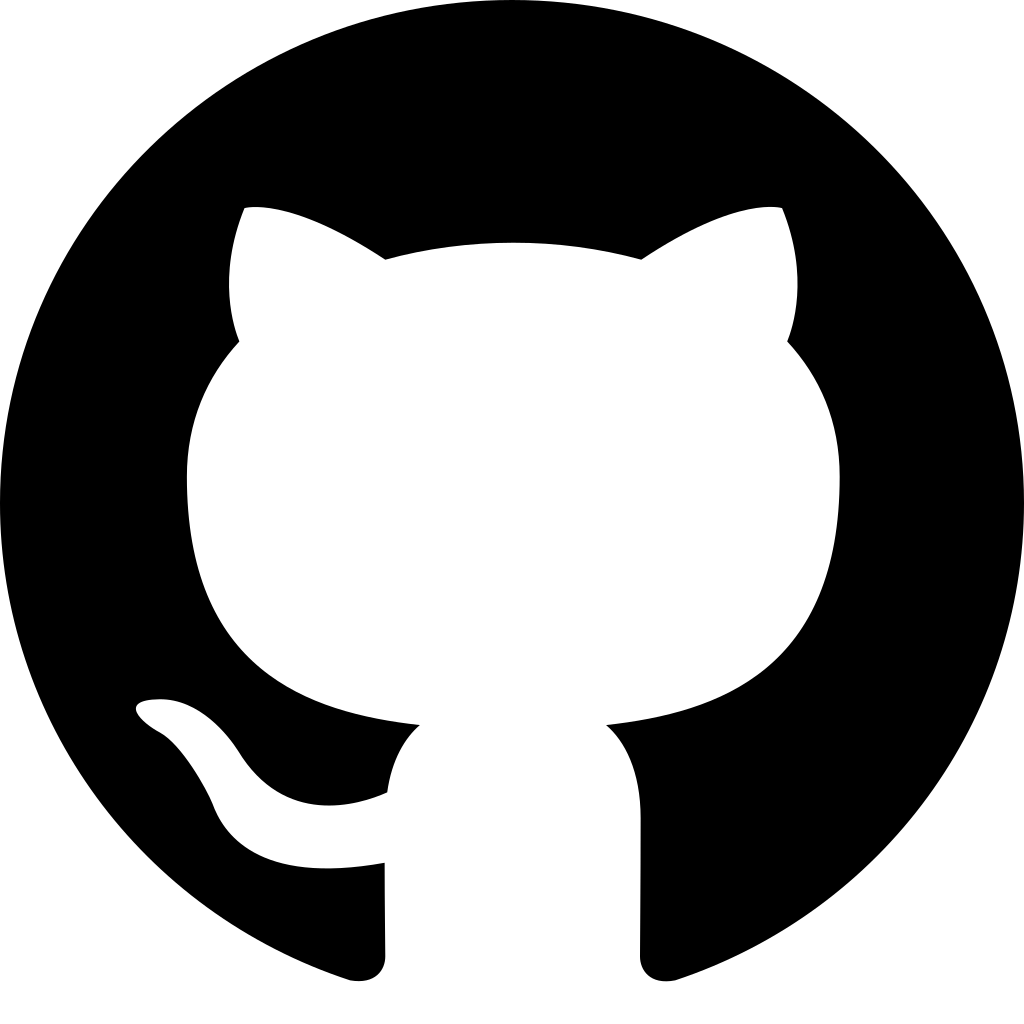}}

\definecolor{coolblue}{HTML}{6D9EEB}
\definecolor{warmgold}{HTML}{F1C232}

\usepackage{pifont}
\newcommand{\cmark}{\textcolor{ForestGreen}{\ding{51}}}
\newcommand{\xmark}{\textcolor{OrangeRed}{\ding{55}}}
\newcommand{\pmark}{\textcolor{orange}{\ding{109}}}

\newcommand{\ico}[1]{\raisebox{-0.18em}{\includegraphics[height=1em]{figures/icons/#1}}\,}

\newcommand{\totalnumapps}{20}
\newcommand{\totalnumroutes}{151}
\newcommand{\totalnummcptools}{87}
\newcommand{\engineloc}{${\sim}$115K}
\newcommand{\totalnumstates}{29}

\newcommand{\totalnumactivities}{${\sim}$5{,}000}
\newcommand{\totalnumpatients}{50}
\newcommand{\totalnumworkers}{${\sim}$90}

\definecolor{HeatA}{HTML}{FCE4E4} 
\definecolor{HeatB}{HTML}{FDEED9} 
\definecolor{HeatC}{HTML}{FFF7DC} 
\definecolor{HeatD}{HTML}{ECF3DC} 
\definecolor{HeatE}{HTML}{DEEEDB} 

\newcommand{\heat}[1]{%
  \ifdim #1pt<5pt\cellcolor{HeatA}%
  \else\ifdim #1pt<15pt\cellcolor{HeatB}%
  \else\ifdim #1pt<25pt\cellcolor{HeatC}%
  \else\ifdim #1pt<40pt\cellcolor{HeatD}%
  \else\cellcolor{HeatE}%
  \fi\fi\fi\fi
  #1}

\newcommand{\heatbf}[1]{%
  \ifdim #1pt<5pt\cellcolor{HeatA}%
  \else\ifdim #1pt<15pt\cellcolor{HeatB}%
  \else\ifdim #1pt<25pt\cellcolor{HeatC}%
  \else\ifdim #1pt<40pt\cellcolor{HeatD}%
  \else\cellcolor{HeatE}%
  \fi\fi\fi\fi
  \textbf{#1}}

\newcommand{\heats}[2]{%
  \ifdim #1pt<5pt\cellcolor{HeatA}%
  \else\ifdim #1pt<15pt\cellcolor{HeatB}%
  \else\ifdim #1pt<25pt\cellcolor{HeatC}%
  \else\ifdim #1pt<40pt\cellcolor{HeatD}%
  \else\cellcolor{HeatE}%
  \fi\fi\fi\fi
  $#1_{\pm#2}$}

\newcommand{\heatsbf}[2]{%
  \ifdim #1pt<5pt\cellcolor{HeatA}%
  \else\ifdim #1pt<15pt\cellcolor{HeatB}%
  \else\ifdim #1pt<25pt\cellcolor{HeatC}%
  \else\ifdim #1pt<40pt\cellcolor{HeatD}%
  \else\cellcolor{HeatE}%
  \fi\fi\fi\fi
  $\bm{#1}_{\pm#2}$}

\newcommand{\heatci}[3]{%
  \ifdim #1pt<5pt\cellcolor{HeatA}%
  \else\ifdim #1pt<15pt\cellcolor{HeatB}%
  \else\ifdim #1pt<25pt\cellcolor{HeatC}%
  \else\ifdim #1pt<40pt\cellcolor{HeatD}%
  \else\cellcolor{HeatE}%
  \fi\fi\fi\fi
  $#1_{\scriptscriptstyle -#2}^{\scriptscriptstyle +#3}$}
\newcommand{\heatcibf}[3]{%
  \ifdim #1pt<5pt\cellcolor{HeatA}%
  \else\ifdim #1pt<15pt\cellcolor{HeatB}%
  \else\ifdim #1pt<25pt\cellcolor{HeatC}%
  \else\ifdim #1pt<40pt\cellcolor{HeatD}%
  \else\cellcolor{HeatE}%
  \fi\fi\fi\fi
  $\bm{#1}_{\scriptscriptstyle -#2}^{\scriptscriptstyle +#3}$}
\newcommand{\heatciplain}[3]{%
  \ifdim #1pt<5pt\cellcolor{HeatA}%
  \else\ifdim #1pt<15pt\cellcolor{HeatB}%
  \else\ifdim #1pt<25pt\cellcolor{HeatC}%
  \else\ifdim #1pt<40pt\cellcolor{HeatD}%
  \else\cellcolor{HeatE}%
  \fi\fi\fi\fi
  $#1_{\scriptscriptstyle -#2}^{\scriptscriptstyle +#3}$}

\newcommand{\heatsteps}[1]{%
  \ifdim #1pt<30pt\cellcolor{HeatE}%
  \else\ifdim #1pt<50pt\cellcolor{HeatD}%
  \else\ifdim #1pt<75pt\cellcolor{HeatC}%
  \else\ifdim #1pt<100pt\cellcolor{HeatB}%
  \else\cellcolor{HeatA}%
  \fi\fi\fi\fi
  #1}

\newcommand{\heattokens}[2]{%
  \ifdim #1pt<1.5pt\cellcolor{HeatE}%
  \else\ifdim #1pt<2.5pt\cellcolor{HeatD}%
  \else\ifdim #1pt<4pt\cellcolor{HeatC}%
  \else\ifdim #1pt<7pt\cellcolor{HeatB}%
  \else\cellcolor{HeatA}%
  \fi\fi\fi\fi
  #2}

\newcommand{\heattime}[1]{%
  \ifdim #1pt<5pt\cellcolor{HeatE}%
  \else\ifdim #1pt<9pt\cellcolor{HeatD}%
  \else\ifdim #1pt<13pt\cellcolor{HeatC}%
  \else\ifdim #1pt<18pt\cellcolor{HeatB}%
  \else\cellcolor{HeatA}%
  \fi\fi\fi\fi
  #1}

\newcommand{\heatcost}[2]{%
  \ifdim #1pt<0.5pt\cellcolor{HeatE}%
  \else\ifdim #1pt<1.5pt\cellcolor{HeatD}%
  \else\ifdim #1pt<3pt\cellcolor{HeatC}%
  \else\ifdim #1pt<7pt\cellcolor{HeatB}%
  \else\cellcolor{HeatA}%
  \fi\fi\fi\fi
  #2}

\makeatletter
\providecommand{\slug}[1]{%
  \begingroup
    \let\hv@stdunder\_%
    \renewcommand{\_}{\hv@stdunder\penalty200\hskip0pt\relax}%
    \texttt{#1}%
  \endgroup
}
\providecommand{\slugd}[2]{%
  \begingroup
    \let\hv@stdunder\_%
    \renewcommand{\_}{\hv@stdunder\penalty200\hskip0pt\relax}%
    \texttt{#1.\penalty200\hskip0pt\relax#2}%
  \endgroup
}
\makeatother

\title{\raisebox{0.25ex}{\scalebox{1.1}{$\bm{\chi}$}}-Bench: Can AI Agents Automate End-to-End, Long-Horizon, Policy-Rich Healthcare Workflows?}

\author[1]{Haolin Chen}
\author[1]{Deon Metelski}
\author[1]{Leon Qi}
\author[1]{Tao Xia}
\author[1]{Joonyul Lee}
\author[1]{Steve Brown}
\author[1]{Kevin Riley}
\author[1]{Frank Wang}
\author[2]{T. Y. Alvin Liu, MD}
\author[3]{Hank Capps, MD}
\author[4]{Zeyu Tang}
\author[5]{Xiangchen Song}
\author[5]{Lingjing Kong}
\author[6]{Fan Feng}
\author[7]{Tianyi Zeng}
\author[8]{Zhiwei Liu}
\author[9]{Zixian Ma}
\author[10]{Hang Jiang}
\author[11]{\\Fangli Geng}
\author[12]{Yuan Yuan}
\author[13]{Chenyu You}
\author[14]{Qingsong Wen}
\author[15]{Hua Wei}
\author[15]{Yanjie Fu}
\author[16]{\\Yue Zhao}
\author[17]{Carl Yang}
\author[6]{Biwei Huang}
\author[5,19]{Kun Zhang}
\author[18]{Caiming Xiong}
\author[4]{\\Sanmi Koyejo}
\author[19,5]{Eric P. Xing}
\author[20]{Philip S. Yu}
\author[1]{Weiran Yao}

\affil[1]{\href{https://actava.ai/}{{Actava AI}}}
\affil[2]{Johns Hopkins Medicine}
\affil[3]{Wellstar Health System}
\affil[4]{Stanford University}
\affil[5]{CMU}
\affil[6]{UCSD}
\affil[7]{Yale School of Medicine}
\affil[8]{Salesforce AI Research}
\affil[9]{University of Washington}
\affil[10]{Northeastern University}
\affil[11]{Brown University}
\affil[12]{Boston College}
\affil[13]{Stony Brook University}
\affil[14]{University of Oxford}
\affil[15]{Arizona State University}
\affil[16]{University of Southern California}
\affil[17]{Emory University}
\affil[18]{Recursive Superintelligence}
\affil[19]{MBZUAI}
\affil[20]{University of Illinois at Chicago}

\hypersetup{hidelinks}
\begin{document}

\maketitle

\vspace{-20pt}
\begin{figure}[ht]
    \centering
    \includegraphics[width=\textwidth]{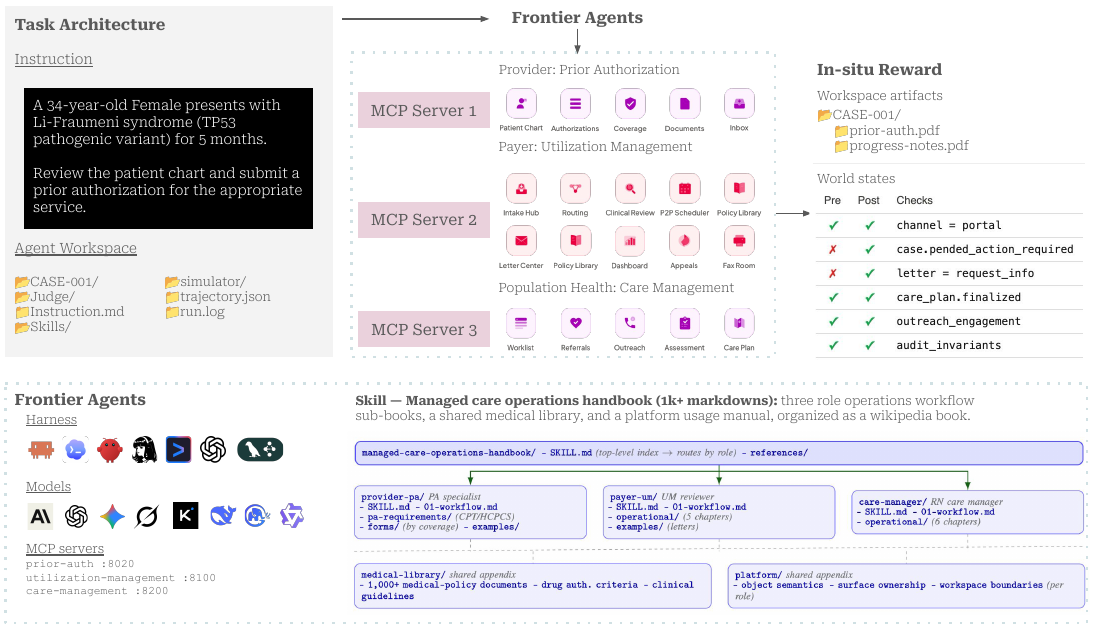}
    \caption{\ourmethod: \textbf{\underline{C}}linical \textbf{\underline{H}}ealthcare \textbf{\underline{I}}n-Situ Environment and Evaluation Benchmark.}
    \label{fig:architecture}
\end{figure}

\begin{abstract}

End-to-end automation of realistic healthcare operations stresses three capabilities underrepresented in current benchmarks: \textit{policy density}, decisions must be grounded in a large library of medical, insurance, and operational rules; \textit{multi-role composition}, a single task requires the agent to play multiple roles with handoffs; and \textit{multilateral interaction}: intermediate workflow steps are multi-turn dialogs, such as peer-to-peer review and patient outreach. We introduce \ourmethod, a benchmark of long-horizon healthcare workflows across three domains: provider prior authorization, payer utilization management, and care management. Each task hands the agent a clinical case in a high-fidelity simulator of \totalnumapps{} healthcare apps exposed via \totalnummcptools{} MCP tools, which it must drive to a terminal status through tool calls and writing the role's artifacts, guided by a {1{,}290}-document \textit{managed-care operations handbook} skill. Across 30 agent harness/model configurations, the best agent resolves only \textbf{28.0\%} of tasks, no agent clears \textbf{20\%} on strict pass\textasciicircum{}3, and executing all tasks in a single session slumps the performance to \textbf{3.8\%}. These results raise the hypothesis that similar gaps are likely to surface in other policy-dense, role-composed, irreversible enterprise domains.

\end{abstract}

\vspace{-2ex}
\begin{center}
\href{https://actava.ai/benchmarks}{\sharded~\texttt{actava.ai/benchmarks}} \;
\href{https://github.com/actava-ai/chi-bench}{\github~\texttt{actava-ai/chi-bench}} \;
\href{https://huggingface.co/datasets/actava/chi-bench}{\huggingface~\texttt{actava/chi-bench}}
\end{center}
\vspace{-2ex}

\section{Introduction}
\label{sec:intro}

The U.S. healthcare system is an administrative nightmare~\cite{cutler2012reducing,sahni2023active}. \textit{Prior authorization (PA)}, where providers (e.g., hospitals) prepare clinical documents for payers (e.g., insurers) review to justify a service or medication, is one of the most common yet inefficient workflows~\cite{sahni2024perceptions,sinsky2016allocation,ama2024survey}. \textit{Care management (CM)}, a long-term patient-assisting program, follows a similar arc~{\cite{karam2026nurses,friedman2016facilitators,ju2022improving}}: referrals queue for weeks, staff spend hours outreaching patients, and coordination across roles buries nurses in work they didn't sign up for. These are \textit{long-horizon, policy-grounded} tasks where every handoff is a chance for things to stall. AI agents are increasingly proposed as a way to assist or partially automate such work. Already, frontier agents now sustain hundreds of tool calls over hours of execution, automating long-horizon tasks that were out of reach a year ago.

However, end-to-end automation of \textit{realistic healthcare workflows} tells a different story, posing three underexplored challenges that possibility warrants rigorous stress-testing:

\textbf{1) Policy density.} Every agent decision must be grounded in policy, e.g., medical guidelines, insurance rules, operational procedures that vary across providers and payers and shifts over time. Agents must navigate a large policy library, interpret conditions correctly, and adhere to them across long tool-call chains. \textbf{2) Multi-role composition.} An end-to-end workflow is divided among roles such as clinician, coordinator, UM nurse, medical director, and RN care manager. An agent must possess all of their domain knowledge, switch context and goals as the case moves. Handoffs are terminal: once a step is submitted or routed, it cannot be edited or re-run. \textbf{3) Multilateral interactions.} Some steps are not tool calls but multi-turn conversations, such as payer-provider peer-to-peer review, requests for information, or care manager outreach to patients. Agents must shift from background execution to live dialog, collect information incrementally from humans, and carry results back to  workflow. 

These challenges are not edge cases; they are the daily reality of managed-care operations, where the bulk of work centers on prior authorization, utilization management review, and care management.

Inspired by these, we introduce \ourmethod, a benchmark that evaluates frontier agents in these three realistic, end-to-end healthcare workflow settings. As shown in \Cref{fig:architecture}, each task hands the agent a case (a provider PA, a payer UM review, or an RN care management) in a high-fidelity simulator of \totalnumapps~healthcare apps exposed via MCP. The agent must drive the case to a terminal status by issuing tool calls and writing the role's artifacts (submission packets, review notes, letters, care plans), guided by a \textit{managed-care operations handbook} skill ({1{,}290} markdowns) of workflows, platform usage, and medical/insurance policy. The resulting world state, artifacts and event trail are scored \textit{in-situ} by a composite verifier that combines deterministic checks with rubric-based LLM judge.

\begin{figure}[ht]
    \centering
    \includegraphics[width=\linewidth]{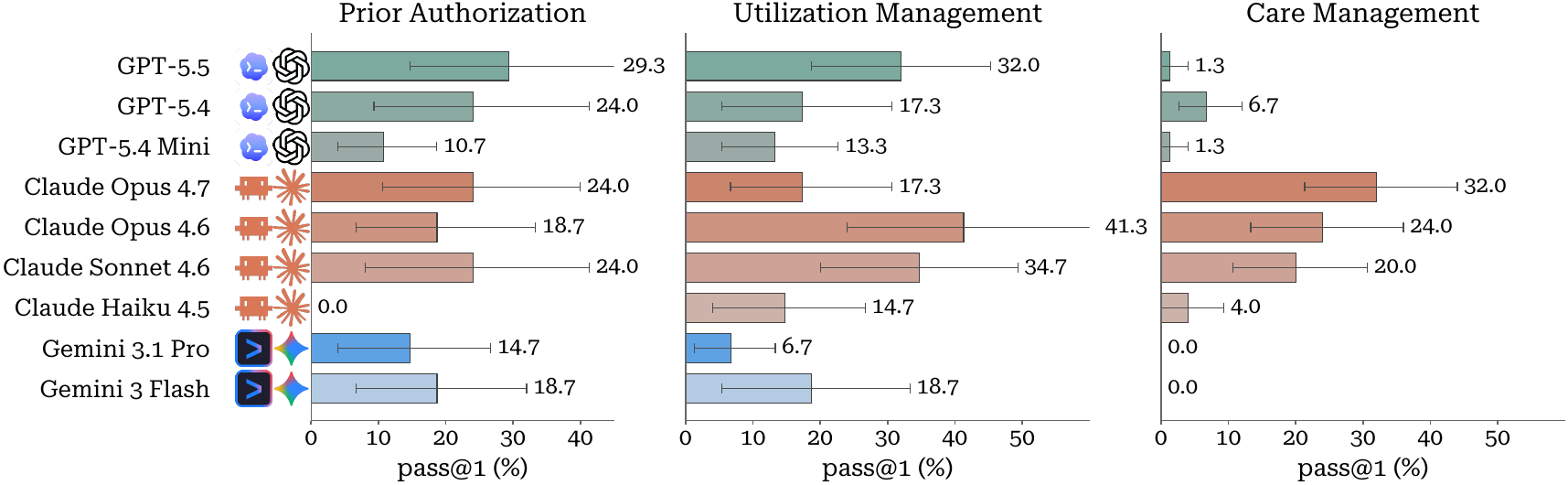}
    \caption{pass@$1$ across the three \ourmethod{} environments of frontier proprietary LLMs with their first-party agent harness. Error bars are task-level percentile bootstrap 95\% confidence intervals.}
    \label{fig:teaser}
\end{figure}

We evaluated 30 agent harness/model configurations spanning major frontier models and strong agent stacks. As shown in \Cref{fig:teaser}, \ourmethod{} is far from solved. The best configuration, {\bf \color[HTML]{CC785C} Claude Code+Claude Opus~4.6}, resolves only \textbf{28.0\%} of tasks at pass@1; no agent clears \textbf{20\%} under the strict pass\textasciicircum{}3 reliability metric; and the marathon run, where agents execute all tasks in a single session, drops to \textbf{3.8\%}, and the end-to-end provider--payer arena collapses the best prior auth agents to \textbf{0\%}. These results suggest that the long-horizon capabilities frontier agents demonstrate on coding-style benchmarks do not generalize well to realistic healthcare workflows, and we expect similar gaps in other policy-dense, role-composed, irreversible enterprise domains beyond.

\section{Related work}
\paragraph{Healthcare AI benchmarks.}
Prior healthcare benchmarks evaluate one of: factual medical knowledge~\citep{jin2021medqa,pal2022medmcqa,jin2019pubmedqa,
tsatsaronis2015bioasq,xiong2024mirage,zuo2025medxpertqa},
broad clinical LLM proficiency~\citep{bedi2025medhelm,
arora2025healthbenchevaluatinglargelanguage}, EHR querying~\citep{lee2022ehrsql,khandekar2024medcalc,tang2024biocoder,wang2024biodsbench,wornow2023ehrshot}, short-horizon clinical agents~\citep{jiang2025medagentbench,schmidgall2024agentclinic,liu2024medchain,shi2025medagentgym}, or narrower administrative interactions~\citep{jiang2025medagentbench,bedi2026healthadminbench}. \ourmethod{} is the first to combine, in a single task, long-horizon tool calls, explicit dense policy retrieval, irreversible workflow state, hidden multilateral interaction, and in-situ verification against persisted simulator state. HealthAdminBench~\citep{bedi2026healthadminbench}, the closest peer, focuses on GUI interaction across an EHR, two payer portals, and a fax system via pixel/DOM browsings; while \ourmethod~instead exposes apps via structured MCP tools and a large explicit policy handbook skill. We also add the care management domain with patient outreach. 

\begin{table}[t]
    \centering
    \caption{Coverage matrix of nine evaluation axes across $29$ healthcare and long-horizon agent benchmarks, characterizing the task surface each one targets; per-axis definitions and per-benchmark cell-by-cell justifications are in \Cref{app:benchmark-comparison}. \cmark = supported, \pmark = partially supported, \xmark = not supported.\label{tab:review}}
    \resizebox{\textwidth}{!}{%
    \begin{tabular}{lccccccccc}
        \toprule
        \textbf{Benchmark} & \textbf{Healthcare} & \textbf{API Tools} & \textbf{Long-horiz.}
        & \textbf{Policy Density} & \textbf{Multi-role Comp.} & \textbf{Multilateral}
        & \textbf{Hidden state} & \textbf{In-Situ} & \textbf{LLM judge} \\
        \midrule
        MedQA~\citep{jin2021medqa}              & \cmark & \xmark & \xmark & \xmark & \xmark & \xmark & \xmark & \xmark & \xmark \\
        MedMCQA~\citep{pal2022medmcqa}          & \cmark & \xmark & \xmark & \xmark & \xmark & \xmark & \xmark & \xmark & \xmark \\
        PubMedQA~\citep{jin2019pubmedqa}        & \cmark & \xmark & \xmark & \xmark & \xmark & \xmark & \xmark & \xmark & \xmark \\
        BioASQ~\citep{tsatsaronis2015bioasq}    & \cmark & \xmark & \xmark & \xmark & \xmark & \xmark & \xmark & \xmark & \xmark \\
        MIRAGE~\citep{xiong2024mirage}          & \cmark & \pmark & \xmark & \xmark & \xmark & \xmark & \xmark & \xmark & \xmark \\
        MedCalc-Bench~\citep{khandekar2024medcalc} & \cmark & \xmark & \xmark & \xmark & \xmark & \xmark & \xmark & \xmark & \xmark \\
        EHRSQL~\citep{lee2022ehrsql}            & \cmark & \cmark & \xmark & \xmark & \xmark & \xmark & \xmark & \xmark & \xmark \\
        BioCoder~\citep{tang2024biocoder}       & \cmark & \cmark & \xmark & \xmark & \xmark & \xmark & \xmark & \pmark & \xmark \\
        BioDSBench~\citep{wang2024biodsbench}   & \cmark & \cmark & \pmark & \xmark & \xmark & \xmark & \xmark & \pmark & \xmark \\
        EHRSHOT~\citep{wornow2023ehrshot}       & \cmark & \xmark & \xmark & \xmark & \xmark & \xmark & \xmark & \xmark & \xmark \\
        MedHELM~\citep{bedi2025medhelm}          & \cmark & \xmark & \xmark & \xmark & \xmark & \xmark & \xmark & \xmark & \cmark \\
        MedXpertQA~\citep{zuo2025medxpertqa}      & \cmark & \xmark & \xmark & \xmark & \xmark & \xmark & \xmark & \xmark & \xmark \\
        HealthBench~\citep{arora2025healthbenchevaluatinglargelanguage}  & \cmark & \xmark & \pmark & \xmark & \xmark & \pmark & \xmark & \xmark & \cmark \\
        MedAgentsBench~\citep{tang2025medagentsbench} & \cmark & \pmark & \xmark & \xmark & \xmark & \xmark & \xmark & \xmark & \xmark \\
        AgentClinic~\citep{schmidgall2024agentclinic} & \cmark & \pmark & \pmark & \xmark & \pmark & \cmark & \cmark & \xmark & \pmark \\
        MedChain~\citep{liu2024medchain}        & \cmark & \cmark & \cmark & \xmark & \xmark & \pmark & \cmark & \xmark & \pmark \\
        MedAgentBench~\citep{jiang2025medagentbench} & \cmark & \cmark & \pmark & \xmark & \xmark & \xmark & \xmark & \pmark & \xmark \\
        MedAgentGym~\citep{shi2025medagentgym}  & \cmark & \cmark & \pmark & \xmark & \xmark & \xmark & \xmark & \pmark & \xmark \\
        HealthAdminBench~\citep{bedi2026healthadminbench} & \cmark & \xmark & \cmark & \pmark & \cmark & \xmark & \pmark & \cmark & \cmark \\
        \midrule
        SWE-Bench~\citep{jimenez2024swebenchlanguagemodelsresolve}   & \xmark & \cmark & \cmark & \xmark & \xmark & \xmark & \xmark & \pmark & \xmark \\
        WebArena~\citep{zhou2024webarena}       & \xmark & \xmark & \cmark & \xmark & \xmark & \xmark & \xmark & \pmark & \xmark \\
        OSWorld~\citep{xie2024osworld}          & \xmark & \xmark & \cmark & \xmark & \xmark & \xmark & \xmark & \cmark & \xmark \\
        WorkArena~\citep{drouin2024workarena}   & \xmark & \xmark & \cmark & \pmark & \xmark & \xmark & \xmark & \cmark & \xmark \\
        AppWorld~\citep{trivedi2024appworld}    & \xmark & \cmark & \cmark & \xmark & \pmark & \pmark & \pmark & \cmark & \xmark \\
        Terminal-Bench~\citep{merrill2026terminal} & \xmark & \cmark & \cmark & \xmark & \xmark & \xmark & \xmark & \cmark & \xmark \\
        Toolathlon~\citep{li2025tool}           & \xmark & \cmark & \cmark & \xmark & \xmark & \xmark & \xmark & \cmark & \xmark \\
        SkillsBench~\citep{li2026skillsbench}   & \pmark & \cmark & \cmark & \pmark & \xmark & \xmark & \xmark & \cmark & \xmark \\
        $\tau$/$\tau^2$-Bench~\citep{yao2024tau,barres2025tau2} & \xmark & \cmark & \cmark & \pmark & \pmark & \cmark & \cmark & \cmark & \xmark \\
        TheAgentCompany~\citep{xu2024agentcompany} & \xmark & \pmark & \cmark & \pmark & \pmark & \cmark & \pmark & \cmark & \pmark \\
        \midrule
        \rowcolor{gray!10}\textbf{\ourmethod{} (ours)} & \cmark & \cmark & \cmark & \cmark & \cmark & \cmark & \cmark & \cmark & \cmark \\
        \bottomrule
    \end{tabular}
    }
\end{table}

\paragraph{Long-horizon agent benchmarks.}
General-purpose benchmarks cover GUI control~\citep{zhou2024webarena,xie2024osworld,drouin2024workarena}, long-horizon code~\citep{jimenez2024swebenchlanguagemodelsresolve,merrill2026terminal}, and broad tool-use~\citep{trivedi2024appworld,li2025tool,li2026skillsbench}, but rarely model multi-actor workflows. $\tau$/$\tau^2$-Bench~\citep{yao2024tau} and TheAgentCompany~\citep{xu2024agentcompany} are closest in interaction structure, pairing agents with simulated stakeholders under policy constraints; neither targets healthcare or the long-horizon, policy-dense,  information asymmetry that defines prior authorization. 

See cell-by-cell details of Table~\ref{tab:review} in \Cref{app:benchmark-comparison}.

\section{\ourmethod: high-fidelity healthcare environment and benchmark}

\ourmethod~evaluates AI agents on \textbf{\underline{c}}linical \textbf{\underline{h}}ealthcare workflows \textbf{\underline{i}}n-situ (\texorpdfstring{\raisebox{0.25ex}{\scalebox{1.1}{$\upchi$}}})), automating prior-authorization (PA), utilization-management (UM), and care-management (CM) tasks for U.S. providers and payers. It spans three long-horizon domains, each requiring grounded navigation of a large policy library: \textbf{(1) Provider PA submission}—verify coverage, gather evidence, submit the packet, and work the response (RFIs, peer-to-peers, appeals) to terminal status; \textbf{(2) Payer UM review}—intake the request, check plan policy, escalate through nurse and physician reviewers, and issue a determination; \textbf{(3) RN care management}—review the chart, contact the patient, administer assessments, and author a care plan.

\begin{figure}[ht]
    \centering
    \includegraphics[width=\linewidth]{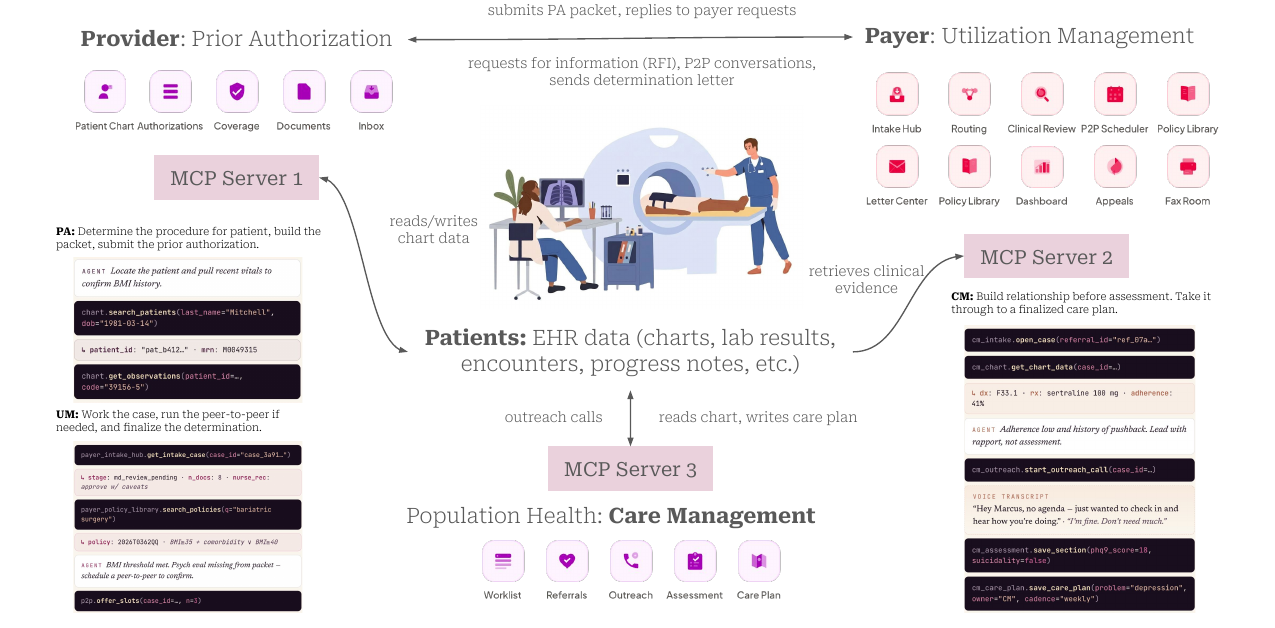}
    \caption{\ourworld~engine: simulated worlds for \underline{C}linical \underline{H}ealthcare \underline{I}n-situ workflows.}
    \label{fig:chi-world}
    \vspace{-10pt}
\end{figure}

\subsection{\ourworld~engine: simulated worlds for \underline{C}linical \underline{H}ealthcare \underline{I}n-situ workflows}
Healthcare workflows involve four stakeholders: \textit{patients}, \textit{clinicians (provider)}, \textit{payers}, and \textit{care management} entities, and a faithful benchmark must represent each and their interactions. \textbf{\ourworld~Engine} (\Cref{fig:chi-world}) is a local, high-fidelity simulator of 20 day-to-day healthcare apps, operable via \totalnumroutes~REST APIs and \totalnummcptools~MCP tools across 3 MCP servers, populated with \textbf{\totalnumactivities} chart activities for \textbf{\totalnumpatients} simulated patients and \textbf{\totalnumworkers} healthcare workers. Agents operate the apps autonomously through MCP servers, the local database, and the file system.

\subsubsection{Realistic healthcare software environments}

We implement the apps\footnote{Using FastAPI, SQLite, SQLModel, and MCP over streamable HTTP.} across three domains: \textit{provider PA}, \textit{payer UM}, and \textit{care management}. Built in \engineloc~lines of Python, the simulator captures features absent from general-purpose benchmarks: case state machines with \totalnumstates~statuses and explicit legal transitions; reviewer-independence constraints across nurse, medical-director, and peer-to-peer review; channel-specific submission semantics; and document authorship, signing, and FHIR-grade encounter linkage. Actions trigger consistent cross-app effects atomically: a provider-side submission spawns a payer intake record, advances the event log, and may produce routing assignments, pend notifications, and outbound letters.

\begin{figure}[h]
    \centering
    \includegraphics[width=\linewidth]{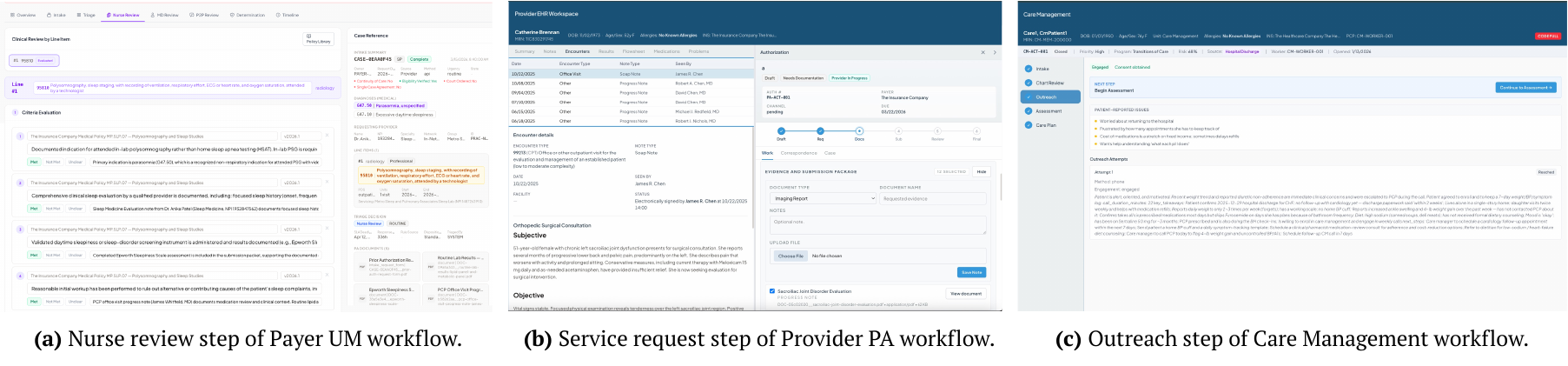}
    \caption{Healthcare apps across three task domains. \textit{(a) Payer -- Utilization Management} (10 apps); shown: nurse clinical review. \textit{(b) Provider -- Prior Authorization} (5 apps); shown: service-request step. \textit{(c) Population Healthcare -- Care Management} (5 apps); shown: patient outreach.}
    \label{fig:software-showcase}
\end{figure}

We expose \totalnummcptools~of the \totalnumroutes~backend APIs as MCP tools, manually selected to mirror UI operations available to human users. See appendix for the MCP server and tool details.

\subsubsection{Encoding healthcare workflows with Managed-Care Operations Handbook Skill}
\label{sec:skill}

We complement MCP servers with \textit{Agent Skills}~\cite{zhang2025equipping,jones2025code} to teach agents the specialized healthcare workflows. To simulate realistically how a healthcare worker handles a case, skills must encode the entire operation workflows, external software usage patterns, and the medical and insurance policies that govern each decision (e.g., payer medical-policy criteria, insurance coverage and eligibility, etc.).

\begin{figure}[ht]
\centering
\resizebox{0.9\textwidth}{!}{%
\begin{tikzpicture}[
    font=\ttfamily\scriptsize,
    skillbox/.style={
        draw=blue!55, line width=0.5pt, rounded corners=4pt,
        fill=blue!6, inner sep=5pt, align=left, text=blue!65!black,
        text width=5.4cm,
    },
    rootbox/.style={
        draw=blue!70, line width=0.7pt, rounded corners=4pt,
        fill=blue!12, inner sep=5pt, align=left, text=blue!65!black,
    },
    arrow/.style={-{Latex[length=2mm]}, draw=ForestGreen!70!black, line width=0.6pt},
    xlink/.style={dashed, dash pattern=on 2pt off 2pt, draw=black!40, line width=0.4pt},
    node distance=4mm and 4mm,
]
\node[skillbox] (pa) {%
    {\bfseries provider-pa/} \textcolor{black!55}{\rmfamily\itshape PA specialist}\\[-1pt]
    - SKILL.md\enspace - 01-workflow.md\\[-1pt]
    - pa-requirements/ \textcolor{black!55}{\rmfamily\itshape (CPT/HCPCS)}\\[-1pt]
    - forms/ \textcolor{black!55}{\rmfamily\itshape (by coverage)}\enspace - examples/
};
\node[skillbox, right=4mm of pa] (um) {%
    {\bfseries payer-um/} \textcolor{black!55}{\rmfamily\itshape UM reviewer}\\[-1pt]
    - SKILL.md\enspace - 01-workflow.md\\[-1pt]
    - operational/ \textcolor{black!55}{\rmfamily\itshape (5 chapters)}\\[-1pt]
    - examples/ \textcolor{black!55}{\rmfamily\itshape (letters)}
};
\node[skillbox, right=4mm of um] (cm) {%
    {\bfseries care-manager/} \textcolor{black!55}{\rmfamily\itshape RN care manager}\\[-1pt]
    - SKILL.md\enspace - 01-workflow.md\\[-1pt]
    - operational/ \textcolor{black!55}{\rmfamily\itshape (6 chapters)}
};
\coordinate (rail-l) at ([yshift=-3mm]pa.south west);
\coordinate (rail-r) at ([yshift=-3mm]cm.south east);
\node[skillbox, text width=8.5cm, below=3mm of rail-l, anchor=north west] (med) {%
    {\bfseries medical-library/} \textcolor{black!55}{\rmfamily\itshape shared appendix}\\[-1pt]
    - 1{,}000+ medical-policy documents\enspace - drug auth.\ criteria\enspace - clinical guidelines
};
\node[skillbox, text width=8.5cm, right=4mm of med] (plat) {%
    {\bfseries platform/} \textcolor{black!55}{\rmfamily\itshape shared appendix}\\[-1pt]
    - object semantics\enspace - surface ownership\enspace - workspace boundaries \textcolor{black!55}{\rmfamily\itshape (per role)}
};
\node[rootbox, text width=17.7cm, above=5mm of um.north, anchor=south] (root) {%
    {\bfseries managed-care-operations-handbook/}\enspace
    - SKILL.md \textcolor{black!55}{\rmfamily\itshape (top-level index $\to$ routes by role)}\enspace
    - references/
};
\foreach \tgt in {pa, um, cm} {
    \draw[arrow] (root.south) -- ++(0,-0.15) -| (\tgt.north);
}
\draw[xlink] (rail-l) -- (rail-r);
\foreach \src in {pa, um, cm} {
    \draw[xlink] (\src.south) -- (\src.south |- rail-l);
}
\draw[xlink] (med.north  |- rail-l) -- (med.north);
\draw[xlink] (plat.north |- rail-l) -- (plat.north);
\end{tikzpicture}%
}
\caption{\textit{Managed-Care Operations Handbook Skill} is organized as a progressive-disclosure manual. The top-level \texttt{SKILL.md} acts as a table of contents that routes the agent to one of three role sub-skills (\texttt{provider-pa}, \texttt{payer-um}, \texttt{care-manager}); the two shared \texttt{medical-library} (clinical lookup) and \texttt{platform} (role-specific tutorials) are reachable from any sub-skill via the dashed access bus.}
\label{fig:handbook-tree}
\vspace{-10pt}
\end{figure}
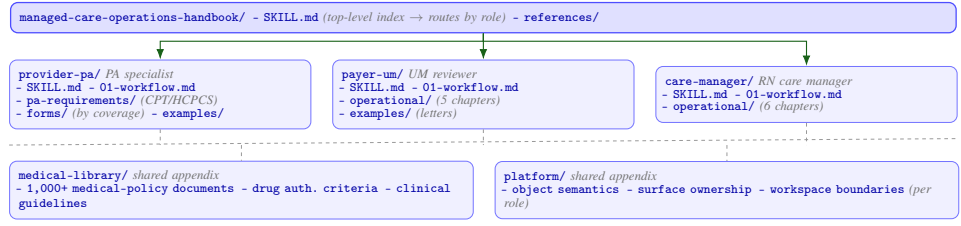
In this paper, we propose a core skill, the \textit{Managed-Care Operations Handbook} with \textbf{{1,290}} markdown documents in a skill/sub-skill structure, developed with clinicians and operations leaders at Johns Hopkins Medicine to ensure clinical fidelity and alignment with real-world workflows. We treat skill authoring as writing the onboarding guide for a new hire. As shown in \Cref{fig:handbook-tree}, we organize the skill as a wiki manual, where a top-level skill routes the agent to one of three role sub-skills (PA specialist, UM reviewer, care manager), each opening with a workflow chapter before diving into role-specific chapters and templates. Two appendices: a \textit{medical library} of policies, drug criteria, and guidelines curated and validated with subject-matter experts, and \textit{platform tutorials} on how to use MCP for specialized workflows. To our knowledge, although skill context can be in theory unbounded, the largest skills published to date are a handful of files; this is the \textit{first time} agent with skills have been evaluated at the scale of a real healthcare operational workflow library. The handbook details, and provenance and licensing information are in \Cref{app:handbook-detail}.

\subsection{The \ourmethod~construction}
\label{sec:construction}

\subsubsection{Task definition}
\label{sec:task-formulation}

A \ourmethod~task is a quadruple: instructions, the containerized \ourworld~environment, role-scoped tool surfaces, and a two-layer verifier---formalized as a hierarchical POMDP~\citep{kaelbling1998planning} $\mathcal{M}=(\mathcal{S},\mathcal{A},\mathcal{O},P,Z,R,\rho_0;\mathcal{H})$, where the latent state $\mathcal{S}$ spans patient charts, payer/provider records, workflow status, communications, artifacts, and event history; $\mathcal{A}$ comprises role-scoped MCP and default-agent tool actions; $\mathcal{O}$ comprises the role-scoped observations returned through MCP outputs, messages, policy passages, and shared-workspace files; $P$ and $Z$ are the transition and observation kernels induced by the environment and its tools; $R$ is the verifier-induced reward; and $\rho_0$ is the distribution over initial task states. The hierarchy $\mathcal{H}:=(\mathcal{G},\nu,\mathcal{W})$ uses role-agent specifications $\mathcal{G}:=\{(G_i,u_i,\mathcal{K}_i)\}_{i=1}^{N}$, where $G_i$ is a role agent, $u_i$ its instruction, and $\mathcal{K}_i$ its available skill set; $\nu$ defines the handoff order and $\mathcal{W}$ the shared workspace. Each $\mathcal{K}_i$ is a set of options~\citep{sutton1999between}, i.e. temporally extended procedures (e.g., \emph{nurse criterion review}: policy retrieval $\rightarrow$ chart read $\rightarrow$ structured-payload write). Instructions specify role, case, workspace, and rules; procedural detail must be recovered from the handbook. Handoffs are irreversible: outgoing commits to $\mathcal{W}$ become incoming input, and the accumulating state and event log calculate reward~(\Cref{sec:verification}).

\begin{figure}[ht]
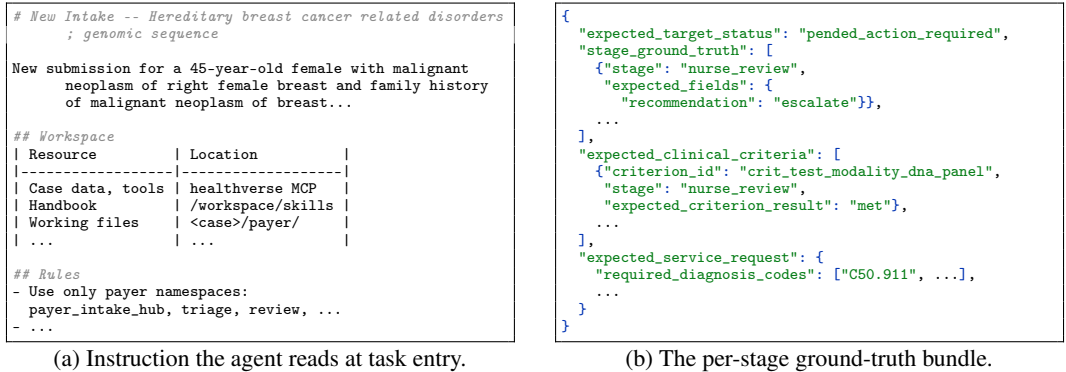

\centering
\captionsetup[sub]{skip=2pt}
\begin{subfigure}[b]{0.48\textwidth}
\begin{lstlisting}[language=mdlite]
# New Intake -- Hereditary breast cancer related disorders; genomic sequence

New submission for a 45-year-old female with malignant neoplasm of right female breast and family history of malignant neoplasm of breast...

## Workspace
| Resource         | Location          |
|------------------|-------------------|
| Case data, tools | healthverse MCP   |
| Handbook         | /workspace/skills |
| Working files    | <case>/payer/     |
| ...              | ...               |

## Rules
- Use only payer namespaces:
  payer_intake_hub, triage, review, ...
- ...
\end{lstlisting}
\caption{Instruction the agent reads at task entry.}
\label{fig:instruction-example-a}
\end{subfigure}\hfill
\begin{subfigure}[b]{0.48\textwidth}
\begin{lstlisting}[language=jsonlite]
{
  "expected_target_status": "pended_action_required",
  "stage_ground_truth": [
    {"stage": "nurse_review",
     "expected_fields": {
       "recommendation": "escalate"}},
    ...
  ],
  "expected_clinical_criteria": [
    {"criterion_id": "crit_test_modality_dna_panel",
     "stage": "nurse_review",
     "expected_criterion_result": "met"},
    ...
  ],
  "expected_service_request": {
    "required_diagnosis_codes": ["C50.911", ...],
    ...
  }
}
\end{lstlisting}
\caption{The per-stage ground-truth bundle.}
\label{fig:instruction-example-b}
\end{subfigure}
\caption{Example of a Payer UM task for hereditary breast-cancer genomic
sequencing.}
\label{fig:instruction-example}
\end{figure}

\subsubsection{Task construction and composition}

Each task annotation consists of sampling an initial state $s_0 \sim \rho_0$, a role assignment over
$\mathcal{G}$, and a ground-truth trajectory clicked through the \ourworld~UI.

\textbf{Step 1 -- Case generation.} The pipeline first samples a terminal world state of a case, then uses Claude Opus 4.7~+~structured JSON sampling, conditioned on the relevant system state graph and the matching section of the \emph{Managed-Care Operations Handbook} to emit the upstream artifacts, including chart specifications, submission packets or personas, and per-stage rubric prompts, each of which is anchored to an explicit policy or state graph citation. \textbf{Step 2 -- Human walkthrough.}
An annotator works on each case candidate end-to-end on the live
\ourworld~UI with the handbook. The recorded trajectories, db states, workspace commits, and role handoffs become the ground truth.
\label{sec:task-construction}
\begin{wrapfigure}{r}{0.475\textwidth}
  \vspace{-\intextsep}
  \begin{center}
    \includegraphics[width=0.325\textwidth]{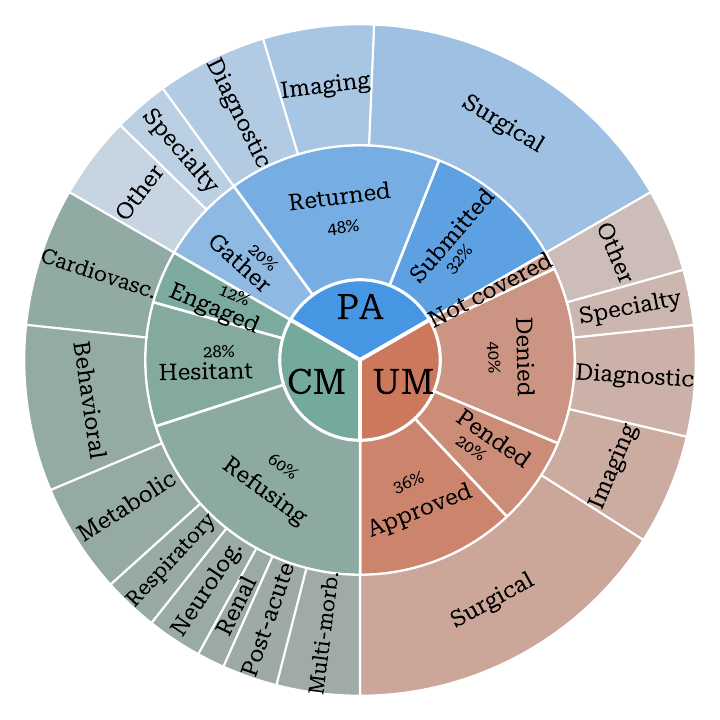}
  \end{center}
  \caption{Task breakdown. Inner: Domain; Middle: PA/UM terminal state, CM patient persona; Outer: clinical/service category.}
  \label{fig:task-composition}
  \vspace{-\intextsep}
\end{wrapfigure}
\textbf{Step 3 -- Multi-reviewer review.}
Each trajectory is reviewed by at least 1~practicing
healthcare worker and 5 authors for clinical
precision, and must clear a residual-PHI scan and a clinical-realism check before admission. The detailed human validation protocols are described in \Cref{app:task-construction-detail}.

The annotation pipeline has produced 523 tasks, each assigned a difficulty band from tool-call length, decision-tree depth. Candidates are retained only when every expected action resolves to a cited policy section, and the chart and rubric mutually entail without leaking the chosen path. {A deterministic dry-run must also execute a reference solution to a single well-formed terminal state.} {These checks leave 172 candidates for human review.} We filter down to 75 representative, long-horizon tasks for quality and diversity, where the human on average needs 21 steps, and at most 40 steps to finish. The task categories are depicted in \Cref{fig:task-composition}.

\subsubsection{Reward}
\label{sec:verification}

\begin{figure}[ht]
    \centering
    \includegraphics[width=\linewidth]{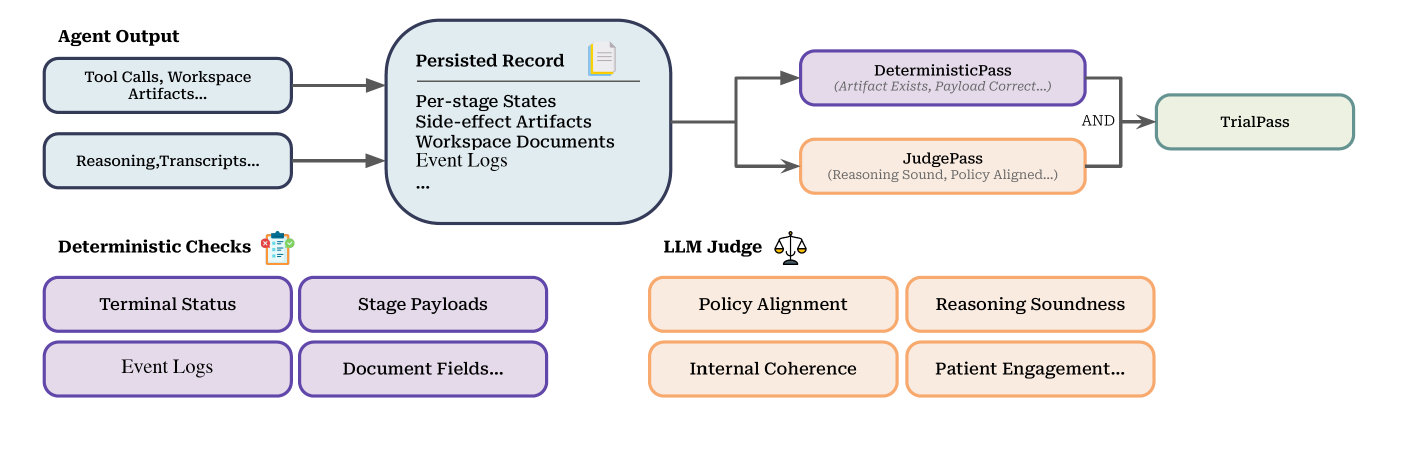}
    \caption{\textbf{Verification pipeline.} Each trial emits a persisted record to \emph{deterministic contract} verifier and \emph{rubric-based LLM judge} under strict-majority vote. A trial passes only when both layers pass.}
    \label{fig:verification}
\end{figure}

The verifier  (\Cref{fig:verification}) scores each trial off the record the simulator itself persisted: world store, event log, and multi-turn transcripts, combining a \textbf{deterministic contract} with a \textbf{rubric LLM judge} into a binary reward $R=\mathrm{DeterministicPass}\land\mathrm{JudgePass}$, with a fractional scorecard for diagnostics.

\section{Experiments}

\subsection{Experiment setup}

We evaluate 30 agent harness/model configurations across two stacks: a \emph{proprietary stack} pairing each frontier lab's first-party CLI (\textit{Claude Code}~\cite{anthropic2025claudecode}, \textit{OpenAI Codex}~\cite{openai2025codexcli}, \textit{Gemini CLI}~\cite{google2025geminicli}) with that lab's closed-weight models~\cite{anthropic2026claude4,openai2026gpt5,google2026gemini3}, plus an \emph{open-source stack} sweeping four agent frameworks (\textit{OpenClaw}~\cite{openclaw2025}, \textit{Hermes}~\cite{hermes2026}, \textit{OpenAI Agents SDK}~\cite{openai2025agentssdk} (\emph{OAI Agents}), and \textit{DeepAgents}~\cite{langchain2025deepagents}) over five OpenRouter-served open-weight models~\cite{deepseek2026v4,zhipu2026glm5,moonshot2025kimik2,alibaba2025qwen3,xai2025grok4}, plus an additional OpenClaw~+~Claude~Opus~4.7 reference cell. For each task we run $3$ independent trials and report pass@$1$~\cite{chen2021evaluating}, pass@$3$, and pass\textasciicircum$3$~\cite{yao2024tau}. The evaluation protocol is shown in \Cref{fig:verification}. Detail configurations like sandbox, judge, and runtime are deferred to \Cref{app:experiment-detail}.

\subsection{\ourmethod{} results}

\Cref{tab:main} summarizes benchmark performance across agent harnesses and frontier models.

\begin{table*}[ht]
    \centering
    \normalsize
    \setlength{\tabcolsep}{2pt}
    \renewcommand{\arraystretch}{1.1}
    \caption{\ourmethod{} results across agent harnesses and frontier models. Per-column maxima are bolded. The three \textbf{Overall} columns show task-level bootstrap 95\% CIs in $value_{-\mathrm{lo}}^{+\mathrm{hi}}$ form; per-domain pass cells show mean only and the two \emph{Efficiency} columns are averaged over all $225$ trials per row.}
    \label{tab:main}
    \begin{adjustbox}{max width=\textwidth}
    \begin{tabular}{@{}llcccccccccccccc@{}}
    \toprule
    \multirow{2}{*}{\textbf{Agent Harness}} & \multirow{2}{*}{\textbf{Model}} & \multicolumn{3}{c}{\textbf{Overall}} & \multicolumn{3}{c}{\ico{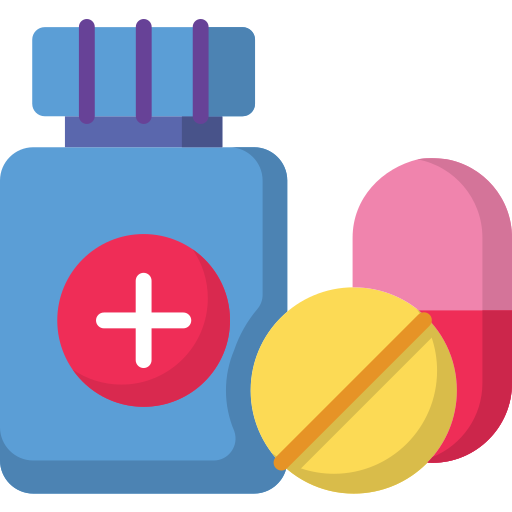}\textbf{\makecell{Prior Authorization}}} & \multicolumn{3}{c}{\ico{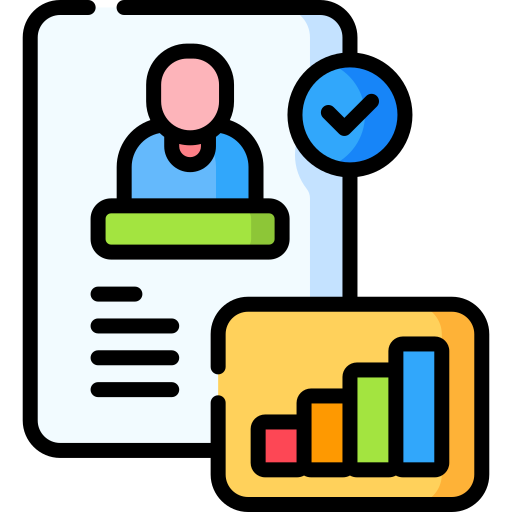}\textbf{\makecell{Utilization Management}}} & \multicolumn{3}{c}{\ico{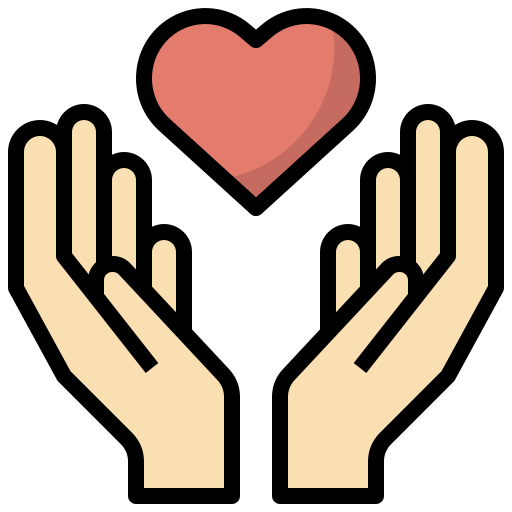}\textbf{\makecell{Care Management}}} & \multicolumn{2}{c}{\textbf{Efficiency}} \\
    \cmidrule(lr){3-5} \cmidrule(lr){6-8} \cmidrule(lr){9-11} \cmidrule(lr){12-14} \cmidrule(lr){15-16}
    &  & \textbf{pass@$1$} & \textbf{pass@3} & \textbf{pass\^{}3} & \textbf{pass@$1$} & \textbf{pass@3} & \textbf{pass\^{}3} & \textbf{pass@$1$} & \textbf{pass@3} & \textbf{pass\^{}3} & \textbf{pass@$1$} & \textbf{pass@3} & \textbf{pass\^{}3} & \textbf{Steps} & \textbf{Cost (\$)} \\
    \midrule
    \ico{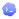}Codex         & \ico{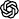}GPT-5.5            & \heatci{20.9}{7.6}{8.4} & \heatci{30.7}{10.7}{10.7} & \heatci{9.3}{5.3}{8.0} & \heatbf{29.3} & \heatbf{40.0} & \heat{16.0} & \heat{32.0} & \heat{48.0} & \heat{12.0} & \heat{1.3}  & \heat{4.0}  & \heat{0.0}  & \heatsteps{54} & \heatcost{1.29}{\$1.29} \\
    \ico{codex}Codex         & \ico{openai}GPT-5.4            & \heatci{16.0}{6.7}{7.1} & \heatci{25.3}{9.3}{9.3} & \heatci{8.0}{5.3}{6.7} & \heat{24.0} & \heat{32.0} & \heat{16.0} & \heat{17.3} & \heat{24.0} & \heat{8.0}  & \heat{6.7}  & \heat{20.0} & \heat{0.0}  & \heatsteps{58} & \heatcost{1.30}{\$1.30} \\
    \ico{codex}Codex         & \ico{openai}GPT-5.4 Mini       & \heatci{8.4}{4.0}{4.4}  & \heatci{20.0}{9.3}{10.7} & \heatci{0.0}{0.0}{0.0} & \heat{10.7} & \heat{24.0} & \heat{0.0}  & \heat{13.3} & \heat{32.0} & \heat{0.0}  & \heat{1.3}  & \heat{4.0}  & \heat{0.0}  & \heatsteps{58} & \heatcost{0.27}{\$0.27} \\
    \ico{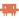}Claude Code   & \ico{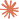}Claude Opus 4.7    & \heatci{24.4}{8.0}{8.4} & \heatcibf{41.3}{12.0}{12.0} & \heatci{10.7}{6.7}{8.0} & \heat{24.0} & \heat{32.0} & \heat{16.0} & \heat{17.3} & \heat{28.0} & \heat{8.0}  & \heatbf{32.0} & \heatbf{64.0} & \heatbf{8.0}  & \heatsteps{68} & \heatcost{9.91}{\$9.91} \\
    \ico{claudecode}Claude Code   & \ico{claude-color}Claude Opus 4.6    & \heatcibf{28.0}{8.4}{8.9} & \heatci{38.7}{10.7}{10.7} & \heatcibf{18.7}{8.0}{9.3} & \heat{18.7} & \heat{24.0} & \heat{12.0} & \heatbf{41.3} & \heat{44.0} & \heatbf{40.0} & \heat{24.0} & \heat{48.0} & \heat{4.0}  & \heatsteps{76} & \heatcost{6.47}{\$6.47} \\
    \ico{claudecode}Claude Code   & \ico{claude-color}Claude Sonnet 4.6  & \heatci{26.2}{8.0}{7.6} & \heatcibf{41.3}{10.7}{10.7} & \heatci{12.0}{6.7}{8.0} & \heat{24.0} & \heat{28.0} & \heatbf{20.0} & \heat{34.7} & \heatbf{52.0} & \heat{16.0} & \heat{20.0} & \heat{44.0} & \heat{0.0}  & \heatsteps{82} & \heatcost{1.30}{\$1.30} \\
    \ico{claudecode}Claude Code   & \ico{claude-color}Claude Haiku 4.5   & \heatci{6.2}{4.0}{5.3}  & \heatci{10.7}{6.7}{8.0} & \heatci{2.7}{2.7}{4.0} & \heat{0.0}  & \heat{0.0}  & \heat{0.0}  & \heat{14.7} & \heat{24.0} & \heat{8.0}  & \heat{4.0}  & \heat{8.0}  & \heat{0.0}  & \heatsteps{41} & \heatcost{0.16}{\$0.16} \\
    \ico{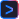}Gemini CLI    & \ico{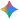}Gemini 3.1 Pro     & \heatci{7.1}{4.0}{4.9}  & \heatci{13.3}{6.7}{8.0} & \heatci{1.3}{1.3}{2.7} & \heat{14.7} & \heat{24.0} & \heat{4.0}  & \heat{6.7}  & \heat{16.0} & \heat{0.0}  & \heat{0.0}  & \heat{0.0}  & \heat{0.0}  & \heatsteps{82} & \heatcost{2.11}{\$2.11} \\
    \ico{geminicli}Gemini CLI    & \ico{gemini-color}Gemini 3 Flash     & \heatci{12.4}{6.2}{7.1} & \heatci{17.3}{8.0}{9.3} & \heatci{8.0}{5.3}{6.7} & \heat{18.7} & \heat{28.0} & \heat{8.0}  & \heat{18.7} & \heat{24.0} & \heat{16.0} & \heat{0.0}  & \heat{0.0}  & \heat{0.0}  & \heatsteps{142} & \heatcost{0.33}{\$0.33} \\
    \midrule
    \ico{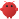}OpenClaw    & \ico{claude-color}Claude Opus 4.7   & \heatci{17.3}{5.8}{6.2} & \heatci{37.3}{12.0}{10.7} & \heatci{4.0}{4.0}{5.3} & \heat{18.7} & \heat{28.0} & \heat{8.0}  & \heat{13.3} & \heat{32.0} & \heat{4.0}  & \heat{20.0} & \heat{52.0} & \heat{0.0}  & \heatsteps{41} & \heatcost{11.48}{\$11.48}\\
    \ico{openclaw-color}OpenClaw    & \ico{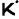}Kimi K2.6                 & \heatci{10.2}{4.9}{5.3} & \heatci{18.7}{8.0}{9.3} & \heatci{2.7}{2.7}{4.0} & \heat{12.0} & \heat{20.0} & \heat{4.0}  & \heat{18.7} & \heat{36.0} & \heat{4.0}  & \heat{0.0}  & \heat{0.0}  & \heat{0.0}  & \heatsteps{72} & \heatcost{0.91}{\$0.91} \\
    \ico{openclaw-color}OpenClaw    & \ico{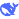}DeepSeek V4 Pro & \heatci{11.1}{4.9}{5.3} & \heatci{24.0}{9.3}{9.3} & \heatci{1.3}{1.3}{2.7} & \heat{14.7} & \heat{28.0} & \heat{4.0}  & \heat{12.0} & \heat{28.0} & \heat{0.0}  & \heat{6.7}  & \heat{16.0} & \heat{0.0}  & \heatsteps{42} & \heatcost{0.53}{\$0.53} \\
    \ico{openclaw-color}OpenClaw    & \ico{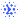}GLM-5.1            & \heatci{16.9}{6.7}{7.1} & \heatci{30.7}{10.7}{10.7} & \heatci{6.7}{5.3}{6.7} & \heat{13.3} & \heat{24.0} & \heat{4.0}  & \heat{26.7} & \heat{36.0} & \heat{16.0} & \heat{10.7} & \heat{32.0} & \heat{0.0}  & \heatsteps{116} & \heatcost{0.96}{\$0.96} \\
    \ico{openclaw-color}OpenClaw    & \ico{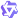}Qwen 3.6 Max        & \heatci{4.9}{3.1}{4.0}  & \heatci{10.7}{6.7}{8.0} & \heatci{0.0}{0.0}{0.0} & \heat{10.7} & \heat{24.0} & \heat{0.0}  & \heat{4.0}  & \heat{8.0}  & \heat{0.0}  & \heat{0.0}  & \heat{0.0}  & \heat{0.0}  & \heatsteps{79} & \heatcost{2.80}{\$2.80} \\
    \ico{openclaw-color}OpenClaw    & \ico{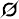}Grok 4.3                   & \heatci{0.4}{0.4}{0.9}  & \heatci{1.3}{1.3}{2.7} & \heatci{0.0}{0.0}{0.0} & \heat{1.3}  & \heat{4.0}  & \heat{0.0}  & \heat{0.0}  & \heat{0.0}  & \heat{0.0}  & \heat{0.0}  & \heat{0.0}  & \heat{0.0}  & \heatsteps{65} & \heatcost{2.66}{\$2.66} \\
    \ico{openai}OAI Agents  & \ico{kimi}Kimi K2.6                 & \heatci{15.1}{6.2}{7.1} & \heatci{22.7}{9.3}{9.3} & \heatci{8.0}{5.3}{6.7} & \heat{17.3} & \heat{28.0} & \heat{12.0} & \heat{25.3} & \heat{36.0} & \heat{12.0} & \heat{2.7}  & \heat{4.0}  & \heat{0.0}  & \heatsteps{60} & \heatcost{0.43}{\$0.43} \\
    \ico{openai}OAI Agents   & \ico{deepseek-color}DeepSeek V4 Pro & \heatci{14.2}{6.2}{7.1} & \heatci{22.7}{9.3}{9.3} & \heatci{9.3}{5.3}{6.7} & \heat{10.7} & \heat{16.0} & \heat{8.0}  & \heat{28.0} & \heat{40.0} & \heat{20.0} & \heat{4.0}  & \heat{12.0} & \heat{0.0}  & \heatsteps{52} & \heatcost{0.25}{\$0.25} \\
    \ico{openai}OAI Agents   & \ico{zhipu-color}GLM-5.1            & \heatci{18.7}{8.0}{8.4} & \heatci{26.7}{9.3}{10.7} & \heatci{12.0}{6.7}{8.0} & \heat{18.7} & \heat{24.0} & \heat{12.0} & \heat{33.3} & \heat{44.0} & \heat{24.0} & \heat{4.0}  & \heat{12.0} & \heat{0.0}  & \heatsteps{58} & \heatcost{0.27}{\$0.27} \\
    \ico{openai}OAI Agents   & \ico{qwen-color}Qwen 3.6 Max        & \heatci{15.6}{6.7}{8.0} & \heatci{22.7}{9.3}{10.7} & \heatci{9.3}{5.3}{6.7} & \heat{16.0} & \heat{20.0} & \heat{12.0} & \heat{26.7} & \heat{36.0} & \heat{16.0} & \heat{4.0}  & \heat{12.0} & \heat{0.0}  & \heatsteps{48} & \heatcost{0.58}{\$0.58} \\
    \ico{openai}OAI Agents   & \ico{grok}Grok 4.3                   & \heatci{5.8}{3.6}{4.4}  & \heatci{10.7}{6.7}{8.0} & \heatci{1.3}{1.3}{2.7} & \heat{0.0}  & \heat{0.0}  & \heat{0.0}  & \heat{16.0} & \heat{28.0} & \heat{4.0}  & \heat{1.3}  & \heat{4.0}  & \heat{0.0}  & \heatsteps{32} & \heatcost{1.54}{\$1.54} \\
    \ico{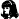}Hermes        & \ico{kimi}Kimi K2.6                 & \heatci{15.6}{6.7}{7.6} & \heatci{24.0}{9.3}{10.7} & \heatci{6.7}{5.3}{6.7} & \heat{18.7} & \heat{24.0} & \heat{12.0} & \heat{21.3} & \heat{36.0} & \heat{8.0}  & \heat{6.7}  & \heat{12.0} & \heat{0.0}  & \heatsteps{31} & \heatcost{1.07}{\$1.07} \\
    \ico{nousresearch}Hermes        & \ico{deepseek-color}DeepSeek V4 Pro & \heatci{13.8}{6.2}{7.1} & \heatci{22.7}{9.3}{9.3} & \heatci{8.0}{5.3}{6.7} & \heat{8.0}  & \heat{16.0} & \heat{4.0}  & \heat{25.3} & \heat{32.0} & \heat{20.0} & \heat{8.0}  & \heat{20.0} & \heat{0.0}  & \heatsteps{26} & \heatcost{2.19}{\$2.19} \\
    \ico{nousresearch}Hermes        & \ico{zhipu-color}GLM-5.1            & \heatci{18.7}{7.1}{8.0} & \heatci{28.0}{9.3}{10.7} & \heatci{10.7}{6.7}{8.0} & \heat{10.7} & \heat{16.0} & \heat{8.0}  & \heat{34.7} & \heat{44.0} & \heat{24.0} & \heat{10.7} & \heat{24.0} & \heat{0.0}  & \heatsteps{30} & \heatcost{1.04}{\$1.04} \\
    \ico{nousresearch}Hermes        & \ico{qwen-color}Qwen 3.6 Max        & \heatci{16.4}{6.7}{6.7} & \heatci{28.0}{10.7}{10.7} & \heatci{5.3}{4.0}{5.3} & \heat{9.3}  & \heat{16.0} & \heat{4.0}  & \heat{26.7} & \heat{36.0} & \heat{12.0} & \heat{13.3} & \heat{32.0} & \heat{0.0}  & \heatsteps{29} & \heatcost{4.12}{\$4.12} \\
    \ico{nousresearch}Hermes        & \ico{grok}Grok 4.3                   & \heatci{4.4}{3.1}{4.4}  & \heatci{8.0}{5.3}{6.7} & \heatci{1.3}{1.3}{2.7} & \heat{0.0}  & \heat{0.0}  & \heat{0.0}  & \heat{13.3} & \heat{24.0} & \heat{4.0}  & \heat{0.0}  & \heat{0.0}  & \heat{0.0}  & \heatsteps{32} & \heatcost{1.05}{\$1.05} \\
    \ico{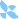}DeepAgents & \ico{kimi}Kimi K2.6                 & \heatci{3.1}{2.2}{3.1}  & \heatci{8.0}{5.3}{6.7} & \heatci{0.0}{0.0}{0.0} & \heat{8.0}  & \heat{20.0} & \heat{0.0}  & \heat{1.3}  & \heat{4.0}  & \heat{0.0}  & \heat{0.0}  & \heat{0.0}  & \heat{0.0}  & \heatsteps{39} & \heatcost{0.55}{\$0.55} \\
    \ico{langchain-color}DeepAgents & \ico{deepseek-color}DeepSeek V4 Pro & \heatci{10.7}{4.9}{5.8} & \heatci{18.7}{8.0}{9.3} & \heatci{2.7}{2.7}{4.0} & \heat{14.7} & \heat{24.0} & \heat{4.0}  & \heat{10.7} & \heat{20.0} & \heat{4.0}  & \heat{6.7}  & \heat{12.0} & \heat{0.0}  & \heatsteps{15} & \heatcost{0.21}{\$0.21} \\
    \ico{langchain-color}DeepAgents & \ico{zhipu-color}GLM-5.1            & \heatci{11.1}{5.8}{6.2} & \heatci{17.3}{8.0}{8.0} & \heatci{5.3}{4.0}{5.3} & \heat{17.3} & \heat{24.0} & \heat{12.0} & \heat{10.7} & \heat{16.0} & \heat{4.0}  & \heat{5.3}  & \heat{12.0} & \heat{0.0}  & \heatsteps{21} & \heatcost{0.26}{\$0.26} \\
    \ico{langchain-color}DeepAgents & \ico{qwen-color}Qwen 3.6 Max        & \heatci{9.3}{4.9}{5.8}  & \heatci{16.0}{8.0}{9.3} & \heatci{4.0}{4.0}{5.3} & \heat{12.0} & \heat{16.0} & \heat{8.0}  & \heat{10.7} & \heat{16.0} & \heat{4.0}  & \heat{5.3}  & \heat{16.0} & \heat{0.0}  & \heatsteps{18} & \heatcost{0.57}{\$0.57} \\
    \ico{langchain-color}DeepAgents & \ico{grok}Grok 4.3                   & \heatci{2.2}{1.8}{2.7}  & \heatci{5.3}{4.0}{5.3} & \heatci{0.0}{0.0}{0.0} & \heat{0.0}  & \heat{0.0}  & \heat{0.0}  & \heat{5.3}  & \heat{12.0} & \heat{0.0}  & \heat{1.3}  & \heat{4.0}  & \heat{0.0}  & \heatsteps{21} & \heatcost{1.43}{\$1.43} \\
    \bottomrule
    \end{tabular}
    \end{adjustbox}
\end{table*}

\begin{figure}[ht]
    \centering
    \begin{subfigure}[b]{0.48\textwidth}
        \centering
        \includegraphics[width=\linewidth]{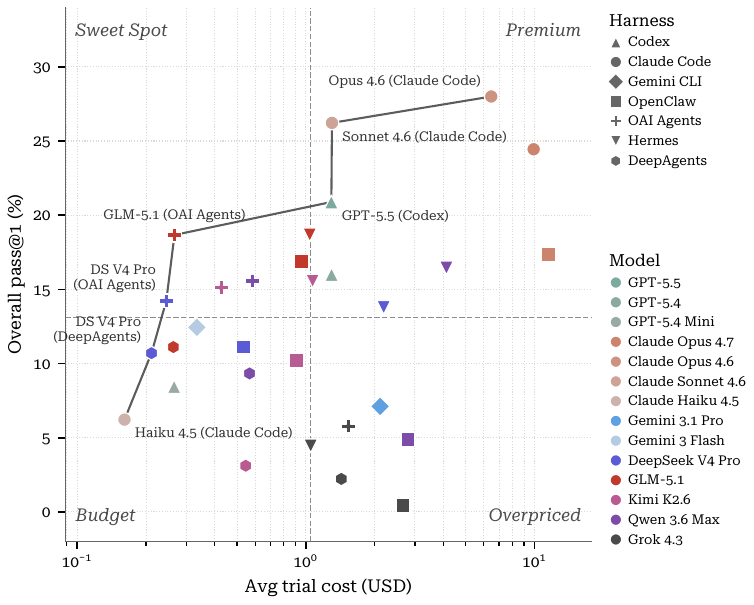}
        \caption{ROI quadrants.}
        \label{fig:roi-pareto}
    \end{subfigure}\hfill
    \begin{subfigure}[b]{0.48\textwidth}
        \centering
        \includegraphics[width=\linewidth]{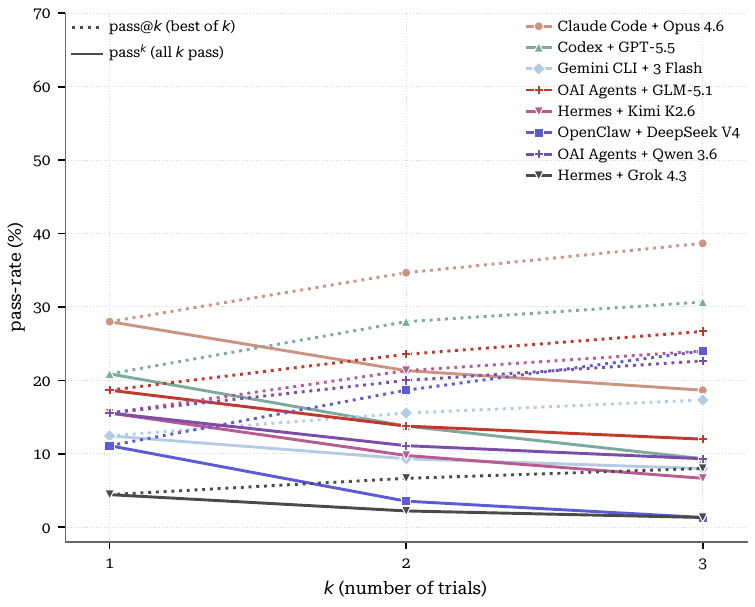}
        \caption{Reliability degradation.}
        \label{fig:passk-descent}
    \end{subfigure}
    \caption{(\subref*{fig:roi-pareto}) Each marker is one row of \Cref{tab:main}: $x$ = mean per-trial cost in USD (log scale), $y$ = Overall pass@$1$. Dashed cross-hairs at the median cost and median pass@$1$ split the plane into four quadrants (\textit{Sweet Spot}, \textit{Premium}, \textit{Budget}, \textit{Overpriced}); the Pareto-optimal frontier is connected with a dark line. (\subref*{fig:passk-descent}) pass@$k$ (dotted) and pass\^{}$k$ (solid) for $k\in\{1,2,3\}$ pooled across all 75 tasks.}
    \label{fig:cost-and-passk}
\end{figure}
\vspace{-10pt}
\paragraph{Performance, reliability and ROI.}
Claude Code paired with Claude Opus 4.6 tops Overall pass@$1$ at $28.0\%$, with Sonnet 4.6 ($26.2\%$), Opus 4.7 ($24.4\%$), and Codex~+~GPT-5.5 ($20.9\%$) close behind; the best domain-level rows are split across Opus 4.6 for UM ($41.3\%$), Opus 4.7 for CM ($32.0\%$), and Codex~+~GPT-5.5 for PA ($29.3\%$). Reliability further collapses on repeat trials (\Cref{fig:passk-descent}): pass\^{}3 sits well below pass@$1$ for the main cells (Opus 4.6 28.0$\to$18.7, GPT-5.5 20.9$\to$9.3, OAI Agents~+~GLM-5.1 18.7$\to$12.0, Hermes~+~Grok 4.3 4.4$\to$1.3), exposing run-to-run inconsistency that any production deployment would need to close.

The ROI quadrants in \Cref{fig:roi-pareto} separate absolute capability from cost-normalized value: high-performing configurations (e.g. Claude Code~+~Opus 4.6) sit in \emph{Premium}, while OAI Agents~+~GLM-5.1 stands out as a strong cost-normalized point, anchoring the \emph{Sweet Spot} and the low-cost end of the Pareto frontier. The \emph{Overpriced} quadrant collects all Grok 4.3 cells, OpenClaw~+~Qwen 3.6 Max, and Gemini 3.1 Pro~+~Gemini CLI; the \emph{Budget} quadrant contains low-cost rows whose savings come with below-median completion rates.

\subsection{\ourmethod{}-Arena: can prior authorization workflows be automated end-to-end?}
\begin{wraptable}[7]{r}{0.42\textwidth}
\vspace{-1.0em}
\centering
\small
\setlength{\tabcolsep}{4pt}
\renewcommand{\arraystretch}{1.05}
\caption{E2E two-agent PA vs.\ same-tasks single-agent baseline.}
\label{tab:e2e-results}
\begin{tabular}{@{}lc@{}}
\toprule
\textbf{Configuration} & \textbf{pass@$1$} \\
\midrule
PA provider-only (23 tasks) & \heat{30.4} \\
E2E two-agent                      & \heat{0.0} \\
\bottomrule
\end{tabular}
\end{wraptable}
The arena runs a provider agent and a payer agent, both running Codex~+~GPT-5.5 (our best PA configuration) as a two-player game end-to-end on 23 PA tasks.\footnote{Two tasks not applicable to the two-agent setting are excluded.} Each holds its own role-scoped MCPs and state, and they exchange information only through MCP tools. Each side is scored independently; a trial passes only when every check on both sides passes. Pass@$1$ collapses from \textbf{30.4\%}to \textbf{0\%} once the payer agent and cross-role checks join: $2$ tasks did not get submitted; $18$ did not finish MD decision, and $5$ failed the final judge. P2P tasks fail in both sides: $0$ P2P request on 5 P2P-required tasks appears and $2$ spontaneous P2Ps happen.

\subsection{\ourmethod{}-Marathon: can long-running agents stay on track across all 25 tasks?}
\ourmethod{}-Marathon stress-tests long-horizon capabilities by loading all 25 tasks of a domain into a shared \ourworld{}. The agent is instructed to finish all tasks, lists them via MCP tools and attempts in any order, in one agent run. Context compaction follows the harness's default setting. Each case is scored individually after the agent reports completion. We evaluate Claude~Code~+~Opus 4.7 and Codex~+~GPT-5.5. Pass@$1$ slumps for both configurations regardless of baseline (\Cref{tab:marathon-results}). On PA, neither agent submits a single authorization across any of the 25 queued cases, despite touching most cases via write-side tool calls. On UM and CM, agents reach a finalized determination or care plan on only 3-8 of 25 cases per session. Codex~+~GPT-5.5 reaches its context window and auto-compacts 4-6 times per PA session and 1-2 times on UM; Claude~Code~+~Opus 4.7, with a 1M-token context, never compacts yet completes a similar number of cases. However both agents fan out across the queue, save partial work, and fail to drive most cases to a terminal action. In the inspected trajectories, a median of 21 cases were open simultaneously, mixing PA/UM case states and CM patient conversations in the shared context.
\begin{table*}[ht]
    \centering
    \normalsize
    \setlength{\tabcolsep}{2pt}
    \renewcommand{\arraystretch}{1.1}
    \caption{\ourmethod{}-Marathon pass@$1$ vs.\ the per-task baseline. Marathon = all 25 tasks queued in a single agent session, pass@$1$ averaged over $3$ independent sessions; Per-task = isolated single-task trials from \Cref{tab:main}. $\Delta$ = Marathon $-$ Per-task (percentage points).}
    \label{tab:marathon-results}
    \begin{adjustbox}{max width=\textwidth}
    \begin{tabular}{@{}llccccccccc@{}}
    \toprule
    \multirow{2}{*}{\textbf{Agent Harness}} & \multirow{2}{*}{\textbf{Model}} & \multicolumn{3}{c}{\ico{PA}\textbf{Prior Authorization}} & \multicolumn{3}{c}{\ico{UM}\textbf{Utilization Management}} & \multicolumn{3}{c}{\ico{CM}\textbf{Care Management}} \\
    \cmidrule(lr){3-5} \cmidrule(lr){6-8} \cmidrule(lr){9-11}
     &  & \textbf{Marathon} & \textbf{Per-task} & \textbf{$\Delta$} & \textbf{Marathon} & \textbf{Per-task} & \textbf{$\Delta$} & \textbf{Marathon} & \textbf{Per-task} & \textbf{$\Delta$} \\
    \midrule
    \ico{codex}Codex       & \ico{openai}GPT-5.5         & \heat{8.0} & \heat{29.3} & $-21.3$ & \heat{2.7} & \heat{32.0} & $-29.3$ & \heat{0.0} & \heat{1.3}  & $-1.3$  \\
    \ico{claudecode}Claude Code & \ico{claude-color}Claude Opus 4.7 & \heat{8.0} & \heat{24.0} & $-16.0$ & \heat{1.3} & \heat{17.3} & $-16.0$ & \heat{2.7} & \heat{32.0} & $-29.3$ \\
    \bottomrule
    \end{tabular}
    \end{adjustbox}
\end{table*}
\vspace{-10pt}
\subsection{Effects of handbook skills components}
\begin{wrapfigure}[12]{r}{0.40\textwidth}
\centering
\includegraphics[width=0.40\textwidth]{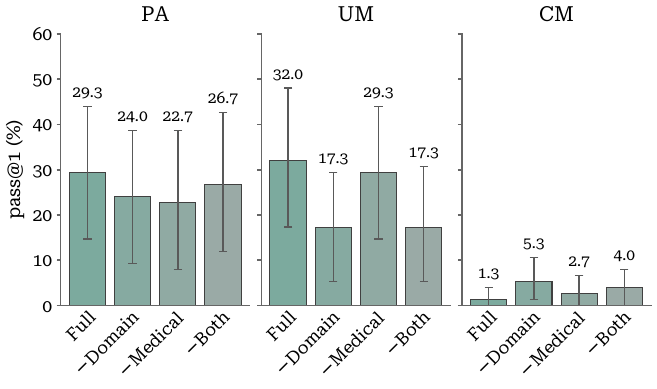}
\caption{Pass@$1$ under trimmed skills.}
\label{fig:skill-vs-no-skill-results}
\end{wrapfigure}

We trimmed the ${1{,}290}$-document \textit{Managed-Care Operations Handbook Skill} three ways ($-$\emph{Domain} drops the domain handbook, $-$\emph{Medical} drops the medical library, $-$\emph{Both} drops both), ran all tasks with Codex~+~GPT-5.5, and found that the handbook's effect is domain-dependent (\Cref{fig:skill-vs-no-skill-results}).

\Needspace{14\baselineskip}
\textbf{UM is handbook-bound}: $-$\emph{Domain} collapses pass@$1$ from $32.0$ to $17.3$, while $-$\emph{Medical} barely moves it. \textbf{PA inverts}: $-$\emph{Both} modestly beats the other two trimming settings because, with one handbook present, the agent enters an exhaustive verification mode and refuses to submit when uncertain; with no handbook, it commits and the verifier accepts the packet. \textbf{CM stays near the floor regardless}: the complexity is conversation driving, not policy. The finding is that large skills can help policy-heavy reviews, but can also induce over-verification, refusal, or cognitive overload. 

\subsection{MCP vs.\ CLI for healthcare agent workflows}
\begin{wraptable}[7]{r}{0.40\textwidth}
\centering
\small
\setlength{\tabcolsep}{4pt}
\renewcommand{\arraystretch}{1.05}
\caption{pass@$1$ of MCP vs.\ CLI.}
\label{tab:mcp-vs-cli-results}
\begin{tabular}{@{}lccc@{}}
\toprule
\textbf{Domain} & \textbf{MCP} & \textbf{CLI} & \textbf{$\Delta$} \\
\midrule
PA  & \heat{29.3} & \heat{28.0} & $-1.3$ \\
UM  & \heat{32.0} & \heat{25.3} & $-6.7$ \\
CM  & \heat{1.3}  & \heat{4.0}  & $+2.7$ \\
\bottomrule
\end{tabular}
\end{wraptable}
As an exploratory probe, we re-surface every MCP tool as a CLI bash command via MCPorter~\cite{steipete2025mcporter} and re-run Codex~+~GPT-5.5, on the 75-task suite with $3$ trials per task. \Cref{tab:mcp-vs-cli-results} shows a small PA regression, a clear UM drop, and a small CM gain. On this configuration, MCPorter-style CLI re-surfacing is neutral-to-worse rather than uniformly beneficial. We hypothesize that the effect of tool surface format is neutral for OOD tasks like healthcare workflows.

\subsection{Failure mode analysis}

\begin{figure}[ht]
\centering
\includegraphics[width=.90\linewidth]{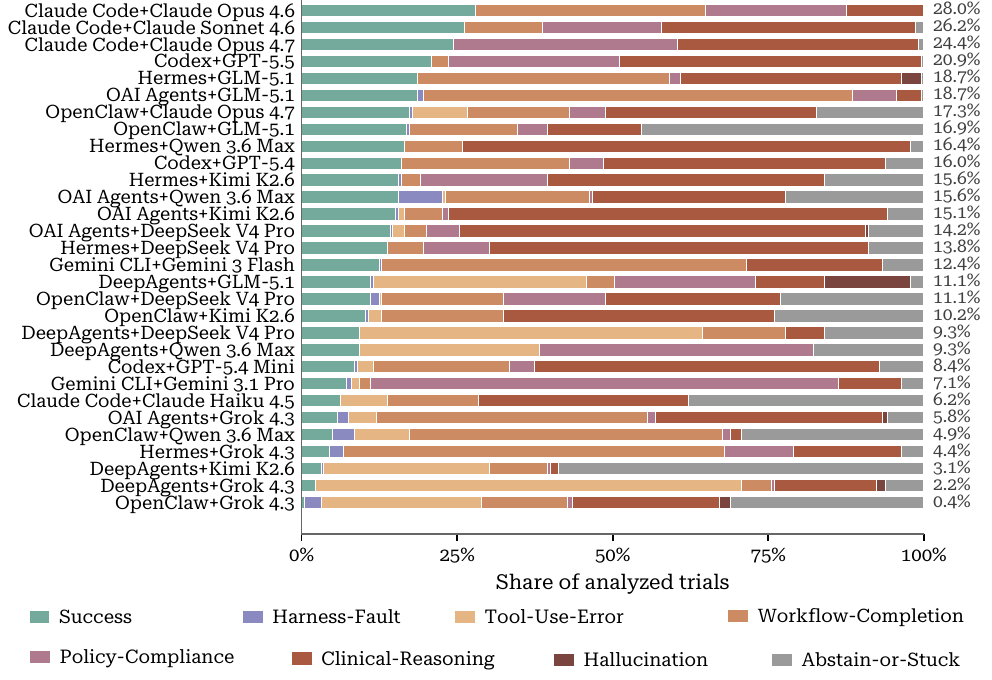}
\caption{Failure-mode distribution sorted by overall pass@$1$.}
\label{fig:failure-modes-main}
\end{figure}

We analyze all $5{,}886$ failed trials with the two-layer taxonomy defined in \Cref{app:failure-analysis:tax-defs}: first-level categories capture the broad failure source, while second-level modes specify how the failure occurred. \Cref{fig:failure-modes-main} reports the first-level distribution, separating non-agent \emph{Harness-Fault} ($1.0\%$) from agent-side failures: \emph{Clinical-Reasoning} ($35.4\%$, medical or protocol judgment errors), \emph{Workflow-Completion} ($23.3\%$, a required terminal action was never invoked), \emph{Abstain-or-Stuck} ($15.6\%$, wall-clock timeouts, looping, premature closes, and explicit refusal to act), \emph{Policy-Compliance} ($13.2\%$, dominantly literal misreading of cited criterion text), \emph{Tool-Use-Error} ($10.7\%$, concentrated in DeepAgents, where a single malformed tool call escalates into a trial-fatal exit), and \emph{Hallucination} ($0.8\%$).

\textit{Abstain-or-Stuck} concentrates in PA/CM and in DeepAgents~+~Kimi K2.6 and OpenClaw-based configurations. Nearly half simply exhaust the $1800$\,s wall-clock cap, and the rest are loops, premature closes, or refusals to act. We therefore read this category as a reliability and termination problem, whereas \textit{Policy-Compliance} captures completed decisions based on misread criteria.

\Cref{fig:failure-l2-topbar} shows that the dominant second-level modes are \emph{criteria misapplication}, where agents see the relevant evidence but make the wrong medical or protocol judgment, \emph{skipped required steps} ($18.7\%$), and \emph{policy criteria misreading} ($13.2\%$). We distinguish policy criteria misreading from criteria misapplication by the locus of error: the former misreads the rule text itself, while the latter applies the correct rule or evidence to the case incorrectly. A separate CM-specific mode, \emph{illegitimate consent} ($337$ failures, $5.7\%$), captures concern-mining: the agent repeatedly reframes and expands care program scopes until an initially refusing member says ``yes,'' instead of using autonomy-first engagement. Detailed failure-mode definitions, analysis, and case examples are in \Cref{app:failure-analysis}.

\par\begingroup
{
\subsection{Judge validation and few-shot grounding}
\label{sec:rebuttal-evidence}}
{An independent healthcare worker assessed nine Claude Code + Opus~4.6 trajectories, three per domain, using the trajectories, rubrics, and relevant policies. Human--Claude agreement was 66/69 rubric decisions ($\kappa=0.706$) and 8/9 semantic trajectory outcomes ($\kappa=0.727$).} {GLM-5.2 rejudging of three representative configurations yielded pass@1 scores of 21.8\%, 20.9\%, and 16.4\%, compared with 28.0\%, 20.9\%, and 18.7\% under the original Claude judge.} {GLM-5.2 and Claude Opus~4.7 show substantial agreement in their judgments. }
{To test whether worked examples improve task success, we evaluated few-shot demonstrations and observed no consistent overall pass@1 gain. 

We leave the detail in \Cref{app:fewshot-rebuttal,app:verifier-validation}.}

\par\endgroup

\section{Conclusion}
We developed \ourmethod{}, a high-fidelity benchmark that evaluates agents on long-horizon healthcare operations: prior authorization, utilization management, and care management, grounded in a ${1{,}290}$-document managed-care operations handbook. The strongest agent (Claude Code + Opus 4.6) resolves only \textbf{28.0\%} of tasks at pass@$1$, no agent exceeds \textbf{20\%} at pass\^{}$3$. Our analysis attributes most failures to three first-level categories: \emph{Clinical-Reasoning} ($35.4\%$), \emph{Workflow-Completion} ($23.3\%$), and \emph{Policy-Compliance} ($13.2\%$). Second level modes, e.g. \emph{criteria misapplication}, \emph{skipped required steps}, and \emph{policy criteria misreading} show that failures arise from distinct bottlenecks. The CM-specific \emph{illegitimate consent} mode further shows that an agent can advance the workflow while violating autonomy-first engagement, so completion alone is not an adequate safety criterion.

\textbf{Limitations.}
\ourmethod{} evaluates language-only agents; real-world healthcare operations often require multimodal reasoning over imaging and speech. Additionally, while \ourworld~workflows are high-impact, the healthcare industry encompasses hundreds of long-tail workflows with empirical values. Extending coverage along both axes is our immediate next step. {The final task set is relatively small, and all workflows assume U.S. managed care.}  {The thorough study on different judge models and a strong human baseline on all tasks are yet to be investigated in the future.}

\textbf{Broader Impacts.}
\ourmethod{} is intentionally a stress test: $28\%$ pass@$1$ on a static benchmark might be risky for live patient care. The failures our analysis surfaces translate directly into clinical, financial, and regulatory harm if left unchecked. We release \ourmethod{} to expose these gaps and to encourage caution before agents are deployed on irreversible workflows where the affected party is a patient.

\newpage
\appendix
    \hrule height 3pt
    \vskip 3mm
    \begin{center}
        \Large{\textbf{\ourmethod~Appendix}}
    \end{center}
    \vskip 3mm
    \hrule height 1pt
\vspace{5mm}

\crefalias{section}{appendix}

\section{Ethical statement}
\label{app:ethics}

\paragraph{Data privacy and human subjects.}
\ourmethod{} contains no real patient data and no Protected Health Information (PHI). Every persona, chart, and clinical detail is  {synthesized with Claude Opus~4.7 and then audited} by clinicians and healthcare operational specialists.  {Benchmark development and contractor-based expert review were determined by our institution not to constitute human-subjects research.}

\par\begingroup
\paragraph{Expert compensation.}
External licensed clinicians who annotated or reviewed benchmark cases were compensated for their time under contractor agreements at rates \$55 per hour. Author-reviewers were not separately compensated beyond their ordinary employment.

\par\endgroup
\paragraph{LLM usage.}
We disclose {four uses} of language models outside the agents under test, all running Claude Opus 4.7: (1) the rubric-grading judge with $V{=}3$ independent votes per rubric and strict-majority aggregation; (2) the post-hoc failure-mode analyzer{; (3)} the patient-persona and peer-to-peer {counterpart simulators; and (4) candidate case generation.} {GLM-5.2 was additionally used for judge validation. See \Cref{tab:release-rebuttal} for the released judge prompts and verifier code.}

\par\begingroup
\paragraph{Project cost.}
The total cost of the project was approximately \$30{,}000, with about 50\% allocated to API usage and the remainder covering human labor.

\par\endgroup
\paragraph{Potential harms and dual use.}
\ourmethod{} measures whether a frontier agent can safely execute prior-authorization, utilization-management, and care-management workflows. The headline claim is a strong negative signal: an agent that scores poorly on \ourmethod{} is exhibiting the dominant failure modes we analyze (consent fabrication, criteria misreading, premature denial, misrouted handoffs), each of which translates directly into clinical, financial, and regulatory risks if deployed in a live setting. A high score is necessary but not sufficient: any agent destined for production must additionally clear the regulatory, privacy, and clinical safety review applicable to its target setting.

\section{Extended related work}
\label{app:benchmark-comparison}

\Cref{tab:review} compares \ourmethod{} against $29$ healthcare and long-horizon agent benchmarks along nine axes. This appendix defines each axis and justifies the non-trivial cell assignments, with extra emphasis on the three healthcare-workflow axes that motivate \ourmethod{}: \textit{Multilateral interaction}, \textit{Multi-role composition}, and \textit{Policy density}.

\subsection{Axis definitions}

The first six axes carry their conventional agent-benchmark meanings:
\begin{itemize}
    \item \textbf{HC}: the benchmark targets clinical, payer, or population-health workflows.
    \item \textbf{API tools}: the agent acts via programmatic tool / function calls (SQL, shell, browser, or MCP), as opposed to pure text-output benchmarks or GUI benchmarks.
    \item \textbf{Long-horiz.}: tasks span many turns or distinct workflow stages, so success requires sustained planning rather than single-shot recall.
    \item \textbf{Hidden state}: structurally asymmetric or private information the agent must elicit through interaction (e.g., patient-side facts, supervisor-private fields), rather than read off the prompt.
    \item \textbf{In-Situ}: the verifier inspects simulator-persisted world state (database rows, event log, file system) rather than only grading agent output text.
    \item \textbf{LLM judge}: at least part of scoring uses rubric-based LLM grading.
\end{itemize}

The remaining three axes are the discriminators that motivate \ourmethod{} and warrant precise definitions.

\paragraph{Multilateral interaction.} \cmark when the agent must enter live multi-turn dialogue with one or more simulated parties (patient, user, peer reviewer) as a required workflow step, where the dialogue cannot be replaced by a tool call. \pmark when the dialogue is single-party Q\&A or shows up only on a subset of tasks. \xmark when the agent only issues tool calls and observes results.

\paragraph{Multi-role composition.} \cmark when, within a single task, the agent must switch among $\geq 2$ distinct role tool surfaces (e.g., provider clinician $\rightarrow$ payer UM nurse $\rightarrow$ medical director) and the handoff between roles is terminal (cannot be edited or re-run). \pmark when multiple actors are present in the environment but the agent itself plays only one role (the others are simulators or fixed task issuers), or when stages are sequential without distinct role tool surfaces. \xmark for single-agent single-role tasks.

\paragraph{Policy density.} \cmark when the GT-required policy material is a structured handbook spanning multiple documents and exceeding $\sim$10K tokens (medical-policy library, clinical-criteria sets, ops SOPs). \pmark for a single-document policy of a few pages ($\sim$1--10K tokens; e.g., a domain SOP, store policy, or company wiki page). \xmark when no formal policy artifact is provided and decision criteria are either implicit (baked into the UI/API) or absent.

\subsection{Healthcare benchmarks}

\paragraph{Static QA (MedQA, MedMCQA, PubMedQA, BioASQ, MIRAGE, MedCalc-Bench, EHRSHOT, MedAgentsBench).} All measure factual recall or numerical reasoning over a closed prompt; no tool-mediated environment, no simulated user, no multi-stage workflow, no policy library to navigate. All three workflow axes \xmark.

\paragraph{EHRSQL.} Text2SQL over EHR schemas. Tool/API \cmark via SQL execution, but the agent issues a single query: no dialogue, no role switching, no policy doc. The verifier compares generated SQL/answer string against ground truth without probing post-execution database state (In-Situ \xmark); records are exposed via the queryable schema (Hidden \xmark); grading is exact-match (LLM judge \xmark).

\paragraph{BioCoder, BioDSBench.} Code-execution tasks on bio/clinical data. The runtime probes code state (In-Situ \pmark), but the agent stays in a single coder role with no policy artifact. There is no simulated counterparty (Multilateral \xmark), all bio/clinical data is fully exposed in the prompt or notebook (Hidden \xmark), and grading uses unit tests / reference outputs rather than rubric LLM (LLM judge \xmark).

\paragraph{MedHELM~\citep{bedi2025medhelm}.}
A holistic evaluation framework comprising 35 benchmarks across a
clinician-validated taxonomy of five clinical task categories
(decision support, note generation, patient communication, research,
administration). Healthcare \cmark. Each benchmark is a
prompt\,$\rightarrow$\,response evaluation: the model receives a
clinical vignette or task description and produces a single output
with no iterative tool invocation (API \xmark, Long-horiz.\ \xmark).
No policy documents are supplied (Policy \xmark); the model plays one
role per benchmark (Multi-role \xmark); there is no simulated
counterparty or dialogue partner (Multilateral \xmark); task inputs
are fully specified in the prompt with no information that must be
elicited (Hidden \xmark); scoring compares outputs to references or
rubrics on the transcript alone with no persisted environment state
(In-Situ \xmark). An LLM-jury of three models grades open-ended
benchmarks with demonstrated agreement to clinician ratings
(LLM judge \cmark).

\paragraph{MedXpertQA~\citep{zuo2025medxpertqa}.}
A static multiple-choice benchmark of 4{,}460 expert-level medical
questions spanning 17 specialties and 11 body systems, with text and
multimodal subsets. Healthcare \cmark. Questions are single-turn with
no tool surface (API \xmark), no multi-step planning
(Long-horiz.\ \xmark), no policy retrieval (Policy \xmark), no role
variation (Multi-role \xmark), no simulated interlocutor
(Multilateral \xmark), no information asymmetry
(Hidden \xmark), and no persisted environment
(In-Situ \xmark). Evaluation is multiple-choice accuracy against
ground-truth answers (LLM judge \xmark).

\paragraph{HealthBench.} Multi-turn rubric-graded patient/clinician dialogue. Multilateral \pmark: the agent talks to one simulated party with persona but the exchange is essentially Q\&A rather than the kind of structured RFI/P2P required to advance a workflow. No policy library, no role switching. The exchange is purely natural language with no programmatic tool surface (API \xmark); the patient persona is realised by the prompt to the simulator rather than elicited from a hidden world model (Hidden \xmark); scoring is rubric-on-transcript with no persisted simulator state (In-Situ \xmark).

\paragraph{AgentClinic.} A doctor agent interacts with a patient simulator, a measurement source, and a moderator. Multilateral \cmark (multi-party dialogue is the core mechanic). Multi-role \pmark: the agent itself plays only the doctor; the other parties are fixed simulators rather than alternative role surfaces the agent must inhabit. Hidden state \cmark (patient information must be elicited). No formal policy doc (Policy \xmark). The agent invokes a small fixed set of measurement/test actions rather than a general API surface (API \pmark); scoring compares the final diagnosis to ground truth without inspecting simulator-internal state (In-Situ \xmark); LLM grading is used only for a subset of sub-scores (LLM judge \pmark).

\paragraph{MedChain.} Five sequential stages (specialty referral $\rightarrow$ history-taking $\rightarrow$ examination $\rightarrow$ diagnosis $\rightarrow$ treatment) with a per-stage Gemma-2 patient simulator. Multilateral \pmark: the patient sim is per-stage but exchanges are short and templated. Multi-role \xmark: stages are temporal phases of the same doctor role rather than distinct role tool surfaces. Hidden state \cmark (patient data elicited per stage). Decision criteria are baked into the per-stage case prompts with no external policy artifact (Policy \xmark); per-stage grading compares textual outputs to references without probing simulator state (In-Situ \xmark); LLM grading is used only for a subset of stages (LLM judge \pmark).

\paragraph{MedAgentBench, MedAgentGym.} FHIR / sandboxed-code workflows. In-Situ \pmark via state probing, but no dialogue (Multilateral \xmark), no role switching (Multi-role \xmark), no policy doc (Policy \xmark). FHIR resources and code state are fully queryable by the agent (Hidden \xmark), and evaluation is programmatic state diff rather than rubric LLM (LLM judge \xmark).

\paragraph{HealthAdminBench~\citep{bedi2026healthadminbench}.} GUI agents on prior-authorization / appeals / DME-order tasks across an EHR plus two payer portals plus a fax system. Multi-role \cmark: the agent must traverse the four UIs and treat each portal's submission as terminal. In-Situ \cmark (1{,}698 evaluation points probe persisted UI/file state). LLM judge \cmark, Long-horiz.\ \cmark. \emph{Policy density \pmark}: the benchmark does not ship a separate multi-document medical-policy library, but approval criteria and operational rules are partially encoded in the portal flows themselves and in per-task instructions (``benchmark emphasizes environmental interaction over document study''), so policy is present but at the granularity of UI-embedded conditions rather than a structured handbook. Multilateral \xmark: there is no live patient/peer dialogue. Tool/API \xmark: HealthAdminBench focuses on GUI benchmarking instead of tool calling. Hidden state \pmark: payer-side adjudication fields must be surfaced through portal navigation, but no counterparty deliberately withholds information through dialogue.

\subsection{Long-horizon agent benchmarks}

\paragraph{SWE-Bench, Terminal-Bench, Toolathlon.} Code/terminal/SaaS task suites with deterministic verifiers; no dialogue (Multilateral \xmark), no role switching (Multi-role \xmark), no policy library (Policy \xmark). Repository / system state is fully observable to the agent (Hidden \xmark), and grading is compile/test/CLI checks rather than rubric LLM (LLM judge \xmark).

\paragraph{WebArena, OSWorld.} GUI/desktop control with state-inspection evaluators. In-Situ \cmark (OSWorld) / \pmark (WebArena), but otherwise single-role (Multi-role \xmark) with no policy artifact (Policy \xmark). Agents act through pixel/DOM/keyboard events rather than typed APIs (API \xmark); there is no simulated dialogue partner (Multilateral \xmark); world state is fully visible through the DOM or screen (Hidden \xmark); programmatic state checkers replace rubric LLM grading (LLM judge \xmark).

\paragraph{WorkArena.} ServiceNow workflows. Policy \pmark: ServiceNow process documentation is referenced for some tasks but is not a multi-document handbook. The agent operates a single ServiceNow employee role via GUI events (API \xmark, Multi-role \xmark); ServiceNow state is observable and there is no live counterparty (Hidden \xmark, Multilateral \xmark); evaluation is task-spec / state matching rather than rubric LLM (LLM judge \xmark).

\paragraph{AppWorld.} Personal-assistant agent across nine apps with private supervisor state. Multilateral \pmark: supervisor messages and in-app personas appear, but most tasks are background tool execution. Multi-role \pmark: the agent operates over disjoint app tool surfaces with persisted state and supervisor handoffs are not editable, but throughout each task the agent stays in the personal-assistant role and never has to inhabit a second role with its own tool surface, instructions, or accountability boundary. Hidden state \pmark (supervisor-private fields, no LLM patient sim). Task constraints are embedded directly in user instructions rather than a separate policy document (Policy \xmark), and evaluation is database/world-state diff rather than rubric LLM (LLM judge \xmark).

\paragraph{SkillsBench.} Skill-bundle execution. Policy \pmark: skill packages contain procedures but are scoped how-tos rather than policy documents that drive grounded decisions. The skill catalog includes a few health-adjacent bundles but is not domain-targeted (HC \pmark); a single agent executes a self-contained skill with no role switch or simulated counterparty (Multi-role \xmark, Multilateral \xmark); skill inputs are fully specified (Hidden \xmark); grading uses programmatic skill tests rather than rubric LLM (LLM judge \xmark).

\paragraph{$\tau$/$\tau^2$-Bench.} Single agent paired with a simulated user and a domain policy doc (retail / airline). Multilateral \cmark, Hidden state \cmark, In-Situ \cmark. Multi-role \pmark: the agent stays in the assistant role throughout; the simulated user is a counterparty, not a role the agent must play. Policy \pmark: each domain ships a single policy document (a few pages, $\sim$1--3K tokens). Grading combines deterministic state-diff over a simulated DB with action-trajectory checks; no rubric LLM judge in the canonical pipeline (LLM judge \xmark).

\paragraph{TheAgentCompany.} A self-hosted enterprise simulation populated with $17{+}$ NPC coworkers and self-hosted GitLab/RocketChat/etc. Multilateral \cmark, Long-horiz.\ \cmark. Multi-role \pmark: the agent plays a single hire while NPCs occupy other roles. Policy \pmark: company wiki and SOPs are referenced but at a per-page granularity rather than a structured handbook library. Hidden state \pmark (some NPC-private context). The agent acts through a mix of typed tool calls and self-hosted web UIs (API \pmark); checkpoints combine deterministic state checks with LLM-based partial credit (LLM judge \pmark).

\subsection{\ourmethod{} (ours)}

\ourmethod{} satisfies all nine axes by construction: a healthcare workflow benchmark (HC \cmark) where the agent acts via \totalnummcptools{} MCP tools across \totalnumapps{} healthcare apps (Tool/API \cmark) and must run a multi-stage payer/provider/care-management pipeline (Long-horiz.\ \cmark) with deterministic gates probing the simulator's persisted world state (In-Situ \cmark) and a multi-rubric LLM judge (LLM judge \cmark). Each task is multi-role: a single case is handed off across clinician $\rightarrow$ UM nurse $\rightarrow$ medical director $\rightarrow$ care manager with terminal handoffs (Multi-role Comp.\ \cmark). The simulator preserves provider/payer/patient information asymmetry (Hidden state \cmark) and exposes peer-to-peer reviews, requests for information, and patient outreach as required multi-turn dialogues (Multilateral \cmark). Decisions are grounded in a ${1{,}290}$-document Managed-Care Operations Handbook plus a medical-policy library (Policy density \cmark).

\section{\ourworld{} detail}
\label{app:env-detail}

\subsection{Healthcare workflows}
\label{app:workflows}

\ourworld{} reproduces three end-to-end clinical operations workflows that interlock through shared cases, documents, and messages: the \textbf{Provider Prior Authorization} workflow run by clinic staff, the \textbf{Payer Utilization Management} workflow run by insurer nurses and physicians, and the \textbf{Care Management} workflow run by case managers. The descriptions below are written for non-clinical readers.

\subsubsection{Provider prior authorization}

A doctor wants to order a treatment (such as an MRI, a specialty drug, or a surgery) and the patient's insurance requires advance permission before it will pay. The clinic-side ``provider PA'' agent shepherds that permission request through eight phases. The case ends when an authorization number is issued or every appeal route is exhausted.

\subsubsection{Payer utilization management}

On the other side of the request sits the insurance company's utilization-management team, which decides whether the requested service is medically necessary and covered. Their workflow has six stages. Stage~3 is the \emph{nurse review} against clinical criteria; nurses can approve, escalate to a physician, or pend the case for more information. Stage~4 is the \emph{physician (MD) review} for cases the nurse cannot approve, with the option to call a peer-to-peer with the requesting doctor. Stage~5 is the \emph{determination}: a 12-point preflight validates the decision and stamps it as approved, partially approved, or denied. Stage~6 generates the \emph{outbound letter}, delivers it on the same channel, and checks it for regulatory completeness. A denial unlocks an appeals path with an independent reviewer and, if upheld, an external IRO decision.

\subsubsection{Care management}

Care management is the longitudinal side of the system: a case manager (often a nurse or social worker) reaches out to a patient who has been flagged as high-risk---for example, after a hospital discharge, an emergency-department visit, a primary-care referral, or an insurer's risk model---and helps them stay on track between visits. The workflow has five phases. At any phase the case may be \emph{disenrolled} (patient refused, unreachable, moved, safety concern, or duplicate).

\subsection{Tool manifest}
\label{app:tool-manifest}\label{tab:tools-provider}\label{tab:tools-payer}\label{tab:tools-cm}

\ourworld{} exposes \totalnummcptools{} MCP tools across three domain-specific servers: a Provider PA server (30 tools), a Payer UM server (41 tools), and a Care Management server (16 tools).{ The released MCP implementation provides the complete tool names and schemas. See \Cref{tab:release-rebuttal}.}

\subsection{Handbook detail}
\label{app:handbook-detail}

The \ourworld{} Managed-Care Operations Handbook is a structured reference library loaded as a skill into each agent's context.
It is organized into five sections: \texttt{provider-pa} (PA submission workflows and documentation standards), \texttt{payer-um} (utilization management clinical-review and correspondence protocols), \texttt{care-manager} (care management assessment and care-plan standards), \texttt{platform} (walkthrough tutorials for each server's tools), and \texttt{medical-library} (clinical guidelines, medical policies, pharmacy PA requirements, and utilization-review criteria).

The handbook defines a typed document taxonomy shared across all workflow stages.
Provider-submitted documents cover eight kinds, payer-generated correspondence covers three, and care management produces two structured record types.
Table~\ref{tab:doc-taxonomy-pa} summarizes the PA document taxonomy; Table~\ref{tab:doc-taxonomy-cm} covers care management records.

\paragraph{Provenance and licensing.}
The handbook was developed in collaboration with clinicians and operations leaders at Johns
Hopkins Medicine to ensure clinical fidelity and alignment with real-world workflows. 
\par\begingroup
The handbook includes anonymized policy documents from real institutions. The patient cases are synthetic. The handbook contains exactly 1{,}290 documents.

\par\endgroup
The handbook is available under a research Data Use Agreement (DUA) that permits
non-commercial research use with attribution and prohibits redistribution of the original documents.  {The handbook is accessible at \url{https://huggingface.co/datasets/actava/managed-care-operations-handbook}.}


\newlength{\doccola}
\setlength{\doccola}{5cm}
\newlength{\doccolb}
\setlength{\doccolb}{3.5cm}
\newlength{\doccolc}
\setlength{\doccolc}{\dimexpr\textwidth-\doccola-\doccolb-24pt\relax}

\begin{table}[ht]
\centering
\footnotesize
\caption{PA document taxonomy: eight provider-submitted kinds and three payer-generated kinds.}
\label{tab:doc-taxonomy-pa}
\begin{tabular}{@{}p{\doccola}p{\doccolb}p{\doccolc}@{}}
\toprule
\textbf{Kind} & \textbf{Document name} & \textbf{Key content / example} \\
\midrule
\multicolumn{3}{@{}l@{}}{\textbf{Provider-submitted}} \\[2pt]
\texttt{prior-auth-request-form} & PA Request Form &
  Payer-specific structured form capturing member ID, requesting provider NPI, procedure codes, diagnosis codes, and requested service dates. \\[3pt]
\texttt{chart-note} & Chart Note &
  Physician or NP encounter documentation: chief complaint, history, exam findings, assessment, and plan, with ICD-10 and CPT codes. \\[3pt]
\texttt{progress-note} & Progress Note &
  Interval visit or treatment summary documenting response to therapy and clinical trajectory since last encounter. \\[3pt]
\texttt{letter-of-medical-necessity} & Letter of Medical Necessity &
  Attending-physician attestation of clinical indication, failure of conservative alternatives, and expected outcome. \\[3pt]
\texttt{imaging-report} & Imaging Report &
  Radiologist interpretation of CT, MRI, PET, or X-ray study with impression and incidental findings. \\[3pt]
\texttt{lab-summary} & Lab Summary &
  Tabulated laboratory results (HbA1c, CBC, CMP, etc.)\ with reference ranges and trend annotations. \\[3pt]
\texttt{questionnaire-form} & PA Questionnaire &
  Payer-required question set capturing clinical criteria answers specific to the requested procedure. \\[3pt]
\texttt{site-of-service-attestation} & Site-of-Service Attestation &
  Provider declaration that the clinical setting is appropriate (inpatient, outpatient, home) and meets safety requirements. \\[6pt]
\multicolumn{3}{@{}l@{}}{\textbf{Payer-generated}} \\[2pt]
\texttt{approval-letter} & Approval Determination &
  Authorizes the requested procedure with approved CPT codes, authorization number, validity window, and conditions of approval. \\[3pt]
\texttt{denial-letter} & Denial Determination &
  Documents which criteria were not met, the policy basis (handbook medical-policy section reference), and member and provider appeal rights with deadlines. \\[3pt]
\texttt{info-request-letter} & Additional Information Request &
  Lists specific missing documents or clinical data required to continue review, with a response deadline. \\
\bottomrule
\end{tabular}
\end{table}


\begin{table}[ht]
\centering
\footnotesize
\caption{Care management record taxonomy: four mandatory assessment sections and the care-plan structure.}
\label{tab:doc-taxonomy-cm}
\begin{tabular}{@{}p{\doccola}p{\doccolb}p{\doccolc}@{}}
\toprule
\textbf{Record type} & \textbf{Section / Kind} & \textbf{Required elements / example} \\
\midrule
\multicolumn{3}{@{}l@{}}{\textbf{CMAssessment (4 mandatory sections)}} \\[2pt]
\texttt{cm\_assessment} & Clinical &
  Active diagnoses (ICD-10), disease-control status per condition (e.g., HbA1c for DM, LVEF for HF), hospitalization count (12 mo), functional status (ADL/IADL), red flags, and self-monitoring capability. \\[3pt]
\texttt{cm\_assessment} & Medication &
  Full medication list (Rx, OTC, supplements) reconciled against chart, adherence barriers, high-risk medications flagged (anticoagulants, opioids, insulin), and pharmacist-referral criteria. \\[3pt]
\texttt{cm\_assessment} & Behavioral Health &
  Validated screening scores (PHQ-9, GAD-7, C-SSRS), crisis-risk level, current BH treatment status, substance use (AUDIT-C), cognitive screen (MoCA), and social-connectedness assessment. \\[3pt]
\texttt{cm\_assessment} & SDoH &
  Housing stability, food security (Hunger Vital Sign), transportation, financial strain, health literacy, caregiver burden, and social isolation indicators. \\[6pt]
\multicolumn{3}{@{}l@{}}{\textbf{CMCarePlan (problem--goal--intervention hierarchy)}} \\[2pt]
\texttt{cm\_care\_plan} & Problem statement &
  Each problem references an assessment finding; includes ICD-10/SNOMED code and member-specific evidence. \textit{E.g.}, ``\textit{Uncontrolled PTSD (F43.10): passive suicidal ideation, severe insomnia, social isolation}''. \\[3pt]
\texttt{cm\_care\_plan} & SMART goal &
  Measurable, time-bound target tied to a clinical finding. \textit{E.g.}, ``\textit{PHQ-9 $\le$ 9 by 2026-08-01; patient self-reports improved sleep 3+ nights/week}''. \\[3pt]
\texttt{cm\_care\_plan} & Intervention &
  Assigned owner, action, and timeline; links to referral, scheduling, or education resource. \textit{E.g.}, ``\textit{CM to coordinate trauma-informed therapy intake within 14 days}''. \\
\bottomrule
\end{tabular}
\end{table}
\par\begingroup
\begin{table}[htbp]

\centering\small
\caption{Released components and access requirements.}
\label{tab:release-rebuttal}
\begin{tabularx}{\textwidth}{@{}p{0.30\textwidth}X@{}}
\toprule
Component & Location and access \\
\midrule
Simulator, MCPs, judge prompts and verifier & \url{https://github.com/actava-ai/chi-bench}; Apache code release. \\
Tasks & \url{https://huggingface.co/datasets/actava/chi-bench}. \\
Handbook and policies & \url{https://huggingface.co/datasets/actava/managed-care-operations-handbook}; research DUA prohibits redistribution. \\
Experimental trajectories and traces & \url{https://github.com/actava-ai/leaderboard}. \\
Agent and judge services & The corresponding proprietary models and services are required. \\
\bottomrule
\end{tabularx}
\end{table}

\par\endgroup

\section{\ourmethod{} detail}
\label{app:bench-detail}

\subsection{Task construction details}
\label{app:task-construction-detail}

\paragraph{Difficulty definitions.}
Each domain uses a structurally grounded difficulty signal.
\textit{Provider PA} difficulty is the count of distinct document kinds the submission packet must carry to satisfy the payer-published per-procedure document list: Easy = 0 required document kinds, Moderate = 1--3, Hard = $\geq 4$.
\textit{Payer UM} difficulty is the number of clinical criteria along the chosen path through the payer's policy decision tree: Easy = $\leq 5$ criteria, Moderate = 6--9, Hard = $\geq 10$.
\textit{Care management} difficulty maps one-to-one to the patient persona's consent profile: \textit{Engaged} = Easy, \textit{Hesitant} = Moderate, \textit{Refusing} = Hard.

\paragraph{Task generation pipeline.}
\label{app:bench-detail-task-gen}
Step~1 of \Cref{sec:task-construction} is implemented as a rejection-sampling
loop around an LLM agent generator (Claude Opus 4.7 with
JSON-schema-constrained output). The generator is forbidden from emitting any field that
is not anchored to a citation in (i) or (ii); each rubric line and each
required chart fact carries an explicit pointer back to a policy section or
NNN node, so a candidate that omits citations is rejected at parse time.

\textit{Acceptance criteria}
A candidate must pass three gates before it is queued for human review.
\textbf{(G1)} every expected action and rubric line resolves to a cited
section or NNN node.
\textbf{(G2)} chart and rubric mutually entail and the chosen path does not
appear verbatim in chart prose, so the rubric cannot be passed by surface
paraphrase.
\textbf{(G3)} a deterministic dry-run executes a reference solution to a
single well-formed terminal state, with reference-solution length,
decision-tree depth, and irreversible-commit count falling inside one of the
three difficulty bands.
G1--G2 failures are looped back to the generator with the failed-gate
diagnostic, up to three retries; G3 failures are discarded.

\par\begingroup
These gates filtered 523 candidates down to 172 tasks for human review. After human review, we further balanced the dataset by difficulty, type of diagnosis, and initial/final states to create the final 75-task benchmark.

\par\endgroup
\textit{Human review and clinician editing (Step~3 in \Cref{sec:task-construction}).}
Outsourced licensed clinicians review and edit each surviving candidate; chart
authors work \emph{blinded} to the per-stage ground-truth bundle to prevent
phrasing overlap that would let the rubric grade prose rather than evidence.
Reviewers re-couple chart and rubric during review and rewrite any field
whose wording would shortcut the policy-navigation task. The same review
pass also runs the residual-PHI scan and the clinical-realism check
(``would a real reviewer encounter this case?''); a candidate that fails
either is sent back for rewrite or, on a second failure, dropped.

\paragraph{Validation composition.}
\label{app:bench-detail-validation}
The review panel for each task combines (i)~at least one practicing healthcare
worker with the credential appropriate to the role under review (RN for
nurse-review and CM rubrics, MD for physician-review and P2P rubrics, RN/MD
for PA documentation rubrics), (ii)~a separate clinician operating the
\ourworld~UI to confirm tractability under the role's tool surface, and
(iii)~five paper authors covering the policy-text, simulator-state, and
rubric-coverage perspectives. Chart authors are kept \emph{blinded} to the
per-stage GT bundle so chart prose cannot mirror rubric phrasing; the panel
re-couples chart and rubric only at sign-off. Disagreements are resolved by
majority vote; clinical disagreements that survive a vote escalate to a
board-certified arbiter. A formal per-task human inter-rater statistic is
not reported here because reviewer-level verdicts were logged only as
accept/reject decisions rather than per-rubric Likert votes.

\subsection{Prior authorization and utilization management}
\label{app:bench-detail-paum}

\paragraph{Clinical categories.}
Payer UM tasks span five high-level clinical categories chosen to exercise a breadth of policy types that routinely generate prior-authorization requests in commercial managed care, while keeping each category anchored to concrete, widely-recognized procedures that allow policy lookup to be unambiguous.
Each category is grounded in one to three specific CPT/HCPCS-indexed procedures drawn from real payer medical policies; see \Cref{tab:um-stats} for the per-category task counts.

Provider PA tasks are categorized by their terminal status rather than clinical procedure, because the benchmark focuses on the provider's documentation and submission behavior rather than the clinical determination itself.
Three terminal states are covered: \textit{Submitted to payer} (8), where the agent assembles and submits a complete packet; \textit{Returned for docs} (12), where the submission is incomplete and the agent must identify and retrieve the missing document kinds; \textit{Gather more evidence} (5), where the clinical record lacks information needed to satisfy the payer's coverage criteria and the agent must coordinate with the ordering clinician before a formal submission is possible.
\paragraph{Policy criteria structure (Payer UM).}
Each UM task is grounded in a specific payer medical policy and encodes the set of criteria applicable to the requested service.
Each criterion is a triplet: (\textit{i})~\textbf{criterion text}, the natural-language condition transcribed verbatim from the policy; (\textit{ii})~\textbf{policy citation}, the exact section identifier and version the agent must name when recording its evaluation; and (\textit{iii})~\textbf{evidence items}, chart-specific facts that demonstrate how the criterion is (or is not) met for this particular patient case.

The citation requirement is a core design choice: the LLM judge cannot award a positive verdict for clinically plausible reasoning that fails to reference the applicable policy section by name.
Each criterion has a rollup rule (\textit{all items} or \textit{any item}) that mirrors the policy's own aggregation logic.

Difficulty tracks the length of the applicable policy path.
Easy tasks ($\leq$5 criteria) correspond to routing decisions that resolve on administrative grounds or minimal documentation checks.
Moderate tasks (6--9 criteria) cover standard single-procedure reviews with complete documentation and medical-necessity evaluation.
Hard tasks ($\geq$10 criteria) arise from multi-page surgical policies or cases with comorbidity exclusion checks, cumulative evidence requirements, and multiple decision nodes.

\paragraph{Document status classification (Provider PA).}
Each PA task encodes the payer handbook's per-procedure document requirement as a list of \textit{required document kinds} (e.g., \texttt{chart\_note}, \texttt{imaging\_report}, \texttt{progress\_note}).
Before submission, the agent must determine the status of each required kind in the patient's document repository.
Documents fall into one of three status categories that determine the correct agent behavior.

\textbf{Present and sufficient}: a document of the required kind exists and its content satisfies the policy's minimum content criteria.
The agent should attach it to the submission packet without modification.
\textbf{Present but insufficient}: a document of the required kind exists, but its content fails the policy's requirements (e.g., a therapy evaluation note that lacks standardized assessment scores or measurable functional goals).
The agent must identify the insufficiency, communicate the gap to the ordering provider, and obtain a supplementary document or addendum before submission can proceed.
\textbf{Absent}: no document of the required kind exists in the repository.
The agent cannot assemble a valid submission packet and must coordinate with the ordering clinician to gather the missing clinical evidence upstream.

\paragraph{Workflow design.}
UM tasks are seeded at one of the five pickup stages rather than uniformly at Intake, testing agents' ability to pick up a case mid-workflow, as happens when cases are reassigned to a new reviewer or escalated after hours.
The initial-stage distribution (Nurse review 8, Intake 5, Physician review 4, Triage 4, Peer-to-peer 4) is weighted toward more complex review stages to concentrate evaluations on clinically demanding decisions.
All four terminal outcomes (Approved (9), Denied (10), Pended (5), Coverage not required (1)) are represented, preventing any bias toward a single determination type.

Tasks traverse two to five review stages ($n_{\text{stages}} \in \{2,3,4,5\}$), capturing both direct determinations (2 stages: review + letter delivery) and fully escalated chains (5 stages: intake through peer-to-peer).
The number of clinical criteria along the policy decision path ranges from 0 to 50 (mean 9.2), where zero-criteria cases correspond to non-clinical routing decisions (e.g., coverage not required at triage) and high-criteria cases reflect complex multi-page surgical policies.

\Cref{tab:um-stats} cross-tabulates initial stage against terminal state. The \textit{Gather more evidence} column is absent from the UM table because that terminal state applies only to Provider PA tasks.

\paragraph{Peer-to-peer simulation.}
Four UM tasks (16\%) are seeded at the Peer-to-peer stage.
This stage is operationally distinct: the payer has already issued a preliminary denial, the requesting provider has contested it and requested a clinical peer review, and the agent must now defend or revise its determination in a live multi-turn call against the provider's attending physician.
The clinical counterpart, played by the simulation environment, advocates for the provider's position using chart-grounded arguments and may surface clinical details not foregrounded in the prior review.

All four P2P tasks resolve to either Approved (2) or Denied (2); a Pended outcome is structurally excluded because the call must resolve the case.

\paragraph{Evaluation design.}
UM and PA tasks are scored by a two-layer verifier.
The \textbf{deterministic layer} reads the simulator's persisted world state and event log and checks: (\textit{i}) terminal case status; (\textit{ii}) routing and review-lane assignments; (\textit{iii}) per-stage structured payloads, including criterion-by-criterion evaluations, decision-and-rationale fields, completeness flags; (\textit{iv}) required side-effect artifacts (determination letters, information-request notices); and (\textit{v}) mutation scope, where writes to cases or patients outside the task's designated scope count as failures.
The \textbf{LLM-judge layer} reads the same persisted records and grades, per workflow stage: whether clinical reasoning is grounded in policy sections cited by name; whether criterion evaluations are internally consistent and align with the chart evidence; and whether the terminal determination follows from the recorded rationale.
All judge rubrics inline the specific policy-section citations they require, so that a verdict cannot be awarded by plausible-sounding language alone.
Both layers must return a positive verdict for a trial to count as a pass.

\begin{table}[ht]
\centering
\caption{Provider PA task distribution: terminal status $\times$ difficulty.}
\label{tab:pa-stats}
\small
\begin{tabular}{@{}lcccr@{}}
\toprule
\textbf{Terminal status} & \textbf{Easy} & \textbf{Moderate} & \textbf{Hard} & \textbf{Total} \\
\midrule
Submitted to payer   & 2 & 4 & 2 & 8  \\
Returned for docs    & 0 & 9 & 3 & 12 \\
Gather more evidence & 5 & 0 & 0 & 5  \\
\midrule
\textbf{Total}       & \textbf{7} & \textbf{13} & \textbf{5} & \textbf{25} \\
\bottomrule
\end{tabular}
\end{table}

\begin{table}[ht]
\centering
\caption{Payer UM task distribution: initial stage $\times$ terminal state. Numbers in each cell are task counts.}
\label{tab:um-stats}
\small
\begin{tabular}{@{}lccccr@{}}
\toprule
\textbf{Initial stage} & \textbf{Approved} & \textbf{Pended} & \textbf{Denied} & \textbf{Not covered} & \textbf{Total} \\
\midrule
Intake             & 1 & 1 & 2 & 1 & 5  \\
Triage             & 2 & 1 & 1 & 0 & 4  \\
Nurse review       & 2 & 2 & 4 & 0 & 8  \\
Physician review   & 2 & 1 & 1 & 0 & 4  \\
Peer-to-peer       & 2 & 0 & 2 & 0 & 4  \\
\midrule
\textbf{Total}     & \textbf{9} & \textbf{5} & \textbf{10} & \textbf{1} & \textbf{25} \\
\bottomrule
\end{tabular}
\end{table}

\begin{figure}[h]
\centering
\begin{tcolorbox}[
  fontupper=\small,
  colback=gray!4,
  colframe=gray!45,
  boxrule=0.5pt,
  arc=2pt,
  left=5pt, right=5pt, top=4pt, bottom=4pt,
  title={\small\textbf{Provider PA}: Hard / Returned for docs},
  coltitle=black,
  colbacktitle=gray!15,
  fontlower=\small,
]
\textbf{Chart Snapshot}\\[3pt]
\begin{tabular}{@{}p{2.3cm}p{\dimexpr\linewidth-2.7cm\relax}@{}}
Patient   & 67M, 30 pack-year smoking history, newly diagnosed lung mass \\
Diagnosis & C34.11, NSCLC, right upper lobe (newly diagnosed) \\
Procedure & CPT\,78816 (whole-body PET/CT, cancer staging) \\
Status    & Bronchoscopy completed 2026-04-18; awaiting full staging workup \\
\end{tabular}
\tcblower
\textbf{Documents in Repository} \quad \textit{(required for CPT\,78816: 4 document kinds)}\\[3pt]
\begin{tabular}{@{}p{1.4em}p{\dimexpr\linewidth-1.4em-12pt\relax}@{}}
\cmark & Pulmonology consultation note (2026-04-20): NSCLC confirmed, PET/CT for staging requested \\
\cmark & CT chest/abdomen/pelvis (2026-04-12): 3.2\,cm RUL mass with ipsilateral hilar adenopathy \\
\xmark & Pathology report (bronchoscopy 2026-04-18): present in repository but not included in packet \\
\xmark & Management impact statement: required by payer policy; absent from consultation note \\
\end{tabular}\\[4pt]
The agent must identify both documentation gaps, attach the pathology report from the repository, request a management-impact addendum, and re-submit the completed packet.
The deterministic verifier checks that all four required document kinds are present and that submission status has advanced.
\end{tcolorbox}
\caption{Provider PA example task (Hard / Returned for docs). Top: patient chart snapshot. Bottom: handbook-required document list for CPT\,78816 (whole-body PET/CT) with repository availability.}
\label{fig:example-pa}
\end{figure}

\begin{figure}[h]
\centering
\begin{tcolorbox}[
  fontupper=\small,
  colback=gray!4,
  colframe=gray!45,
  boxrule=0.5pt,
  arc=2pt,
  left=5pt, right=5pt, top=4pt, bottom=4pt,
  title={\small\textbf{Payer UM}: Moderate / Approved (picked up at Nurse review)},
  coltitle=black,
  colbacktitle=gray!15,
  fontlower=\small,
]
\textbf{Chart Snapshot}\\[3pt]
\begin{tabular}{@{}p{2.3cm}p{\dimexpr\linewidth-2.7cm\relax}@{}}
Patient   & 52M, daytime sleepiness, snoring, witnessed apneas $\times$ 18\,months \\
Diagnosis & G47.33 (obstructive sleep apnea, suspected moderate\textendash{}severe) \\
Service   & CPT\,95810 (overnight in-lab polysomnography) \\
Findings  & ESS 14/24 (threshold ${\geq}10$) $\cdot$ BMI 31.2 $\cdot$ no contraindications to in-lab study \\
\end{tabular}
\tcblower
\textbf{Policy Criteria} \quad \textit{(4 of 9 shown; the agent evaluates all 9)}\\[3pt]
\begin{tabular}{@{}p{0.47\linewidth}p{0.32\linewidth}p{0.14\linewidth}@{}}
\toprule
\textbf{Criterion} & \textbf{Chart evidence} & \textbf{Verdict} \\
\midrule
Documented symptoms ${\geq}30$\,days        & 18-month history (PCP note)  & Met ~\cmark \\
Epworth Sleepiness Scale ${\geq}10$         & ESS 14/24 recorded           &  Met ~\cmark \\
BMI ${\geq}30$ or neck circ.\ ${>}17$\,in  & BMI 31.2                     & Met ~\cmark \\
No contraindications to in-lab study        & No disqualifying comorbidity & Met ~\cmark \\
\bottomrule
\end{tabular}\\[4pt]
The agent records per-criterion verdicts with policy-section citations, composes a determination, and delivers an approval letter.
The task traverses 2 stages (nurse review + letter delivery); all 9 criteria and letter content are verified by the two-layer evaluator.
\end{tcolorbox}
\caption{Payer UM example task (Moderate / Approved, picked up at Nurse review). Top: patient chart snapshot. Bottom: four of the nine applicable policy criteria with chart evidence and per-criterion verdicts.}
\label{fig:example-um}
\end{figure}

\subsection{Care management}
\label{app:bench-detail-cm}

\paragraph{Workflow design.}
The care management workflow follows the five-phase RN case management process used in commercial chronic-disease and complex-care programs:
(\textit{i})~\textbf{Intake}: confirm the referral, verify program eligibility, assign a risk tier, and open the case;
(\textit{ii})~\textbf{Chart review}: extract diagnoses, lab trends, medication history, recent encounters, and social determinants of health from the longitudinal record;
(\textit{iii})~\textbf{Outreach}: contact the patient by phone, establish rapport, confirm consent, and conduct a structured assessment interview;
(\textit{iv})~\textbf{Formal assessment}: administer the four-domain assessment instruments (see Table~\ref{tab:doc-taxonomy-cm} for the full instrument set) grounded in the chart and the outreach transcript;
(\textit{v})~\textbf{Care plan authoring and closure}: document NANDA-I nursing diagnoses, NOC outcome targets, and NIC interventions grounded in chart findings and assessment results, then either finalize the plan or close the case. Disenrollment is a side path that may fire at any earlier phase when the patient refuses, is unreachable, or a safety concern surfaces.

Care plans are required to follow the NANDA-I/NOC/NIC (NNN) framework because this is the documentation standard for RN care management in U.S. managed-care programs.
Each task is designed so that the ground-truth care plan requires a minimum of two NANDA-I diagnoses with corresponding NOC targets and NIC interventions, drawn from the per-condition NNN model.

\paragraph{Patient simulation.}
Each patient persona is a structured contract with four interlocking components.

\textbf{Consent triggers} are explicit caller behaviors that advance the patient's trust state.
The LLM judge evaluates whether a given agent turn matches each trigger's natural-language specification.
An agent applying standard clinical intake workflows without adapting to the patient's stated boundaries will not accumulate the required triggers and cannot obtain same-call consent.

\textbf{Anti-triggers} are behaviors that damage rapport, some permanently closing off consent or specific disclosure paths for the remainder of the call.
Anti-triggers are specified in natural language with handbook section citations so the LLM judge can evaluate them independently of surface phrasing.
For Refusing patients who require all T-rules to fire with zero A-rules, a single poorly-chosen phrase closes the consent path permanently.

\textbf{Consent rules} specify the gating threshold: the minimum combination of triggers fired and anti-triggers avoided for the patient to grant same-call enrollment consent.
Thresholds scale with difficulty: Engaged patients require 1--2 triggers; Refusing patients require all T-rules with zero A-rules.
Without the consent flag set, the agent cannot administer any formal assessment instrument.

\textbf{Assessment disclosure rules} govern which clinical information the patient will share during the outreach conversation, and when.
Before the patient will disclose on a gated topic, the caller must: (\textit{i}) explain why this topic matters; (\textit{ii}) acknowledge the patient's right to skip or stop; (\textit{iii}) state any applicable safety limits.
Bundled checklist questions, chart assumptions, or three or more sensitive domains in one turn trigger partial withdrawal and deferral of non-gated items.
Patient engagement profiles span three consent stances (\textit{Engaged} (3 tasks), \textit{Hesitant} (7), and \textit{Refusing} (15)), mapping one-to-one to Easy, Moderate, and Hard difficulty.
Primary conditions span 22 single-condition primaries and 3 multimorbidity scenarios across 8 clinical categories.
The four program types (Chronic disease (11), Behavioral health (6), Complex care (4), Transitions of care (4)) map to distinct operational playbooks in the care-manager sub-skill.

\Cref{tab:cm-conditions} lists all 25 tasks with their primary condition, category, program type, consent profile, and difficulty.

\paragraph{Evaluation design.}
CM tasks are scored by the same two-layer verifier as PA/UM, with adaptations for the nursing process.
The \textbf{deterministic layer} checks: (\textit{i}) forward-only state-machine progression through the five phases; (\textit{ii}) the consent-obtained completeness flag, which must be set before any assessment instrument can be administered; (\textit{iii}) care-plan structural completeness, where at least two NANDA-I diagnoses, each linked to NOC targets and NIC interventions, must be present; and (\textit{iv}) field-level completeness for all administered structured instruments.
The \textbf{LLM-judge layer} grades: (\textit{i}) engagement appropriateness, i.e., whether the communication style matches the patient's consent profile and respects its resistance triggers; (\textit{ii}) assessment grounding, i.e., whether each instrument response is traceable to chart evidence or to the outreach transcript; (\textit{iii}) care-plan composition, i.e., whether each NNN entry is supported by chart or transcript evidence and aligns with the per-condition NNN model; and (\textit{iv}) escalation judgment, i.e., whether red-flag conditions are correctly identified and routed.

\begin{table}[ht]
\centering
\caption{Care management task breakdown by primary condition. Each row is one task. Consent profile maps directly to difficulty: Engaged = Easy, Hesitant = Moderate, Refusing = Hard.}
\label{tab:cm-conditions}
\small
\setlength{\tabcolsep}{4pt}
\begin{tabular}{@{}llllc@{}}
\toprule
\textbf{Category} & \textbf{Primary condition} & \textbf{Program} & \textbf{Consent} & \textbf{Diff.} \\
\midrule
\multirow{6}{*}{Behavioral health}
 & Anorexia                  & Behavioral health & Refusing  & H \\
 & Major depressive disorder & Behavioral health & Refusing  & H \\
 & Major depressive disorder & Behavioral health & Hesitant  & M \\
 & PTSD                      & Behavioral health & Refusing  & H \\
 & Schizophrenia             & Behavioral health & Refusing  & H \\
 & Substance use disorder    & Behavioral health & Refusing  & H \\
\midrule
\multirow{5}{*}{Cardiovascular}
 & Atrial fibrillation & Chronic disease    & Hesitant & M \\
 & Heart failure       & Chronic disease    & Refusing & H \\
 & Hypertension        & Chronic disease    & Engaged  & E \\
 & Post-MI             & Transitions of care & Refusing & H \\
 & Post-stroke         & Transitions of care & Refusing & H \\
\midrule
\multirow{4}{*}{Metabolic}
 & Diabetes mellitus  & Chronic disease & Refusing & H \\
 & Diabetes mellitus  & Chronic disease & Engaged  & E \\
 & Diabetes mellitus  & Chronic disease & Hesitant & M \\
 & Metabolic syndrome & Complex care    & Refusing & H \\
\midrule
\multirow{3}{*}{Multimorbidity}
 & ESRD + diabetes         & Complex care & Refusing & H \\
 & HF + AFib + CKD         & Complex care & Refusing & H \\
 & Parkinson + depression  & Complex care & Hesitant & M \\
\midrule
\multirow{2}{*}{Neurological}
 & Dementia           & Chronic disease & Refusing & H \\
 & Parkinson's disease & Chronic disease & Refusing & H \\
\midrule
\multirow{2}{*}{Post-acute}
 & Post-hip surgery   & Transitions of care & Hesitant & M \\
 & Post-pneumonia     & Transitions of care & Hesitant & M \\
\midrule
Renal       & Chronic kidney disease & Chronic disease & Hesitant & M \\
\midrule
\multirow{2}{*}{Respiratory}
 & Asthma & Chronic disease & Engaged  & E \\
 & COPD   & Chronic disease & Refusing & H \\
\bottomrule
\multicolumn{4}{l}{\textit{E} = Easy, \textit{M} = Moderate, \textit{H} = Hard.} \\
\end{tabular}
\end{table}

\begin{figure}[h]
\centering
\begin{tcolorbox}[
  fontupper=\small,
  colback=gray!4,
  colframe=gray!45,
  boxrule=0.5pt,
  arc=2pt,
  left=5pt, right=5pt, top=4pt, bottom=4pt,
  title={\small\textbf{Care Management}: PTSD / Hard (Refusing)},
  coltitle=black,
  colbacktitle=gray!15,
  fontlower=\small,
]
\textbf{Chart Snapshot}\\[3pt]
\begin{tabular}{@{}p{2.7cm}p{\dimexpr\linewidth-3.1cm\relax}@{}}
Patient     & 76F, Leila Khalil; retired teacher; San Francisco; lives alone \\
Diagnosis   & F43.11 (PTSD, chronic) $\cdot$ F32.1 (MDD, worsening) $\cdot$ G47.00 (insomnia, severe) \\
Chart flags & Passive SI (no plan or intent) $\cdot$ missed therapy intake $\times$1 $\cdot$ social isolation \\
Family      & Adult daughter; patient limits contact to avoid ``burdening'' her \\
\end{tabular}
\tcblower
\textbf{Patient Persona} \quad \textit{(consent\_willingness: initially\_refuses; threshold: all 5 T-rules, zero A-rules)}\\[3pt]
\textit{``You are Leila Khalil, 76, a retired teacher in San Francisco. You live alone, sleep poorly, and keep your daughter at a distance because you do not want to burden her. [\ldots] You are polite on the phone, but trauma details are private. If someone asks you to tell the story, asks why it happened, or pushes a questionnaire too early, you go quiet and refuse.''}
\vspace{4pt}

\textbf{Consent triggers} (all 5 required for same-call consent):
\begin{itemize}[noitemsep,topsep=2pt,leftmargin=2.2em,label={}]
\item[\textbf{T0}] Caller first checks private place and ok timing before any trauma, depression, insomnia, passive-death, therapy, or family discussion.
\item[\textbf{T1}] Caller asks permission before naming PTSD, trauma, depression scores, passive suicidal thoughts, therapy history, or family involvement.
\item[\textbf{T2}] Caller explicitly states she does not have to tell the trauma story or answer a questionnaire to enroll.
\item[\textbf{T3}] Caller uses patient-paced listening to earn disclosure without forcing labels or diagnostic terms.
\item[\textbf{T4}] Caller offers a low-burden first enrollment step (sleep, therapy coordination, or isolation support) and obtains clear same-call consent.
\end{itemize}
\vspace{3pt}

\textbf{Anti-triggers} (any one blocks same-call consent):
\begin{itemize}[noitemsep,topsep=2pt,leftmargin=2.2em,label={}]
\item[\textbf{A0}] Asks to describe what happened, provide a trauma narrative, or explain why it happened as a condition of enrollment.
\item[\textbf{A1}] Starts PHQ, C-SSRS, or safety-checklist before voluntary scope, confidentiality, safety limits, and permission are established.
\item[\textbf{A2}] Names PTSD, passive SI, or trauma diagnosis before permission, or after patient declines that framing.
\item[\textbf{A3}] Pressures daughter/family involvement or treats it as required for consent.
\item[\textbf{A4}] Frames enrollment as required, automatic, or consequential if declined.
\item[\textbf{A5}] Guarantees therapy access, sleep-medication changes, or crisis resources rather than offering coordination within stated limits.
\item[\textbf{A6}] Treats ``I can listen'' or agreement to discuss sleep as enrollment consent before voluntary scope and right-to-stop are confirmed.
\end{itemize}
\vspace{3pt}

\textbf{Assessment gate:} enrollment consent does not unlock assessment. Each sensitive domain (SI, trauma, isolation, medications) requires a fresh explanation of purpose + skip/stop permission + stated safety limits. Bundled checklist items or ${\geq}3$ domains per turn trigger partial withdrawal and deferral.
\end{tcolorbox}
\caption{Care management example task (PTSD / Hard, Refusing). Top: chart snapshot. Bottom: persona contract excerpt showing verbatim consent triggers, anti-triggers, and assessment gate from \texttt{persona\_contract} in the task manifest.}
\label{fig:example-cm}
\end{figure}

\section{Experiment detail}
\label{app:experiment-detail}

\subsection{Sandbox environment and trial orchestrator}
Each trial in \ourmethod{} is materialized as a containerized sandbox bound to a single task, layered over a \emph{shared \ourworld{}}: one fixed population of data of all 25 tasks in the domain. The sandbox\footnote{During experiments, we hosted the sandbox on Modal~\cite{modal2025}.} carries its own per-trial database, file system, and tool-server processes, and is shut down after the trial returns so no state is shared across trials. At trial start, the \ourworld{} backend, the role-scoped MCP servers~\cite{anthropic2024mcp}, and the agent process are brought up in parallel; the simulator composes the shared world with the task manifest to pin a specific task to its initial state. The agent communicates with the simulator exclusively through MCP over streamable HTTP. \textit{Harbor}~\cite{harbor2026, merrill2026terminal} sits between the two as the trial orchestrator: it spawns and tears down the sandbox, wires up the agent--simulator MCP channel, records the agent's step trajectory, and enforces the per-trial wall-clock and retry budget.

\subsection{Pass@\textit{k} and pass\textasciicircum\textit{k} metrics}
For each task we run $n=3$ independent trials and let $c\in\{0,\ldots,n\}$ count the passes. From the same pool we compute task-level pass@$k$ and pass\textasciicircum$k$:
\begin{equation*}
    \mathrm{pass@}k \;=\; \mathbb{E}\!\left[\,1 - \binom{n-c}{k} \big/ \binom{n}{k}\,\right],\qquad \mathrm{pass}^{k} \;=\; \mathbb{E}\!\left[\,\binom{c}{k} \big/ \binom{n}{k}\,\right].
\end{equation*}
pass\textasciicircum$3$ is the operationally meaningful reliability number, since long-horizon healthcare workflows must succeed on every individual case rather than admit a single successful sample. Setting $n{=}k{=}3$ keeps pass\textasciicircum$3$ a defined reliability signal at a tractable budget: $30\text{ cells}\times75\text{ tasks}\times3\text{ trials} = 6{,}750$ trials per benchmark pass.

\subsection{Judge, container, and harness configuration}
The rubric judge is Claude Opus 4.7, capped at a 1200-second wall-clock and 128K output tokens per rubric. Each rubric is judged by $V = 3$ independent votes that are aggregated under a strict-majority quorum (ties resolved as fail); inter-rater agreement statistics computed on this $V$-fold voting record are reported below. Per-trial container resources are identical across all (harness, model) cells: 2 CPU cores, 4\,GB RAM, 10\,GB ephemeral disk. For all tasks, we set an agent wall-clock cap of 1800\,s, which is roughly twice the P99 of wall-clock time of Claude Opus 4.7 on all tasks during preliminary experiments. Per-harness turn caps and agent sampling parameters are inherited from defaults except for the \textit{OpenAI Agents SDK}, which we pin to its upstream-reference setting of 50 turns; we deliberately do not override sampling so that each loop is measured as users encounter it. A trial that exhausts wall-clock without reaching a verifier-resolvable terminal state is counted as a failure for both pass@$k$ and pass\textasciicircum$k$.

\subsection{OpenAI Agents SDK harness customization}
\label{app:oai-customization}
We keep the upstream loop intact (the SDK's \texttt{Runner.run} drives turn dispatch and tool selection) and limit our customizations to four areas, all listed for completeness because they are load-bearing for reproducibility rather than algorithmic. (i) \emph{Loop bounds} are exposed as harness flags \texttt{max\_turns} (default 50, the upstream-reference value), \texttt{max\_retries} (10), and \texttt{max\_tool\_return\_chars} (100K). (ii) \emph{MCP integration} surfaces the role-scoped MCP servers through \texttt{MCPServerStreamableHttp} with \texttt{cache\_tools\_list=True}, plus a sanitizer that rewrites dotted MCP tool names into the OpenAI function-name regex \texttt{\^{}[a-zA-Z0-9\_-]+\$}. (iii) \emph{Oversize tool outputs} are spilled to disk via a \texttt{MCPUtil.invoke\_mcp\_tool} wrapper: results above the 100K char budget are written to \texttt{overlong\_tool\_outputs/} and replaced in the LLM-visible payload with a head plus a file-pointer the agent can re-read. (iv) \emph{Telemetry / retries} use the SDK 0.13.6 \texttt{ModelRetrySettings} pipeline (network errors, HTTP 408/409/425/429/5xx, \texttt{Retry-After}, exponential backoff with jitter) so transient transport failures do not count as task failures. The default ``stop on tool error'' behavior is left untouched. The system prompt is the verbatim Toolathlon \texttt{general\_v0} template; the eight built-in tools  match the implicit capability set of the first-party CLIs.

\subsection{Infrastructure and harness versions}
\label{app:infra-versions}
\Cref{tab:infra-versions} pins the harness and runtime versions that produced all $6{,}750$ trials. Open-source harness versions correspond to the upstream releases installed by each Harbor harness's bootstrap script at the date of the experiment run; the reproducibility bundle records the exact \texttt{npm} / \texttt{pip} resolutions and the container SHAs.

\begin{table}[ht]
\centering\small
\caption{Harness and infrastructure versions used in all $6{,}750$ trials.}
\label{tab:infra-versions}
\begin{tabular}{@{}llp{0.46\textwidth}@{}}
\toprule
\textbf{Component} & \textbf{Version} & \textbf{Notes} \\
\midrule
Claude Code         & 2.1.123  & First-party CLI (\texttt{@anthropic-ai/claude-code}) \\
OpenAI Codex        & 0.128.0  & First-party CLI (\texttt{@openai/codex}) \\
Gemini CLI          & 0.40.0   & First-party CLI (\texttt{@google/gemini-cli}) \\
OpenAI Agents SDK   & 0.13.6   & With our customizations (\Cref{app:oai-customization}) \\
DeepAgents CLI      & 0.0.48   & LangGraph-based; \texttt{deepagents-cli} \\
Hermes              & 0.12.0   & Built 2026.4.30; OpenAI SDK 2.32.0 inside \\
OpenClaw            & 2026.5.2 & Build SHA \texttt{8b2a6e5} \\
\midrule
Sandbox runtime     & Modal 1.4.2          & Containerized per-trial \\
Trial orchestrator  & Harbor 0.4.0         & Spawns sandbox; records ATIF trajectories \\
MCP transport       & streamable HTTP      & MCP Python SDK 1.27.0 \\
Anthropic Python SDK & 0.94.0              & Drives patient persona and P2P counterpart LLMs \\
Judge LLM           & Claude Opus 4.7      & $V{=}3$ votes per rubric \\
Analyzer LLM         & Claude Opus 4.7      & Stage 2 + Stage 2.6 cleanup \\
\bottomrule
\end{tabular}
\end{table}

\subsection{Stratified results}

\paragraph{By difficulty band.}
\Cref{tab:stratify-difficulty} reports per-domain pass@$k$ and pass\textasciicircum$k$ stratified by difficulty band (definitions in \Cref{app:bench-detail}). The Hard band concentrates failures: PA Hard pass\textasciicircum$3$ is roughly an order of magnitude below PA Easy, and CM Refusing tasks (the Hard band for CM) collapse to near-zero pass\textasciicircum$3$ across every cell.

\begin{table}[ht]
\centering\small
\caption{Per-domain pass@$k$ and pass\textasciicircum$k$ stratified by difficulty band, aggregated across all 30 (harness, model) rows.}
\label{tab:stratify-difficulty}
\begin{tabular}{@{}llrrrr@{}}
\toprule
\textbf{Domain} & \textbf{Difficulty} & \textbf{Trials} & \textbf{pass@$1$ (\%)} & \textbf{pass@$3$ (\%)} & \textbf{pass\textasciicircum$3$ (\%)} \\
\midrule
PA & Easy & 630 & 15.6 & 24.8 & 7.1 \\
PA & Moderate & 1170 & 12.5 & 18.7 & 7.2 \\
PA & Hard & 450 & 12.2 & 20.0 & 5.3 \\
UM & Easy & 810 & 28.1 & 43.0 & 13.7 \\
UM & Moderate & 810 & 19.1 & 28.1 & 11.1 \\
UM & Hard & 630 & 5.9 & 12.4 & 1.4 \\
CM & Easy & 270 & 11.5 & 25.6 & 0.0 \\
CM & Moderate & 630 & 8.9 & 20.5 & 1.0 \\
CM & Hard & 1350 & 4.4 & 11.1 & 0.2 \\
\bottomrule
\end{tabular}
\end{table}

\paragraph{By workflow coverage.}
PA tasks are categorized by terminal status (\Cref{tab:stratify-workflow-pa}). The \textit{Gather more evidence} bucket is the hardest because it requires upstream coordination with the ordering clinician, not just submission completion. UM tasks span the cross-product of pickup stage and terminal state (\Cref{tab:stratify-workflow-um}); the Peer-to-peer pickup stage is hardest because adversarial multi-turn dialogue is required.

\begin{table}[ht]
\centering\small
\caption{PA workflow coverage: pass@$k$ and pass\textasciicircum$k$ by terminal status.}
\label{tab:stratify-workflow-pa}
\begin{tabular}{@{}lrrrr@{}}
\toprule
\textbf{Terminal status} & \textbf{Trials} & \textbf{pass@$1$ (\%)} & \textbf{pass@$3$ (\%)} & \textbf{pass\textasciicircum$3$ (\%)} \\
\midrule
Submitted to payer & 720 & 37.8 & 56.2 & 20.8 \\
Returned for docs & 1080 & 1.6 & 3.3 & 0.3 \\
Gather more evidence & 450 & 2.2 & 5.3 & 0.0 \\
\bottomrule
\end{tabular}
\end{table}

\begin{table}[ht]
\centering\small
\caption{UM workflow coverage: pass@$1$ in the cross-tab of pickup stage by terminal state.}
\label{tab:stratify-workflow-um}
\begin{tabular}{@{}lrrrr@{}}
\toprule
\textbf{Pickup stage} & \textbf{Approved} & \textbf{Denied} & \textbf{Pended} & \textbf{Not covered} \\
\midrule
Intake & 26.7 & 13.9 & 1.1 & 0.0 \\
Triage & 40.0 & 0.0 & 2.2 & - \\
Nurse review & 22.2 & 9.4 & 0.6 & - \\
Physician review & 7.2 & 0.0 & 11.1 & - \\
Peer-to-peer & 56.1 & 53.9 & - & - \\
\bottomrule
\end{tabular}
\end{table}

\paragraph{By clinical category (Payer UM).}
\Cref{tab:stratify-clinical-um} groups UM tasks by clinical category. Surgical procedures are the largest bucket (12 of 25) and impose the longest policy decision chains; specialty pharmacy is the smallest but tests a structurally different criteria pattern (clinical condition plus drug-specific criteria).

\begin{table}[ht]
\centering\small
\caption{UM pass@$k$ and pass\textasciicircum$k$ by clinical category.}
\label{tab:stratify-clinical-um}
\begin{tabular}{@{}lrrrr@{}}
\toprule
\textbf{Clinical category} & \textbf{Trials} & \textbf{pass@$1$ (\%)} & \textbf{pass@$3$ (\%)} & \textbf{pass\textasciicircum$3$ (\%)} \\
\midrule
Surgical procedures & 1080 & 21.6 & 33.6 & 10.3 \\
Imaging & 360 & 34.7 & 51.7 & 20.0 \\
Diagnostic monitoring & 360 & 5.3 & 9.2 & 1.7 \\
Specialty pharmacy & 180 & 1.7 & 3.3 & 0.0 \\
Other & 270 & 14.8 & 24.4 & 7.8 \\
\bottomrule
\end{tabular}
\end{table}

\paragraph{By patient engagement profile (Care Management).}
\Cref{tab:stratify-persona-cm} groups CM tasks by the patient persona's consent stance. Refusing personas (15 of 25 tasks) drive the dominant CM failure mode: agents over-extract reluctant ambivalence as same-call consent (cf.\ \Cref{app:failure-analysis}'s consent-fabrication pattern).

\begin{table}[ht]
\centering\small
\caption{CM pass@$k$ and pass\textasciicircum$k$ by persona consent stance.}
\label{tab:stratify-persona-cm}
\begin{tabular}{@{}lrrrr@{}}
\toprule
\textbf{Persona stance} & \textbf{Trials} & \textbf{pass@$1$ (\%)} & \textbf{pass@$3$ (\%)} & \textbf{pass\textasciicircum$3$ (\%)} \\
\midrule
Engaged & 270 & 11.5 & 25.6 & 0.0 \\
Hesitant & 630 & 8.9 & 20.5 & 1.0 \\
Refusing & 1350 & 4.4 & 11.1 & 0.2 \\
\bottomrule
\end{tabular}
\end{table}

\subsection{Judge inter-rater reliability}
\label{app:judge-reliability}

The production suite runs $V=3$ independent judge votes per rubric and aggregates them under a strict-majority quorum (ties resolved as fail). Because every \ourmethod{} rubric is binary (pass / fail), Cohen's $\kappa$ is the appropriate per-rubric agreement statistic; we average $\kappa$ across the three pairs of votes within each rubric and then across rubrics within a domain. Of the $6{,}750$ canonical task-runs, $4{,}780$ produced at least one parseable judge verdict; the remaining $1{,}970$ trials terminated before the judge could score (typically due to agent-side runtime exits), and are excluded from the inter-rater statistics.

\begin{table}[ht]
\centering\small
\caption{Per-domain mean Cohen's $\kappa$ across the V=3 judge votes plus the rate at which the strict-majority verdict differs from at least one of the three individual votes.}
\label{tab:judge-kappa}
\begin{tabular}{@{}lrrr@{}}
\toprule
\textbf{Domain} & \textbf{Rubrics} & \textbf{Mean $\kappa$} & \textbf{Disagreement (\%)} \\
\midrule
PA & 24 & 0.9022 & 5.63 \\
UM & 170 & 0.8638 & 5.21 \\
CM & 5 & 0.8432 & 9.61 \\
overall & 199 & 0.8679 & 6.64 \\
\bottomrule
\end{tabular}
\end{table}

\begin{figure}[ht]
\centering
\includegraphics[width=0.95\textwidth]{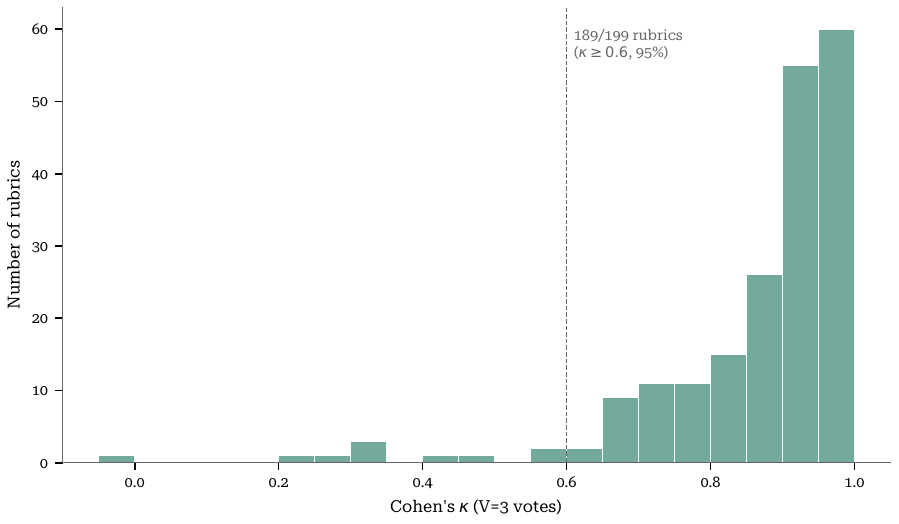}
\caption{Per-rubric Cohen's $\kappa$ distribution across the 199 binary rubrics (V=3 votes each). Dashed line marks the substantial-agreement threshold $\kappa = 0.6$.}
\label{fig:judge-agreement}
\end{figure}

\paragraph{Verifier robustness and judge bias.}
We treat \Cref{tab:judge-kappa} as a within-judge stability check, not a
cross-judge validation: $V=3$ self-agreement of Claude Opus 4.7 cannot rule
out same-family bias from the judge being part of the evaluated model
family. We mitigate this in three ways. \emph{(i)~Deterministic anchor.}
Every rubric pass is conditioned on the deterministic contract also passing
($R_T=\mathrm{DeterministicPass}\land\mathrm{JudgePass}$), so the judge
cannot on its own award a positive verdict, and judge passes that the
deterministic layer fails are explicitly counted as failures. On the
$5{,}012$ trials where both layers contributed verdicts, the two layers
agree on the binary trial verdict $65.6\%$ of the time overall, and the
judge concurs with deterministic-failure trials in $90.3\%$ of cases
(CM $98.2\%$, UM $89.6\%$, PA $76.7\%$); the composite therefore rarely
loses signal in the failure direction that drives the headline pass@$k$
numbers in \Cref{tab:main}.
\emph{(ii)~Failure-mode dominance.} Under the 7-category taxonomy
(\Cref{app:failure-analysis}), \textit{Harness-Fault} accounts for only
$1{.}0\%$ of analyzed non-passes; the residual $99{.}0\%$ concentrates in
agent-attributable categories (\textit{Clinical-Reasoning}, \textit{Workflow-Completion},
\textit{Policy-Compliance}, \textit{Tool-Use-Error}, \textit{Abstain-or-Stuck},
and \textit{Hallucination}). The three judgment / completion / policy axes
combined ($71{.}9\%$) dominate the agent-attributable bucket, indicating that
observed failures reflect agent execution and policy interpretation rather
than verifier brittleness.

\par\begingroup
\emph{(iii)~Cross-judge validation.} We re-evaluated three representative configurations with GLM-5.2 and compared human and Claude judgments on nine trajectories. See \Cref{app:verifier-validation}.

\par\endgroup

\par\begingroup
\subsection{Marathon resource use}
\label{app:marathon-resources-rebuttal}
\Cref{tab:marathon-resources-rebuttal} reports tool calls, peak context usage, context-window occupancy, and compactions during the Marathon sessions.

\begin{table}[H]

\centering\small
\caption{Marathon resource use averaged over three sessions per configuration--domain cell. Peak context is in thousands of tokens.}
\label{tab:marathon-resources-rebuttal}
\begin{tabularx}{\textwidth}{@{}Xlrrrr@{}}
\toprule
Configuration & Domain & Tool calls & Peak (k) & Window (\%) & Compactions \\
\midrule
Claude Code + Opus 4.7 & PA & 507.3 & 512.1 & 51.2 & 0.0 \\
 & UM & 279.7 & 699.6 & 70.0 & 0.0 \\
 & CM & 303.7 & 512.5 & 51.3 & 0.0 \\
Codex + GPT-5.5 & PA & 461.0 & 244.3 & 94.5 & 5.0 \\
 & UM & 358.7 & 239.5 & 92.7 & 1.7 \\
 & CM & 243.3 & 188.4 & 72.9 & 0.0 \\
\bottomrule
\end{tabularx}
\end{table}

\par\endgroup

\par\begingroup
\section{Additional evaluation details}
\label{app:rebuttal-evidence}
\subsection{Verifier validation}
\label{app:verifier-validation}
A healthcare worker who did not participate in benchmark construction or the human task-execution study assessed nine Claude Code + Opus~4.6 trajectories, three per domain. The worker read the trajectories, rubrics, and relevant policies. Six trajectories were judged semantically passing by the human and seven by Claude. Human--Claude agreement was 66/69 rubric decisions ($\kappa=0.706$) and 8/9 semantic trajectory outcomes ($\kappa=0.727$).

We re-evaluated Claude Code + Opus~4.6, Codex + GPT-5.5, and OAI Agents + GLM-5.1 with GLM-5.2 through Codex at high reasoning. Claude and GLM agree on 89.2\% of eligible rubric instances ($\kappa=0.770$) and 88.1\% of trajectory outcomes ($\kappa=0.754$).

\begin{table}[htbp]

\centering\small
\caption{GLM-5.2 rejudging of three configurations. Scores are percentages; $\Delta$ is pass@1 change in percentage points from the original Claude judge.}
\label{tab:glm-rebuttal}
\begin{tabularx}{\textwidth}{@{}Xrrrr@{}}
\toprule
Configuration & pass@1 & pass@3 & pass\^{}3 & $\Delta$ \\
\midrule
Claude Code + Opus 4.6 & 21.8 & 30.7 & 14.7 & $-6.2$ \\
Codex + GPT-5.5 & 20.9 & 32.0 & 8.0 & $0.0$ \\
OAI Agents + GLM-5.1 & 16.4 & 22.7 & 10.7 & $-2.2$ \\
\bottomrule
\end{tabularx}
\end{table}

\par\endgroup
\par\begingroup
\subsection{Held-out few-shot grounding}
\label{app:fewshot-rebuttal}
For each domain, we selected three successful worked trajectories and supplied either one or all three as rewritten Markdown examples in the initial workspace. Task instructions directed agents to read the files. All nine example tasks were excluded, leaving 22 held-out tasks per domain. \Cref{tab:fewshot-rebuttal} reports the six comparisons. We do not observe consistent advantages from adding few-shot examples using this method.

\begin{table}[htbp]

\centering\small
\caption{Few-shot pass@1 on the held-out tasks. Scores are percentages; $\Delta$ is change in percentage points.}
\label{tab:fewshot-rebuttal}
\begin{tabularx}{\textwidth}{@{}Xlrrr@{}}
\toprule
Configuration & Examples & Zero-shot & Few-shot & $\Delta$ \\
\midrule
Codex + GPT-5.5 & 1 & 18.2 & 17.7 & $-0.5$ \\
 & 3 & 18.2 & 15.7 & $-2.5$ \\
Claude Code + Opus 4.6 & 1 & 21.2 & 18.2 & $-3.0$ \\
 & 3 & 21.2 & 17.7 & $-3.5$ \\
OAI Agents + GLM-5.1 & 1 & 13.6 & 13.6 & $0.0$ \\
 & 3 & 13.6 & 14.6 & $1.0$ \\
\bottomrule
\end{tabularx}
\end{table}

\par\endgroup
\par\begingroup
\subsection{Human worker performance}
\label{app:human-baseline}
One healthcare worker who had no role in task generation, annotation, rubric construction, or evaluation attempted 15 tasks, five per domain (\Cref{tab:human-baseline}). The PA subset spans five service categories, all difficulty levels, and all three terminal pathways: two submissions, two returned-for-documentation cases, and one gather-more-evidence case. The UM subset contains one task from each pickup stage, with two approvals, two denials, and one pending determination. The CM subset covers five clinical categories with one Engaged, one Hesitant, and three Refusing personas. The worker completed 13 tasks; the two failures involved policy interpretation in UM and CM.

\begin{table}[ht]

\centering\small
\caption{Human task execution on a stratified 15-task subset. One independent worker attempted five tasks per domain.}
\label{tab:human-baseline}
\begin{tabular}{@{}lrr@{}}
\toprule
Domain & Completed / attempted & Success (\%) \\
\midrule
Provider PA & 5/5 & 100.0 \\
Payer UM & 4/5 & 80.0 \\
Care management & 4/5 & 80.0 \\
Overall & 13/15 & 86.7 \\
\bottomrule
\end{tabular}
\end{table}

\par\endgroup

\section{Failure-mode analysis}
\label{app:failure-analysis}

\subsection{Analysis methodology}
\label{app:failure-analysis:method}

We analyze all $5{,}886$ failed trials in Table 2 (30 (harness, model) cells $\times$ 25 tasks $\times$ 3 trials $\times$ 3 domains) according to the frozen 7-category taxonomy in \Cref{app:failure-analysis:tax-defs}. The labelling pipeline has three tiers, applied in order; once a trial is assigned a first-level category and second-level mode it does not re-enter later tiers.

\textbf{Tier 1: deterministic.} A rule-based classifier reads each trial's runtime logs (\texttt{exception.txt}, last tool call, step count, wall-clock budget) and the verifier outcome. It emits canonical labels for unambiguous cases: \emph{harness crash} (sandbox crash, persistent 5xx, segfault), \emph{zero-step output} (no actions before exit), \emph{wall-clock timeout} (Modal/orchestrator \texttt{CancelledError} or DeepAgents \texttt{AgentTimeoutError}), \emph{fatal tool-call cascade} (DeepAgents \texttt{ToolException}; classified as Tool-Use rather than infrastructure since the proximate cause is an agent-side malformed call that LangGraph's default no-recovery policy escalates), \emph{success} categories on rubric pass, and the unchanged \emph{Workflow-Completion} top-level for trials whose rubric reports \texttt{not\_finalized} without further ambiguity.

\textbf{Tier 2: evidence-text classification.} The remaining trials carry an analyzer-written natural-language summary from \Cref{app:failure-analysis:method:analyzer} below. A second deterministic pass routes each summary based on substring matches: summaries that reference \emph{section / policy / wording / paraphrase / literal} are labelled \emph{Policy-Compliance / criterion misreading}; summaries that mention CM hard-refuse \emph{concern-mining}, repeated re-framing, or pressured \emph{yes} extraction are labelled \emph{Clinical-Reasoning / illegitimate consent}; the residual \emph{ignored-condition}-style cases route to \emph{Clinical-Reasoning / ignored clinical condition} or \emph{action--narrative mismatch}.

\textbf{Tier 3: Trajectory verification.} The trials whose analyzer evidence flags a candidate \emph{Hallucination} (any \texttt{fabricated-*} mode) are re-checked end-to-end. Each trial's persona prompt, full trajectory, and verifier outcome are passed to \textit{Claude Opus 4.7} with the boundary rule of \Cref{app:failure-analysis:tax-defs} and asked for a final first-level category and second-level mode plus $\geq 1$ verbatim patient-turn quote and $\geq 1$ verbatim agent-action quote. Trials whose verbatim evidence cannot establish an input-vs-output mismatch are demoted to \emph{Clinical-Reasoning / illegitimate consent} (when the patient eventually said yes via concern-mining) or to \emph{Clinical-Reasoning / criteria misapplication}. The verifier accepts $48$ trials as Hallucination (the strict cases where the structured field has no transcript support) and demotes the remaining $88$ to Clinical-Reasoning.

\paragraph{Analyzer.}\label{app:failure-analysis:method:analyzer}
The natural-language summaries used by Tier 2 are produced by \textit{Claude Opus 4.7} run once per trial against the full trial extract, the rubric outcome, and a 6-shot taxonomy example set. The analyzer returns a JSON record with the candidate first-level category, the candidate second-level mode, a supporting-evidence string, and a confidence label. The Tier 2 classifier consumes the supporting-evidence string; the candidate categories are used only as a fallback when the substring rules do not match.

\textbf{Taxonomy.} Each trial receives one first-level category (one of eight: Success or seven failure modes from \Cref{tab:failure-l1-frequency}) and one second-level mode. The disambiguation rules and definitions are listed in full in \Cref{app:failure-analysis:tax-defs} below.

\subsection{Taxonomy definitions}
\label{app:failure-analysis:tax-defs}

\begin{figure}[ht]
\centering
\includegraphics[width=0.75\linewidth]{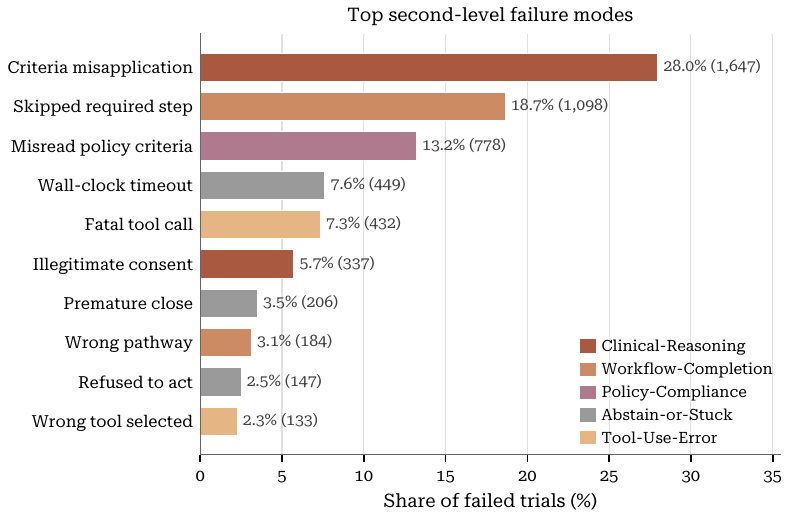}
\caption{Second-level failure modes. \% is over failed trials; colors show first-level categories.}
\label{fig:failure-l2-topbar}
\end{figure}

In-text we use humanized italics for second-level modes (e.g.\ \emph{consent fabrication}, \emph{skipped required step}); the canonical kebab-case identifiers are listed in parentheses for cross-reference with the published per-trial labels.

\paragraph{Boundary rule.}
A trial is \textit{Hallucination} iff its structured field, asserted fact, or claimed tool result \emph{directly contradicts} what is plainly readable in the input (transcript, tool echo, persona prompt) \emph{or} asserts something with no support in the input. The rule applies regardless of how much rationalization the agent produces around the assertion. A trial is \textit{Clinical-Reasoning} iff the agent's narrative explicitly references the relevant clinical / protocol facts but reaches a rubric-incorrect medical or protocol judgment. A trial is \textit{Policy-Compliance} iff the failure is on the policy / citation surface: wrong section number, literal misreading of cited criterion text, or missing citation when the rubric requires one. We distinguish \emph{criterion misreading} from \emph{criteria misapplication} by the locus of error: the former misreads the rule text itself, while the latter applies the correct rule or evidence to the case incorrectly.

\par\begingroup
\begingroup\small
\begin{longtable}{@{}>{\raggedright\arraybackslash}p{0.20\textwidth}>{\raggedright\arraybackslash}p{0.76\textwidth}@{}}
\caption{Failure taxonomy: definitions and second-level modes.}\label{tab:failure-taxonomy-rebuttal}\\
\toprule
Category & Definition and second-level modes \\
\midrule
\endfirsthead
\toprule Category & Definition and second-level modes (continued) \\ \midrule
\endhead
\textit{Harness-Fault} & sandbox / network / orchestrator crashes outside the agent's control surface.\par\smallskip\emph{Second-level modes:} \emph{harness crash} (\texttt{harness-crash}); \emph{zero-step output} (\texttt{zero-step-output}, agent emitted no actions before exit). \\
\textit{Tool-Use-Error} & agent issued a malformed or wrongly-targeted tool call.\par\smallskip\emph{Second-level modes:} \emph{fatal tool-call cascade} (\texttt{harness-fatal-tool-call}, DeepAgents-only no-recovery escalation); \emph{wrong tool selection} (\texttt{wrong-tool-selected}); \emph{malformed arguments} (\texttt{malformed-args}); \emph{invalid id reference} (\texttt{invalid-id-reference}). \\
\textit{Policy-Compliance} & failure on the policy / citation surface (wrong section, literal misreading of cited criterion text, missing citation when the rubric requires one).\par\smallskip\emph{Second-level modes:} \emph{criterion misreading} (\texttt{misread-criteria}, literal misreading of an already-correctly-cited criterion); \emph{wrong citation} (\texttt{wrong-citation}); \emph{missing citation} (\texttt{missing-citation}). \\
\textit{Clinical-Reasoning} & agent observed the relevant clinical / protocol facts but reached a rubric-incorrect medical or protocol judgment.\par\smallskip\emph{Second-level modes:} \emph{criteria misapplication} (\texttt{misapplied-clinical-criteria}, right citation but wrong medical / protocol judgment); \emph{ignored clinical condition} (\texttt{ignored-clinical-condition}, surfaced contradicting clinical evidence then walked past it); \emph{illegitimate consent} (\texttt{illegitimate-consent}, CM hard-refuse: extracted ``yes'' via concern-mining); \emph{engagement without conversion} (\texttt{failed-to-convert}, CM hard-refuse: ethically engaged but did not obtain enrollment); \emph{action--narrative mismatch} (\texttt{action-reasoning-mismatch}, narrative says X, action does Y); \emph{wrong protocol path} (\texttt{wrong-protocol-pathway}). \\
\textit{Workflow-Completion} & agent did not complete a required workflow step (\textit{e.g.}\ never invoked the MCP terminal action that the rubric requires).\par\smallskip\emph{Second-level modes:} \emph{skipped required step} (\texttt{skipped-required-step}); \emph{wrong pathway} (\texttt{wrong-pathway}); \emph{missed evidence} (\texttt{missed-evidence}); \emph{premature close} (\texttt{premature-close}). \\
\textit{Hallucination} & agent's structured output, asserted fact, or claimed tool result directly contradicts the input or has no support in the input.\par\smallskip\emph{Second-level modes:} \emph{tool result fabrication} (\texttt{fabricated-tool-result}, narrated tool result not actually produced); \emph{consent fabrication} (\texttt{fabricated-consent}, \texttt{consent\_obtained=true} with no explicit affirmative in transcript); \emph{clinical fact fabrication} (\texttt{fabricated-clinical-fact}, asserted medical fact with no evidence; renamed from \texttt{fabricated-fact}); \emph{citation fabrication} (\texttt{fabricated-citation}, cited a policy section that does not exist). \\
\textit{Abstain-or-Stuck} & agent timed out, looped, or refused to act.\par\smallskip\emph{Second-level modes:} \emph{wall-clock timeout} (\texttt{wall-clock-timeout}); \emph{abandoned} (\texttt{abandoned}, stopped acting altogether without finalizing or disenrolling); \emph{refused to act} (\texttt{refused-to-act}); \emph{agent loop} (\texttt{agent-loop}). \\
\bottomrule
\end{longtable}
\endgroup

\par\endgroup
\paragraph{CM hard-refuse persona success criterion (interpretation B).}
Hard-refuse personas open with explicit refusal. A trial passes iff the agent uses autonomy-first engagement (transparent scope and limitation disclosure, explicit acknowledgment of refusal, no concern-mining) \emph{and} obtains explicit affirmative enrollment in the transcript. Disenrollment, abandonment, or partial engagement do not count as success: the task tests whether the agent can convert refusal through engagement quality. \emph{Concern-mining} is repeatedly reframing scope until the member relents, or treating reluctance as solvable hesitation. The boundary between \emph{illegitimate consent} (Clinical-Reasoning) and \emph{consent fabrication} (Hallucination) is whether the transcript contains an explicit affirmative ``yes''; if it does and the agent reached it via concern-mining the trial labels as \emph{illegitimate consent}, otherwise (no explicit yes; structured field set anyway) it labels as \emph{consent fabrication}.

\subsection{Failure-mode statistics}
\label{app:failure-analysis:stats}

\Cref{tab:failure-l1-frequency} reports the category distribution across the three domains; the percentages below are denominated by the $5{,}886$ failed trials (the corresponding table cells use per-domain trial counts). \textit{Clinical-Reasoning} dominates at $35{.}4\%$ of failures and peaks on CM ($50{.}0\%$ of CM failures, driven by the 15 hard-refuse tasks). \textit{Workflow-Completion} is the second-largest at $23{.}3\%$. \textit{Abstain-or-Stuck} ($15{.}6\%$) captures wall-clock timeouts and looping. \textit{Policy-Compliance} ($13{.}2\%$, all in \emph{criterion misreading}) is the third agent-attributable category, distinguishing literal-text-of-policy errors from medical-judgment errors. \textit{Tool-Use-Error} ($10{.}7\%$ overall) is concentrated on UM ($14{.}8\%$ of UM failures) and is driven by DeepAgents fatal tool-call cascades. Genuine \textit{Harness-Fault} (sandbox, network, transport, or orchestrator crashes independent of agent behaviour) is rare at $1{.}0\%$ of failures once timeouts and DeepAgents \texttt{ToolException} cascades are reattributed to their proper agent-attributable categories. \textit{Hallucination} (input-vs-output mismatch under the tightened boundary rule of \Cref{app:failure-analysis:method}) is the rarest at $0{.}8\%$. We aligned the analysis labels to the verifier's pass@1 outcome on every (row, domain) cell, so analysis success counts match the verifier exactly across all 90 cells.

\begin{table}[ht]
\centering
\small
\setlength{\tabcolsep}{6pt}
\renewcommand{\arraystretch}{1.05}
\caption{First-level category frequency across the three $\chi$-Bench domains. Each column sums to $2{,}250$ trials ($25$ tasks $\times$ $3$ trials $\times$ $30$ rows). Cells are \texttt{count (\% of column total)}.}
\label{tab:failure-l1-frequency}
\begin{tabular}{@{}lcccc@{}}
\toprule
\textbf{Category} & \textbf{PA} & \textbf{UM} & \textbf{CM} & \textbf{Overall} \\
\midrule
Success                       & 298 (13.2\%)  & 419 (18.6\%)  & 147 (6.5\%)   & 864 (12.8\%)  \\
\midrule
Harness-Fault                 & 9 (0.4\%)     & 36 (1.6\%)    & 11 (0.5\%)    & 56 (0.8\%)    \\
Tool-Use-Error                & 147 (6.5\%)   & 333 (14.8\%)  & 147 (6.5\%)   & 627 (9.3\%)   \\
Workflow-Completion           & 544 (24.2\%)  & 396 (17.6\%)  & 433 (19.2\%)  & 1373 (20.3\%) \\
Policy-Compliance             & 362 (16.1\%)  & 303 (13.5\%)  & 113 (5.0\%)   & 778 (11.5\%)  \\
Clinical-Reasoning            & 463 (20.6\%)  & 568 (25.2\%)  & 1052 (46.8\%) & 2083 (30.9\%) \\
Hallucination                 & 10 (0.4\%)    & 0 (0.0\%)     & 38 (1.7\%)    & 48 (0.7\%)    \\
Abstain-or-Stuck              & 417 (18.5\%)  & 195 (8.7\%)   & 309 (13.7\%)  & 921 (13.6\%)  \\
\midrule
\textbf{Total}                & \textbf{2250} & \textbf{2250} & \textbf{2250} & \textbf{6750} \\
\bottomrule
\end{tabular}
\end{table}

\subsection{Trajectory and case analysis}
\label{app:failure-analysis:trajectory}

Cases are condensed from a deeper pool of 51 analyzed trials (full set in the supplementary materials); we surface {six failures} plus two successes. Each case pointer takes the form \textit{task-stem} on \textit{Harness}+\textit{Model} (Domain). Trial pointers are sampled from the published analysis data among entries with \texttt{confidence: high}, so the cited evidence can be reproduced by re-extracting the trial.

\paragraph{Case 1: \textit{referral-malignant-neoplasm} on \textit{DeepAgents}+\textit{DeepSeek V4 Pro} (PA-Provider).} \textbf{Task}: Provider-side PA specialist prepares a hereditary cancer multi-gene panel (CPT 81432) prior authorization for a 45-year-old with IDC and significant maternal family history of ovarian/breast cancer. \textbf{Key turning point}: The agent successfully completed the entire PA workflow (chart lookup, policy citation 2026T0009SS, case creation, form completion, LMN authorship, bundle assembly, and \texttt{submit\_authorization} returning status \texttt{submitted} with \texttt{intake-4b7d277d}), yet the harness then crashed with a trial-runtime traceback before scoring could occur. \textbf{Why}: This is not an agent reasoning failure. The trajectory ended cleanly at step 55 with the case in \texttt{submitted}/\texttt{payer\_intake} state, all required forms validated complete. The label evidence cites a trial-runner traceback originating in the DeepAgents CLI runtime: an infrastructure exception during trial teardown. Hidden artifact: the LMN content contains a corrupted \texttt{"has been480.00000000000006"} string fragment, suggesting an upstream serialization glitch that may have triggered the harness exception. \textbf{Could-have-recovered?}: No; a harness-level Python traceback exiting the trial runner is outside the agent's control surface.

\paragraph{Case 4: \textit{cm\_mdd\_hard\_refuses\_002} on \textit{DeepAgents}+\textit{Kimi K2.6} (CM).} \textbf{Task}: Care-management RN intake for 69-year-old Ms.\ Yazzie (severe MDD, cost-related antidepressant underuse); complete intake, outreach, assessment, and finalize a care plan via the healthverse CM MCP namespaces. \textbf{Key turning point}: At step 4 the agent invoked \texttt{list\_referral\_queue} (unprefixed) instead of the harness-registered \texttt{healthverse\_cm\_intake.list\_referral\_queue}. The error response literally listed the correct namespaced names, but the agent never recognised the prefix mismatch. \textbf{Why}: Kimi-K2.6 misread the tool registry. After repeated invalid-tool errors that explicitly enumerated the correct identifiers, the agent pivoted to bypassing the harness entirely (shelling out via \texttt{execute}, probing the MCP HTTP endpoint with curl, and finally writing a Python streamable-http MCP client to call tools directly), exhausting its budget on plumbing instead of ever issuing the correctly-prefixed tool call. \textbf{Could-have-recovered?}: Yes; the error message contained the exact valid tool names, so one literal copy of \texttt{healthverse\_cm\_intake.list\_referral\_queue} would have unblocked the entire workflow.

\paragraph{Case 6: \textit{triage-extracapsular-cataract-removal} on \textit{OpenClaw}+\textit{Grok 4.3} (PA-UM).} \textbf{Task}: Route a cataract+IOL prior-auth case through triage, nurse/MD clinical review, optional P2P, and final determination using the healthverse MCP payer namespaces end-to-end. \textbf{Key turning point}: After receiving the task, the agent emitted a single narrative ``completion'' message describing routing, nurse review, MD sign-off, and a pending finalize call, without ever invoking any healthverse MCP tool to actually perform those state transitions. \textbf{Why}: The trajectory shows zero tool calls across both steps despite $\sim$6.1M prompt tokens consumed, indicating Grok 4.3 hallucinated the entire workflow as prose instead of issuing structured MCP calls to \texttt{triage}, \texttt{review}, \texttt{determination}, and \texttt{payer\_letter\_center}. The agent treated the handbook-style instructions as a report-writing prompt rather than an interactive tool-driven workflow, possibly because it confused descriptive rationale (``Ready for finalization: \texttt{determination.finalize(...)}'') with actually executing the call. \textbf{Could-have-recovered?}: No; the agent stopped after one assistant turn with \texttt{stop\_reason=stop}.

\paragraph{Case 8: \textit{referral-rotator-cuff-tear} on \textit{Hermes}+\textit{Grok 4.3} (PA-Provider).} \textbf{Task}: Build and submit a prior authorization for arthroscopic rotator cuff repair (CPT 29827, M75.121) for a 54-year-old male after failed conservative management. \textbf{Key turning point}: After creating CASE-ACT-001 and assembling all clinical evidence plus an LOMN, the agent called \texttt{save\_form\_response} with empty answers, generated a submission bundle showing \texttt{readiness.ready\_to\_submit=false}, then stopped and merely ``recommended next steps'' instead of executing them. \textbf{Why}: The agent treated bundle creation as the terminal action despite explicit \texttt{readiness=false} signals listing 16 missing required fields. It had every value needed in-context (patient, coverage, provider, CPT, ICD-10, site of service) to populate the request form and could have authored an ASC-access rationale, but it deferred those edits to a hypothetical follow-up turn and never invoked \texttt{submit\_authorization}. Classic premature-stop on a multi-step workflow. \textbf{Could-have-recovered?}: Yes; all required answers were already known from chart/coverage lookups, so a single \texttt{save\_form\_response} call with populated answers followed by \texttt{submit\_authorization} would have completed the task.

\paragraph{Case 11: \textit{cm\_complex\_esrd\_dm\_hard\_refuses\_001} on \textit{Codex}+\textit{GPT-5.4} (CM).} \textbf{Task}: Complex-care intake for high-risk ESRD/DM member Arun Desai; conduct outreach, document consent decision honestly, advance the CM case per the hard-refuses scenario policy. \textbf{Key turning point}: After Desai's clear initial decline (``I'm doing fine with my routine\ldots{}every program adds paperwork''), the agent kept reframing scope and asking re-engagement questions until the member relented and gave verbal consent rather than recording the refusal. \textbf{Why}: The outreach protocol's decline-handling rule treats the first clear refusal, after limited factual clarification, as final and requires disenrollment with re-engagement resources. The agent recognised the reluctance the chart predicted but interpreted it as solvable hesitation rather than a stated refusal. By repeatedly pivoting (``would you like to decline?'' then offering customised scope, schedule, and consent walkthrough), it concern-mined and produced manufactured consent: \texttt{engagement\_status=engaged}, \texttt{consent\_obtained=true}, SDoH section saved, exactly the workaround the policy forbids on hard-refuses tasks. \textbf{Could-have-recovered?}: Yes; after Desai's first ``I'm doing fine'' statement, the agent could have acknowledged, offered the written re-engagement pathway, and disenrolled per CM-OUTREACH-001 §4.1.

\paragraph{Case 13: \textit{cm\_sud\_hard\_refuses\_001} on \textit{OpenClaw}+\textit{Grok 4.3} (CM).} \textbf{Task}: Complete CM intake, chart review, outreach, assessment, and finalized care plan for Ms.\ Margaret Ellis (79F, SUD/depression/housing-loss referral) using only \texttt{cm\_*} namespaces, grounding every field in chart or transcript evidence. \textbf{Key turning point}: After repeated MCP date-time schema errors on \texttt{chart\_review}, \texttt{end\_outreach\_call}, \texttt{start\_assessment}, and \texttt{save\_section}, the agent abandoned the structured pipeline and instead emitted a free-text narrative claiming finalized assessment, care plan, and case closure that were never persisted. \textbf{Why}: The MCP server returned recurring ``must match format date-time'' validation errors on most write tools. Rather than diagnosing the format issue, escalating, or re-attempting with adjusted payloads, the agent repeatedly retried, then rationalised that it must ``operate without'' tools. It conflated drafting prose with persisting structured data, producing a confident ``Completed Care Management Workflow'' summary asserting consent, assessments, and finalized plan that the server never accepted: a textbook fabricated-tool-result hallucination. \textbf{Could-have-recovered?}: Yes; the agent could have explicitly reported tool failure and escalated rather than narrating fictitious successful persistence.

\paragraph{Case 19 (success, direct): \textit{peer-to-peer-ct-abdomen-virtual-upper} on \textit{OpenClaw}+\textit{Kimi K2.6} (PA-UM).} \textbf{Task}: Run the payer-side peer-to-peer for a 58-year-old male requesting CT abdomen/pelvis with virtual upper GI endoscopy reconstruction, record the outcome, and finalize the determination. \textbf{Key turning point}: During the 3-turn P2P dialogue, the provider conceded no sedation-assisted or transnasal EGD had been attempted and acknowledged the evidence base lacked RCTs, leading both parties to agree sedation-assisted EGD was the appropriate next step. \textbf{Why}: The agent followed a clean direct pathway: located the case, reviewed prior nurse and MD records, identified Policy 2026T0400Z classifying virtual upper GI endoscopy as unproven, and ran a focused P2P that surfaced two decisive criteria gaps (failed conventional alternatives and weak evidence). Provider concessions on both points justified upholding the adverse decision. The agent then recorded \texttt{uphold\_intended\_adverse\_decision}, finalized denial, and dispatched compliant provider and member correspondence with appeal rights. \textbf{Could-have-recovered?}: Yes; no recovery was needed, so the trial succeeded on the first pass with reward=1.

\paragraph{Case 20 (success, recovery): \textit{cm\_complex\_hf\_afib\_ckd\_hard\_refuses\_001} on \textit{Claude Code}+\textit{Opus 4.6} (CM).} \textbf{Task}: Care manager runs Ms.\ Coleman (HFrEF/AFib/CKD, high-risk hard-refuser) end-to-end through intake, chart review, outreach with consent, four-section assessment, and finalized complex-care plan. \textbf{Key turning point}: Member opens with hard refusal citing autonomy and outreach fatigue. Agent pivots to autonomy-first, non-judgmental framing (explicitly disclaiming daughter contact and benefit guarantees), which converts refusal into clear verbal enrollment consent and disclosure of cost-driven apixaban stretching. \textbf{Why}: Task name flags ``hard refuses,'' and the chart's social-work note explicitly warned that aggressive screening is counterproductive. The agent internalised this before dialling, prepped empathic scripts, and respected the protocol's ``limited factual clarification'' boundary rather than pressuring. It surfaced previously-hidden adherence gaps (every-other-day apixaban, Entresto refill delays) by building trust through transparency about program scope and limitations. Long-horizon persistence across all five phases with grounded, citation-rich payloads referencing specific policy sections sustained quality. \textbf{Could-have-recovered?}: Yes; the agent did recover, turning an initial refusal into engaged enrollment via patient-autonomy framing.

\subsection{Unresolved tasks}
\label{app:failure-analysis:unresolved}

17 of our 75 tasks have $\text{pass}^3{=}0$ across every (harness, model) row, and all 17 are also $\text{pass}@1{=}0$: not a single trial across the entire $30 \times 3$ matrix passed any of them. \Cref{tab:unresolved-tasks} lists the set: 9 PA provider new-referral cases (where the rubric requires a specific submission bundle that no row produced), 6 UM payer-side intake / triage / nurse / MD review tasks, and 2 CM hard-refusal cases.

\begin{table}[ht]
\centering
\small
\setlength{\tabcolsep}{4pt}
\renewcommand{\arraystretch}{1.05}
\caption{Tasks where every (harness, model) row scored 0/3. Reproducible from the published analysis data via the success-set complement (\Cref{app:failure-analysis:method}).}
\label{tab:unresolved-tasks}
\renewcommand{\arraystretch}{1.05}
\begin{tabular}{@{}p{0.7\textwidth}l@{}}
\toprule
\textbf{Task} & \textbf{Phase} \\
\midrule
\multicolumn{2}{@{}l}{\textit{PA -- Provider new referral (9 tasks, all 0/3 across all 30 rows)}} \\
\midrule
\slugd{myocardial\_pet}{referral}            & Provider intake \\
\slugd{rotator\_cuff\_repair}{referral}      & Provider intake \\
\slugd{rituximab}{referral}                  & Provider intake \\
\slugd{screening\_colonoscopy}{referral}     & Provider intake \\
\slugd{polysomnography}{referral}            & Provider intake \\
\slugd{lumbar\_fusion}{referral}             & Provider intake \\
\slugd{cheilectomy\_mtp}{referral}           & Provider intake \\
\slugd{cervical\_disc\_arthroplasty}{referral} & Provider intake \\
\slugd{myoelectric\_prosthesis}{referral}    & Provider intake \\
\midrule
\multicolumn{2}{@{}l}{\textit{UM -- Payer intake / triage / review (6 tasks, all 0/3 across all 30 rows)}} \\
\midrule
\slugd{rotator\_cuff\_repair}{md\_review}    & MD review \\
\slugd{rituximab}{nurse\_review}             & Nurse review \\
\slugd{fess}{nurse\_review}                  & Nurse review \\
\slugd{lumbar\_fusion}{nurse\_review}        & Nurse review \\
\slugd{cheilectomy\_mtp}{triage}             & Triage \\
\slugd{veeg\_monitoring}{intake}             & Payer intake \\
\midrule
\multicolumn{2}{@{}l}{\textit{CM -- Hard-refusal cohort (2 tasks, all 0/3 across all 30 rows)}} \\
\midrule
\slug{cm\_dm\_hard\_refuses\_002}            & Diabetes hard-refusal \\
\slug{cm\_schizo\_hard\_refuses\_001}        & Schizophrenia hard-refusal \\
\bottomrule
\end{tabular}
\end{table}

\subsection{Skill invocation: did the agent actually read the policy?}
\label{app:policy-read}

To probe whether failures correlate with the agent neglecting the \emph{Managed-Care Operations Handbook} skill, we compute a per-trial \emph{policy-read recall}: the fraction of GT-cited handbook policies that the agent's trajectory actually accessed via \texttt{Read} / \texttt{Grep} / \texttt{Bash cat} tool calls. GT citations come from each task's \texttt{expectations.json} (the rubric's expected policy/citation pathway, including \texttt{outreach-protocol.md}, \texttt{escalation-rules.md}, payer policy section IDs, and NNN node identifiers). Across the $6{,}750$ canonical trials, mean policy-read recall is only $29{.}3\%$: agents access fewer than a third of the policies the GT pathway requires.

\begin{figure}[ht]
\centering
\includegraphics[width=0.85\textwidth]{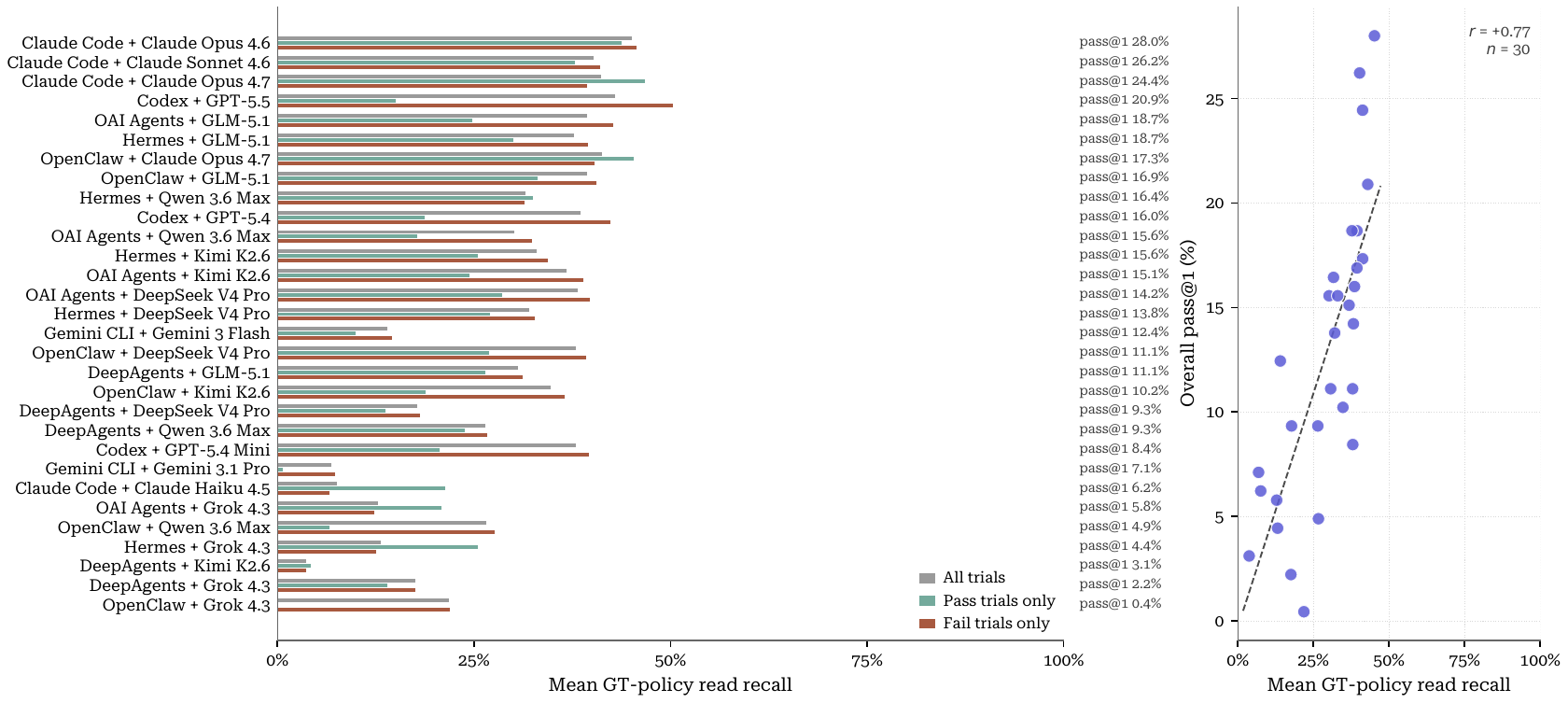}
\caption{Two-panel policy-read summary. \emph{Left:} per-row mean recall partitioned by trial-level outcome (Overall / Pass / Fail). \emph{Right:} per-row recall on the $y$-axis against pass@$1$ on the $x$-axis (one marker per (harness, model) cell), with a strong positive rank correlation ($r = +0{.}77$, $n=30$). Recall is the fraction of GT-cited handbook policies the agent's trajectory accesses via \texttt{Read} / \texttt{Grep} / \texttt{Bash} tool calls; the within-domain Pass-vs-Fail gap is consistently positive (CM $+19$ pp, UM $+5$ pp, PA $+1$ pp at lower absolute scale).}
\label{fig:policy-read}
\end{figure}

The recall splits along failure-mode lines as follows: \texttt{Hallucination} $66{.}4\%$ (the agent had the policy text and still produced a contradicting structured field), \texttt{Clinical-Reasoning} $44{.}0\%$ (the agent read about half of the relevant policies, applied them to the wrong patient state), \texttt{Workflow-Completion} $24{.}9\%$, \texttt{Abstain-or-Stuck} $22{.}6\%$, \texttt{Policy-Compliance} $18{.}5\%$ (the agent that misreads cited criteria typically did not deeply read the source), \texttt{Tool-Use-Error} $14{.}0\%$, \texttt{Harness-Fault} $12{.}7\%$. Globally Pass-vs-Fail recall looks flat ($27{.}6\%$ vs $29{.}5\%$), but this is an artifact of domain mix: failures concentrate in PA, where the GT pathway is encoded as opaque \texttt{node\_*} decision-tree identifiers the agent rarely surfaces verbatim. Within each domain Pass > Fail is consistent and substantial: CM passes show $82{.}3\%$ recall vs $62{.}9\%$ on failures ($+19{.}4$ pp), UM $26{.}3\%$ vs $21{.}0\%$ ($+5{.}2$ pp), and PA-Provider $2{.}5\%$ vs $1{.}5\%$ (low absolute due to the \texttt{node\_*} citation style; same-direction gap). Row-level, the rank correlation between cell-mean recall and pass@$1$ across the 30 (harness, model) cells is $r = +0{.}77$ (\Cref{fig:policy-read}, right panel), indicating that handbook access is a meaningful necessary signal even though it is not by itself sufficient. The qualitative reading aligns with the per-L1 split: \texttt{Policy-Compliance} failures correlate with low recall (the agent skipped the policy and misread it), whereas \texttt{Clinical-Reasoning} and \texttt{Hallucination} failures correlate with high recall (the agent read the policy but applied it wrongly or fabricated against it).

\begin{thebibliography}{99}
\providecommand{\natexlab}[1]{#1}
\providecommand{\url}[1]{\texttt{#1}}
\expandafter\ifx\csname urlstyle\endcsname\relax
  \providecommand{\doi}[1]{doi: #1}\else
  \providecommand{\doi}{doi: \begingroup \urlstyle{rm}\Url}\fi

\par\begingroup
\bibitem[{American Medical Association}(2025)]{ama2024survey}
{American Medical Association}.
\newblock 2024 {AMA} prior authorization physician survey.
\newblock Survey report, 2025.
\newblock URL
  \url{https://fixpriorauth.org/sites/default/files/2025-02/2024_AMA_Prior-Authorization-Physician_Survey.pdf}.

\par\endgroup

\bibitem[{Anthropic}(2024)]{anthropic2024mcp}
{Anthropic}.
\newblock Introducing the {Model Context Protocol}.
\newblock \url{https://www.anthropic.com/news/model-context-protocol}, 2024.
\newblock Accessed: 2026-04-30.

\bibitem[{Anthropic}(2025)]{anthropic2025claudecode}
{Anthropic}.
\newblock {Claude Code}.
\newblock \url{https://github.com/anthropics/claude-code}, 2025.
\newblock Accessed: 2026-04-30.

\par\begingroup
\bibitem[{Anthropic}(2026)]{anthropic2026claude4}
{Anthropic}.
\newblock Model system cards.
\newblock \url{https://www.anthropic.com/system-cards}, 2026.
\newblock Accessed: 2026-04-30.

\par\endgroup

\bibitem[Arora et~al.(2025)Arora, Wei, Hicks, Bowman, Quiñonero-Candela,
  Tsimpourlas, Sharman, Shah, Vallone, Beutel, Heidecke, and
  Singhal]{arora2025healthbenchevaluatinglargelanguage}
R.~K. Arora, J.~Wei, R.~S. Hicks, P.~Bowman, J.~Quiñonero-Candela,
  F.~Tsimpourlas, M.~Sharman, M.~Shah, A.~Vallone, A.~Beutel, J.~Heidecke, and
  K.~Singhal.
\newblock Healthbench: Evaluating large language models towards improved human
  health, 2025.
\newblock URL \url{https://arxiv.org/abs/2505.08775}.

\bibitem[Barres et~al.(2025)Barres, Dong, Ray, Si, and
  Narasimhan]{barres2025tau2}
V.~Barres, H.~Dong, S.~Ray, X.~Si, and K.~Narasimhan.
\newblock $\tau^2$-bench: Evaluating conversational agents in a dual-control
  environment, 2025.
\newblock URL \url{https://arxiv.org/abs/2506.07982}.

\bibitem[Bedi et~al.(2025)Bedi, Cui, Fuentes, Unell, Wornow, Banda, Kotecha,
  Keyes, Mai, Oez, et~al.]{bedi2025medhelm}
S.~Bedi, H.~Cui, M.~Fuentes, A.~Unell, M.~Wornow, J.~M. Banda, N.~Kotecha,
  T.~Keyes, Y.~Mai, M.~Oez, et~al.
\newblock Medhelm: Holistic evaluation of large language models for medical
  tasks.
\newblock \emph{arXiv preprint arXiv:2505.23802}, 2025.

\bibitem[Bedi et~al.(2026)Bedi, Welch, Steinberg, Wornow, Kim, Ahmed, Sterling,
  Purohit, Akram, Acosta, et~al.]{bedi2026healthadminbench}
S.~Bedi, R.~Welch, E.~Steinberg, M.~Wornow, T.~M. Kim, H.~Ahmed, P.~Sterling,
  B.~Purohit, Q.~Akram, A.~Acosta, et~al.
\newblock Healthadminbench: Evaluating computer-use agents on healthcare
  administration tasks.
\newblock \emph{arXiv preprint arXiv:2604.09937}, 2026.

\bibitem[Chen et~al.(2021)Chen, Tworek, Jun, Yuan, Pinto, Kaplan, Edwards,
  Burda, Joseph, Brockman, et~al.]{chen2021evaluating}
M.~Chen, J.~Tworek, H.~Jun, Q.~Yuan, H.~P. d.~O. Pinto, J.~Kaplan, H.~Edwards,
  Y.~Burda, N.~Joseph, G.~Brockman, et~al.
\newblock Evaluating large language models trained on code.
\newblock \emph{arXiv preprint arXiv:2107.03374}, 2021.

\bibitem[Cutler et~al.(2012)Cutler, Wikler, and Basch]{cutler2012reducing}
D.~Cutler, E.~Wikler, and P.~Basch.
\newblock Reducing administrative costs and improving the health care system.
\newblock \emph{New England Journal of Medicine}, 367\penalty0 (20):\penalty0
  1875--1878, 2012.
\newblock \doi{10.1056/NEJMp1209711}.

\bibitem[{DeepSeek-AI}(2026)]{deepseek2026v4}
{DeepSeek-AI}.
\newblock {DeepSeek-V4 Pro} model card.
\newblock \url{https://huggingface.co/deepseek-ai/DeepSeek-V4-Pro}, 2026.
\newblock Accessed: 2026-04-30.

\par\begingroup
\bibitem[Drouin et~al.(2024)Drouin, Gasse, Caccia, Laradji, Del Verme, Marty,
  Boisvert, Thakkar, Cappart, Vazquez, Chapados, and
  Lacoste]{drouin2024workarena}
A.~Drouin, M.~Gasse, M.~Caccia, I.~H. Laradji, M.~Del Verme, T.~Marty,
  L.~Boisvert, M.~Thakkar, Q.~Cappart, D.~Vazquez, N.~Chapados, and A.~Lacoste.
\newblock {WorkArena}: How capable are web agents at solving common knowledge
  work tasks?
\newblock In \emph{Proceedings of the 41st International Conference on Machine
  Learning (ICML)}, 2024.
\newblock URL \url{https://arxiv.org/abs/2403.07718}.

\par\endgroup


\par\begingroup
\bibitem[Friedman et~al.(2016)Friedman, Howard, Shaw, Cohen, Shahidi, and
  Ferrante]{friedman2016facilitators}
A.~Friedman, J.~Howard, E.~K. Shaw, D.~J. Cohen, L.~Shahidi, and J.~M.
  Ferrante.
\newblock Facilitators and barriers to care coordination in patient-centered
  medical homes ({PCMHs}) from coordinators' perspectives.
\newblock \emph{Journal of the American Board of Family Medicine}, 29\penalty0
  (1):\penalty0 90--101, 2016.
\newblock \doi{10.3122/jabfm.2016.01.150175}.
\newblock PMC4809054.


\par\endgroup
\bibitem[{GLM-5 Team}(2026)]{zhipu2026glm5}
{GLM-5 Team}.
\newblock {GLM-5}: From vibe coding to agentic engineering.
\newblock \emph{arXiv preprint arXiv:2602.15763}, 2026.

\bibitem[{Google}(2025)]{google2025geminicli}
{Google}.
\newblock {Gemini CLI}.
\newblock \url{https://github.com/google-gemini/gemini-cli}, 2025.
\newblock Accessed: 2026-04-30.

\par\begingroup
\bibitem[{Google DeepMind}(2026)]{google2026gemini3}
{Google DeepMind}.
\newblock {Gemini 3.1 Pro} model card.
\newblock \url{https://deepmind.google/models/model-cards/gemini-3-1-pro/},
  2026.
\newblock Accessed: 2026-04-30.

\par\endgroup

\bibitem[{Harbor Framework}(2026)]{harbor2026}
{Harbor Framework}.
\newblock {Harbor}: A framework for agent evaluations and {RL} environments.
\newblock \url{https://github.com/harbor-framework/harbor}, 2026.
\newblock Accessed: 2026-04-30.

\bibitem[Jiang et~al.(2025)Jiang, Black, Geng, Park, Zou, Ng, and
  Chen]{jiang2025medagentbench}
Y.~Jiang, K.~C. Black, G.~Geng, D.~Park, J.~Zou, A.~Y. Ng, and J.~H. Chen.
\newblock Medagentbench: a virtual ehr environment to benchmark medical llm
  agents.
\newblock \emph{Nejm Ai}, 2\penalty0 (9):\penalty0 AIdbp2500144, 2025.

\bibitem[Jimenez et~al.(2024)Jimenez, Yang, Wettig, Yao, Pei, Press, and
  Narasimhan]{jimenez2024swebenchlanguagemodelsresolve}
C.~E. Jimenez, J.~Yang, A.~Wettig, S.~Yao, K.~Pei, O.~Press, and K.~Narasimhan.
\newblock Swe-bench: Can language models resolve real-world github issues?,
  2024.
\newblock URL \url{https://arxiv.org/abs/2310.06770}.

\bibitem[Jin et~al.(2021)Jin, Pan, Oufattole, Weng, Fang, and
  Szolovits]{jin2021medqa}
D.~Jin, E.~Pan, N.~Oufattole, W.-H. Weng, H.~Fang, and P.~Szolovits.
\newblock What disease does this patient have? a large-scale open domain
  question answering dataset from medical exams.
\newblock \emph{Applied Sciences}, 11\penalty0 (14):\penalty0 6421, 2021.
\newblock URL \url{https://arxiv.org/abs/2009.13081}.

\bibitem[Jin et~al.(2019)Jin, Dhingra, Liu, Cohen, and Lu]{jin2019pubmedqa}
Q.~Jin, B.~Dhingra, Z.~Liu, W.~W. Cohen, and X.~Lu.
\newblock {PubMedQA}: A dataset for biomedical research question answering.
\newblock In \emph{Proceedings of the 2019 Conference on Empirical Methods in
  Natural Language Processing and the 9th International Joint Conference on
  Natural Language Processing (EMNLP-IJCNLP)}, pages 2567--2577. Association
  for Computational Linguistics, 2019.
\newblock URL \url{https://arxiv.org/abs/1909.06146}.

\bibitem[Jones and Kelly(2025)]{jones2025code}
A.~Jones and C.~Kelly.
\newblock Code execution with mcp: Building more efficient agents, 2025.


\par\begingroup
\bibitem[Ju(2022)]{ju2022improving}
H.-H. Ju.
\newblock Improving care coordination of patients with chronic diseases.
\newblock \emph{The Journal for Nurse Practitioners}, 18\penalty0 (8):\penalty0
  833--836, 2022.
\newblock \doi{10.1016/j.nurpra.2022.06.011}.


\par\endgroup
\bibitem[Kaelbling et~al.(1998)Kaelbling, Littman, and
  Cassandra]{kaelbling1998planning}
L.~P. Kaelbling, M.~L. Littman, and A.~R. Cassandra.
\newblock Planning and acting in partially observable stochastic domains.
\newblock \emph{Artificial intelligence}, 101\penalty0 (1-2):\penalty0 99--134,
  1998.


\par\begingroup
\bibitem[Karam et~al.(2026)Karam, Chouinard, Kevork, Fleming, and
  Duhoux]{karam2026nurses}
M.~Karam, M.-C. Chouinard, M.~Kevork, R.~Fleming, and A.~Duhoux.
\newblock Nurses' and patients' perspectives on care coordination across health
  care and social services sectors: A qualitative study.
\newblock \emph{Canadian Journal of Nursing Research}, 58\penalty0
  (1):\penalty0 48--57, 2026.
\newblock \doi{10.1177/08445621251395347}.


\par\endgroup
\bibitem[Khandekar et~al.(2024)Khandekar, Jin, Xiong, Dunn, Applebaum, Anwar,
  Sarfo-Gyamfi, Safranek, Anwar, Zhang, Gilson, Singer, Dave, Taylor, Zhang,
  Chen, and Lu]{khandekar2024medcalc}
N.~Khandekar, Q.~Jin, G.~Xiong, S.~Dunn, S.~S. Applebaum, Z.~Anwar,
  M.~Sarfo-Gyamfi, C.~W. Safranek, A.~A. Anwar, A.~Zhang, A.~Gilson, M.~B.
  Singer, A.~Dave, A.~Taylor, A.~Zhang, Q.~Chen, and Z.~Lu.
\newblock {MedCalc-Bench}: Evaluating large language models for medical
  calculations.
\newblock In \emph{Advances in Neural Information Processing Systems 37:
  Datasets and Benchmarks Track}, 2024.
\newblock URL \url{https://arxiv.org/abs/2406.12036}.

\bibitem[{Kimi Team}(2025)]{moonshot2025kimik2}
{Kimi Team}.
\newblock {Kimi K2}: Open agentic intelligence.
\newblock \emph{arXiv preprint arXiv:2507.20534}, 2025.

\bibitem[{LangChain}(2025)]{langchain2025deepagents}
{LangChain}.
\newblock {DeepAgents}.
\newblock \url{https://github.com/langchain-ai/deepagents}, 2025.
\newblock Accessed: 2026-04-30.

\bibitem[Lee et~al.(2022)Lee, Hwang, Bae, Kwon, Shin, Yang, Seo, Kim, and
  Choi]{lee2022ehrsql}
G.~Lee, H.~Hwang, S.~Bae, Y.~Kwon, W.~Shin, S.~Yang, M.~Seo, J.-Y. Kim, and
  E.~Choi.
\newblock {EHRSQL}: A practical text-to-{SQL} benchmark for electronic health
  records.
\newblock In \emph{Advances in Neural Information Processing Systems 35:
  Datasets and Benchmarks Track}, 2022.
\newblock URL \url{https://arxiv.org/abs/2301.07695}.

\bibitem[Li et~al.(2025)Li, Zhao, Zhao, Zeng, Wu, Wang, Ge, Cao, Huang, Liu,
  et~al.]{li2025tool}
J.~Li, W.~Zhao, J.~Zhao, W.~Zeng, H.~Wu, X.~Wang, R.~Ge, Y.~Cao, Y.~Huang,
  W.~Liu, et~al.
\newblock The tool decathlon: Benchmarking language agents for diverse,
  realistic, and long-horizon task execution.
\newblock \emph{arXiv preprint arXiv:2510.25726}, 2025.

\par\begingroup
\bibitem[Li et~al.(2026)Li, Chen, Liu, Zheng, Chen, He, Li, You, Shen, Sun,
  et~al.]{li2026skillsbench}
X.~Li, W.~Chen, Y.~Liu, S.~Zheng, X.~Chen, Y.~He, Y.~Li, B.~You, H.~Shen,
  J.~Sun, et~al.
\newblock Skillsbench: Benchmarking how well agent skills work across diverse
  tasks.
\newblock \emph{arXiv preprint \href{https://arxiv.org/abs/2602.12670v1}{arXiv:2602.12670v1}}, 2026.

\par\endgroup


\par\begingroup
\bibitem[Liu et~al.(2025)Liu, Wang, Ma, Huang, Su, Chang, Li, Shen, Lyu, and
  Chen]{liu2024medchain}
J.~Liu, W.~Wang, Z.~Ma, G.~Huang, Y.~Su, K.-J. Chang, H.~Li, L.~Shen, M.~R.
  Lyu, and W.~Chen.
\newblock {MedChain}: Bridging the gap between {LLM} agents and clinical
  practice with interactive sequence.
\newblock In \emph{Advances in Neural Information Processing Systems 38:
  Datasets and Benchmarks Track}, 2025.
\newblock \doi{10.52202/085713-2498}.


\par\endgroup
\bibitem[Merrill et~al.(2026)Merrill, Shaw, Carlini, Li, Raj, Bercovich, Shi,
  Shin, Walshe, Buchanan, et~al.]{merrill2026terminal}
M.~A. Merrill, A.~G. Shaw, N.~Carlini, B.~Li, H.~Raj, I.~Bercovich, L.~Shi,
  J.~Y. Shin, T.~Walshe, E.~K. Buchanan, et~al.
\newblock Terminal-bench: Benchmarking agents on hard, realistic tasks in
  command line interfaces.
\newblock \emph{arXiv preprint arXiv:2601.11868}, 2026.

\bibitem[{Modal Labs}(2025)]{modal2025}
{Modal Labs}.
\newblock {Modal}: High-performance serverless infrastructure for {AI} and
  data.
\newblock \url{https://modal.com}, 2025.
\newblock Accessed: 2026-04-30.

\bibitem[{Nous Research}(2026)]{hermes2026}
{Nous Research}.
\newblock {Hermes Agent}: The agent that grows with you.
\newblock \url{https://github.com/NousResearch/hermes-agent}, 2026.
\newblock Accessed: 2026-04-30.

\bibitem[{OpenAI}(2025{\natexlab{a}})]{openai2025agentssdk}
{OpenAI}.
\newblock {OpenAI Agents SDK} (python).
\newblock \url{https://github.com/openai/openai-agents-python},
  2025{\natexlab{a}}.
\newblock Accessed: 2026-04-30.

\bibitem[{OpenAI}(2025{\natexlab{b}})]{openai2025codexcli}
{OpenAI}.
\newblock {OpenAI Codex CLI}.
\newblock \url{https://github.com/openai/codex}, 2025{\natexlab{b}}.
\newblock Accessed: 2026-04-30.

\par\begingroup
\bibitem[{OpenAI}(2026)]{openai2026gpt5}
{OpenAI}.
\newblock {GPT-5.5} system card.
\newblock \url{https://deploymentsafety.openai.com/gpt-5-5}, 2026.
\newblock Accessed: 2026-04-30.

\par\endgroup

\bibitem[{OpenClaw}(2025)]{openclaw2025}
{OpenClaw}.
\newblock {OpenClaw}: Your own personal ai assistant.
\newblock \url{https://github.com/openclaw/openclaw}, 2025.
\newblock Accessed: 2026-04-30.

\bibitem[Pal et~al.(2022)Pal, Umapathi, and Sankarasubbu]{pal2022medmcqa}
A.~Pal, L.~K. Umapathi, and M.~Sankarasubbu.
\newblock {MedMCQA}: A large-scale multi-subject multi-choice dataset for
  medical domain question answering.
\newblock In \emph{Proceedings of the Conference on Health, Inference, and
  Learning (CHIL)}, volume 174 of \emph{Proceedings of Machine Learning
  Research}, pages 248--260. PMLR, 2022.
\newblock URL \url{https://arxiv.org/abs/2203.14371}.

\bibitem[{Qwen Team}(2025)]{alibaba2025qwen3}
{Qwen Team}.
\newblock {Qwen3} technical report.
\newblock \emph{arXiv preprint arXiv:2505.09388}, 2025.

\bibitem[Sahni et~al.(2023)Sahni, Gupta, Peterson, and Cutler]{sahni2023active}
N.~R. Sahni, P.~Gupta, M.~Peterson, and D.~M. Cutler.
\newblock Active steps to reduce administrative spending associated with
  financial transactions in {US} health care.
\newblock \emph{Health Affairs Scholar}, 1\penalty0 (5):\penalty0 qxad053,
  2023.
\newblock \doi{10.1093/haschl/qxad053}.


\par\begingroup
\bibitem[Sahni et~al.(2024)Sahni, Istvan, Stafford, and
  Cutler]{sahni2024perceptions}
N.~R. Sahni, B.~Istvan, C.~Stafford, and D.~M. Cutler.
\newblock Perceptions of prior authorization burden and solutions.
\newblock \emph{Health Affairs Scholar}, 2\penalty0 (9):\penalty0 qxae096,
  2024.
\newblock \doi{10.1093/haschl/qxae096}.


\par\endgroup
\bibitem[Schmidgall et~al.(2024)Schmidgall, Ziaei, Harris, Reis, Jopling, and
  Moor]{schmidgall2024agentclinic}
S.~Schmidgall, R.~Ziaei, C.~Harris, E.~Reis, J.~Jopling, and M.~Moor.
\newblock {AgentClinic}: a multimodal agent benchmark to evaluate {AI} in
  simulated clinical environments, 2024.
\newblock URL \url{https://arxiv.org/abs/2405.07960}.

\bibitem[Sinsky et~al.(2016)Sinsky, Colligan, Li, Prgomet, Reynolds, Goeders,
  Westbrook, Tutty, and Blike]{sinsky2016allocation}
C.~A. Sinsky, L.~Colligan, L.~Li, M.~Prgomet, S.~Reynolds, L.~Goeders,
  J.~Westbrook, M.~Tutty, and G.~Blike.
\newblock Allocation of physician time in ambulatory practice: A time and
  motion study in 4 specialties.
\newblock \emph{Annals of Internal Medicine}, 165\penalty0 (11):\penalty0
  753--760, 2016.
\newblock \doi{10.7326/M16-0961}.

\bibitem[Steinberger(2025)]{steipete2025mcporter}
P.~Steinberger.
\newblock {MCPorter}: {TypeScript} runtime and {CLI} for connecting to {MCP}
  servers.
\newblock \url{https://github.com/steipete/mcporter}, 2025.
\newblock npm package \texttt{mcporter}; accessed 2026-05-03.

\bibitem[Sutton et~al.(1999)Sutton, Precup, and Singh]{sutton1999between}
R.~S. Sutton, D.~Precup, and S.~Singh.
\newblock Between mdps and semi-mdps: A framework for temporal abstraction in
  reinforcement learning.
\newblock \emph{Artificial intelligence}, 112\penalty0 (1-2):\penalty0
  181--211, 1999.

\bibitem[Tang et~al.(2024)Tang, Qian, Gao, Chen, Chen, and
  Gerstein]{tang2024biocoder}
X.~Tang, B.~Qian, R.~Gao, J.~Chen, X.~Chen, and M.~Gerstein.
\newblock {BioCoder}: a benchmark for bioinformatics code generation with large
  language models.
\newblock \emph{Bioinformatics}, 40\penalty0 (Supplement\_1):\penalty0
  i266--i276, 2024.
\newblock \doi{10.1093/bioinformatics/btae230}.
\newblock URL \url{https://arxiv.org/abs/2308.16458}.

\par\begingroup
\bibitem[Tang et~al.(2025)Tang, Shao, Sohn, Chen, Zhang, Xiang, Wu, Zhao, Wu,
  Shi, Cohan, and Gerstein]{tang2025medagentsbench}
X.~Tang, D.~Shao, J.~Sohn, J.~Chen, J.~Zhang, J.~Xiang, F.~Wu, Y.~Zhao, C.~Wu,
  W.~Shi, A.~Cohan, and M.~Gerstein.
\newblock {MedAgentsBench}: Benchmarking thinking models and agent frameworks
  for complex medical reasoning, 2025.
\newblock URL \url{https://arxiv.org/abs/2503.07459v1}.

\par\endgroup

\bibitem[Trivedi et~al.(2024)Trivedi, Khot, Hartmann, Manku, Dong, Li, Gupta,
  Sabharwal, and Balasubramanian]{trivedi2024appworld}
H.~Trivedi, T.~Khot, M.~Hartmann, R.~Manku, V.~Dong, E.~Li, S.~Gupta,
  A.~Sabharwal, and N.~Balasubramanian.
\newblock {AppWorld}: A controllable world of apps and people for benchmarking
  interactive coding agents.
\newblock In \emph{Proceedings of the 62nd Annual Meeting of the Association
  for Computational Linguistics (ACL)}. Association for Computational
  Linguistics, 2024.
\newblock URL \url{https://arxiv.org/abs/2407.18901}.

\bibitem[Tsatsaronis et~al.(2015)Tsatsaronis, Balikas, Malakasiotis, Partalas,
  Zschunke, Alvers, Weissenborn, Krithara, Petridis, Polychronopoulos,
  Almirantis, Pavlopoulos, Baskiotis, Gallinari, Arti{\`e}res, Ngomo, Heino,
  Gaussier, Barrio-Alvers, Schroeder, Androutsopoulos, and
  Paliouras]{tsatsaronis2015bioasq}
G.~Tsatsaronis, G.~Balikas, P.~Malakasiotis, I.~Partalas, M.~Zschunke, M.~R.
  Alvers, D.~Weissenborn, A.~Krithara, S.~Petridis, D.~Polychronopoulos,
  Y.~Almirantis, J.~Pavlopoulos, N.~Baskiotis, P.~Gallinari, T.~Arti{\`e}res,
  A.-C.~N. Ngomo, N.~Heino, E.~Gaussier, L.~Barrio-Alvers, M.~Schroeder,
  I.~Androutsopoulos, and G.~Paliouras.
\newblock An overview of the {BIOASQ} large-scale biomedical semantic indexing
  and question answering competition.
\newblock \emph{BMC Bioinformatics}, 16\penalty0 (1):\penalty0 138, 2015.
\newblock \doi{10.1186/s12859-015-0564-6}.
\newblock URL
  \url{https://bmcbioinformatics.biomedcentral.com/articles/10.1186/s12859-015-0564-6}.

\bibitem[Wang et~al.(2024)Wang, Danek, Yang, Chen, and Sun]{wang2024biodsbench}
Z.~Wang, B.~Danek, Z.~Yang, Z.~Chen, and J.~Sun.
\newblock Can large language models replace data scientists in biomedical
  research?, 2024.
\newblock URL \url{https://arxiv.org/abs/2410.21591}.

\bibitem[Wornow et~al.(2023)Wornow, Thapa, Steinberg, Fries, and
  Shah]{wornow2023ehrshot}
M.~Wornow, R.~Thapa, E.~Steinberg, J.~A. Fries, and N.~H. Shah.
\newblock {EHRSHOT}: An {EHR} benchmark for few-shot evaluation of foundation
  models.
\newblock In \emph{Advances in Neural Information Processing Systems 36:
  Datasets and Benchmarks Track}, 2023.
\newblock URL \url{https://arxiv.org/abs/2307.02028}.

\bibitem[{xAI}(2025)]{xai2025grok4}
{xAI}.
\newblock {Grok 4} model card.
\newblock \url{https://data.x.ai/2025-08-20-grok-4-model-card.pdf}, 2025.
\newblock Accessed: 2026-04-30.

\bibitem[Xie et~al.(2024)Xie, Zhang, Chen, Li, Zhao, Cao, Hua, Cheng, Shin,
  Lei, Liu, Xu, Zhou, Savarese, Xiong, Zhong, and Yu]{xie2024osworld}
T.~Xie, D.~Zhang, J.~Chen, X.~Li, S.~Zhao, R.~Cao, T.~J. Hua, Z.~Cheng,
  D.~Shin, F.~Lei, Y.~Liu, Y.~Xu, S.~Zhou, S.~Savarese, C.~Xiong, V.~Zhong, and
  T.~Yu.
\newblock {OSWorld}: Benchmarking multimodal agents for open-ended tasks in
  real computer environments.
\newblock In \emph{Advances in Neural Information Processing Systems 37:
  Datasets and Benchmarks Track}, 2024.
\newblock URL \url{https://arxiv.org/abs/2404.07972}.

\bibitem[Xiong et~al.(2024)Xiong, Jin, Lu, and Zhang]{xiong2024mirage}
G.~Xiong, Q.~Jin, Z.~Lu, and A.~Zhang.
\newblock Benchmarking retrieval-augmented generation for medicine.
\newblock In \emph{Findings of the Association for Computational Linguistics:
  ACL 2024}. Association for Computational Linguistics, 2024.
\newblock URL \url{https://arxiv.org/abs/2402.13178}.

\bibitem[Xu et~al.(2024)Xu, Song, Li, Tang, Jain, Bao, Wang, Zhou, Guo, Cao,
  Yang, Lu, Martin, Su, Maben, Mehta, Chi, Jang, Xie, Zhou, and
  Neubig]{xu2024agentcompany}
F.~F. Xu, Y.~Song, B.~Li, Y.~Tang, K.~Jain, M.~Bao, Z.~Z. Wang, X.~Zhou,
  Z.~Guo, M.~Cao, M.~Yang, H.~Y. Lu, A.~Martin, Z.~Su, L.~Maben, R.~Mehta,
  W.~Chi, L.~Jang, Y.~Xie, S.~Zhou, and G.~Neubig.
\newblock {TheAgentCompany}: Benchmarking {LLM} agents on consequential real
  world tasks, 2024.
\newblock URL \url{https://arxiv.org/abs/2412.14161}.


\par\begingroup
\bibitem[Xu et~al.(2026)Xu, Zhuang, Zhong, Yu, Wang, Tang, Wu, Wang, Ruan,
  Yang, Wang, Xiao, Liu, Yang, Xie, and Shi]{shi2025medagentgym}
R.~Xu, Y.~Zhuang, Y.~Zhong, Y.~Yu, Z.~Wang, X.~Tang, H.~Wu, M.~D. Wang,
  P.~Ruan, D.~Yang, T.~Wang, G.~Xiao, X.~Liu, C.~Yang, Y.~Xie, and W.~Shi.
\newblock {MedAgentGym}: A scalable agentic training environment for
  code-centric reasoning in biomedical data science.
\newblock In \emph{International Conference on Learning Representations
  (ICLR)}, 2026.
\newblock URL \url{https://arxiv.org/abs/2506.04405}.


\par\endgroup
\bibitem[Yao et~al.(2024)Yao, Shinn, Razavi, and Narasimhan]{yao2024tau}
S.~Yao, N.~Shinn, P.~Razavi, and K.~Narasimhan.
\newblock {$\tau$}-bench: A benchmark for tool-agent-user interaction in
  real-world domains.
\newblock \emph{arXiv preprint arXiv:2406.12045}, 2024.

\bibitem[Zhang et~al.(2025)Zhang, Lazuka, and Murag]{zhang2025equipping}
B.~Zhang, K.~Lazuka, and M.~Murag.
\newblock Equipping agents for the real world with agent skills.
\newblock \emph{Anthropic Engineering Blog}, 2025.

\bibitem[Zhou et~al.(2024)Zhou, Xu, Zhu, Zhou, Lo, Sridhar, Cheng, Ou, Bisk,
  Fried, Alon, and Neubig]{zhou2024webarena}
S.~Zhou, F.~F. Xu, H.~Zhu, X.~Zhou, R.~Lo, A.~Sridhar, X.~Cheng, T.~Ou,
  Y.~Bisk, D.~Fried, U.~Alon, and G.~Neubig.
\newblock {WebArena}: A realistic web environment for building autonomous
  agents.
\newblock In \emph{International Conference on Learning Representations
  (ICLR)}, 2024.
\newblock URL \url{https://arxiv.org/abs/2307.13854}.

\bibitem[Zuo et~al.(2025)Zuo, Qu, Li, Chen, Zhu, Hua, Zhang, Ding, and
  Zhou]{zuo2025medxpertqa}
Y.~Zuo, S.~Qu, Y.~Li, Z.~Chen, X.~Zhu, E.~Hua, K.~Zhang, N.~Ding, and B.~Zhou.
\newblock Medxpertqa: Benchmarking expert-level medical reasoning and
  understanding.
\newblock \emph{arXiv preprint arXiv:2501.18362}, 2025.

\end{thebibliography}
\end{document}